\documentclass{article} %
\usepackage{iclr2022_conference,times}

\usepackage{amsmath, amssymb, natbib, graphicx, url} 
\usepackage[linesnumbered,ruled,algo2e]{algorithm2e}%
\usepackage{etoolbox}
\makeatletter
\patchcmd{\@algocf@start}%
  {-1.5em}%
  {0pt}%
  {}{}%
\makeatother

\title{Generalized Kernel Thinning}
\author{%
Raaz Dwivedi$^1$, 
Lester Mackey$^2$\\
  $^1$ Department of Computer Science, Harvard University and 
  Department of EECS, MIT \\ %
  $^2$ Microsoft Research New England \\
  \texttt{raaz@mit.edu}, \texttt{lmackey@microsoft.com} %
}

\definecolor{mydarkblue}{rgb}{0,0.08,0.45}
\usepackage[colorlinks,citecolor=mydarkblue,urlcolor=black,linkcolor=mydarkblue,pdfborder={0 0 0},pagebackref=true]{hyperref}
\renewcommand*{\backrefalt}[4]{%
    \ifcase #1 \footnotesize{(Not cited.)}%
    \or        \footnotesize{(Cited on page~#2.)}%
    \else      \footnotesize{(Cited on pages~#2.)}%
    \fi}

\usepackage{booktabs} %
\usepackage{caption}
\usepackage[usestackEOL]{stackengine}
\usepackage{enumitem}

\usepackage{etoc}

\usepackage{mathrsfs}
\usepackage[capitalize]{cleveref}  

\usepackage{autonum}
\newtheorem{myproposition}{Proposition}

\newtheorem{mycorollary}{Corollary}

\newtheorem{mylemma}{Lemma}

\newtheorem{mydefinition}{Definition}

\newtheorem{myremark}{Remark}

\crefname{appendix}{App.}{Apps.}
\crefname{subsubsubappendix}{App.}{Apps.}
\crefname{equation}{}{}
\crefname{lemma}{Lem.}{Lems.}
\crefname{theorem}{Thm.}{Thms.}
\crefname{Corollary}{Cor.}{Cors.}
\crefname{algorithm}{Alg.}{Algs.}
\crefname{section}{Sec.}{Secs.}
\crefname{table}{Tab.}{Tabs.}
\crefname{remark}{Rem.}{Rems.}
\crefname{definition}{Def.}{Defs.}
\crefname{Proposition}{Prop.}{Props.}
\crefname{myremark}{Rem.}{Rems.}
\crefname{mylemma}{Lem.}{Lems.}
\crefname{mydefinition}{Def.}{Defs.}
\crefname{myproposition}{Prop.}{Props.}
\crefname{mycorollary}{Cor.}{Cors.}
\crefname{enumi}{}{}
\crefname{name}{}{} %
\SetAlCapHSkip{0pt} %
\LinesNotNumbered %

\newcommand{\tgf}[1]{\tup{#1}}

\newcommand{\rkt}{\tsc{root KT}\xspace}
\newcommand{\Rkt}{\tsc{Root KT}\xspace}
\newcommand{\tkt}{\tsc{target KT}\xspace}
\newcommand{\Tkt}{\tsc{Target KT}\xspace}
\newcommand{\akt}{\tsc{power KT}\xspace}

\newcommand{\pin}{\P_{\tgf{in}}}
\newcommand{\pout}{\P_{\tgf{out}}}
\newcommand{\kdiffgen}[1][\l]{\mc W_{#1, \m}}
\newcommand{\eventequi}{\event[\tgf{equi}]}

\newcommand{\restrict}[1][\wtil{\ball}]{\vert_{#1}}
\newcommand{\aroot}{\alpha}
\newcommand{\kroot}[1][\aroot]{\kernel_{#1}}
\newcommand{\kplus}{\kernel^{\dagger}}
\newcommand{\kappaaroot}[1][\aroot]{\kappa_{#1}}

\newcommand{\kinfsin}[1][\kernel]{\Vert{#1}\Vert_{\infty,\tgf{in}}}
\newcommand{\psplit}{\P_{\tgf{split}}}
\newcommand{\psg}{p_{\tgf{sg}}}

\newcommand{\rin}{\mfk{R}_{\tgf{in}}}

\newcommand{\kball}[1][]{\ball_{\kernel_{#1}}}
\newcommand{\kcover}[1][\kernel]{\cover_{#1, \varepsilon}}

\newcommand{\poutone}{\pout^{(1)}}
\newcommand{\entropy}[1][\kernel]{\mc M_{#1}}

\newcommand{\bspline}{\textsc{B-spline}}
\newcommand{\gauss}{\textsc{Gauss}}
\newcommand{\maternk}{\textsc{\Matern}}
\newcommand{\laplace}{\textsc{Laplace}}
\newcommand{\imq}{\textsc{IMQ}}
\newcommand{\sinc}{\textsc{sinc}}

\newcommand{\vol}{\tgf{Vol}}
\newcommand{\rminnew}[1][]{\rmin_{#1}}

\newcommand{\bndcase}{\textsc{Compact}}
\newcommand{\subgauss}{\textsc{SubGauss}}
\newcommand{\subexp}{\textsc{SubExp}}
\newcommand{\heavytail}{\textsc{HeavyTail}}
\newcommand{\htailparam}{\rho}
\newcommand{\loggrowth}{\textsc{LogGrowth}\xspace}
\newcommand{\polygrowth}{\textsc{PolyGrowth}\xspace}

\newcommand{\order}{\mc{O}}

\newcommand{\mfk}{\mathfrak}

\newcommand{\set}[1]{\mc{#1}}
\newcommand{\ball}{\set{B}}

\newcommand{\indicator}{\mbf 1}

\newcommand{\dims}{d}

\newcommand{\Rd}{\R^{\dims}}

\newcommand{\snorm}[1]{\Vert #1 \Vert}
\newcommand{\sinfnorm}[1]{\snorm{#1}_\infty}

\newcommand{\m}{m}

\newcommand{\fun}{f}

\newcommand{\sless}[1]{\stackrel{#1}{\leq}}

\newcommand{\seq}[1]{\stackrel{#1}{=}}

\newcommand{\x}{x}
\newcommand{\y}{y}
\newcommand{\z}{z}

\newcommand{\axi}[1][i]{\x_{#1}}

\newcommand{\dirac}{\mbi{\delta}}

\newcommand{\kernel}{\mbf{k}}
\newcommand{\kersplit}{\kernel_{\tgf{split}}}
\newcommand{\rkhs}{\mc{H}}
\newcommand{\splitrkhs}{\rkhs_{\trm{split}}}

\newcommand{\knorm}[1]{\Vert{#1}\Vert_{\kernel}}
\newcommand{\krtnorm}[1]{\Vert{#1}\Vert_{\ksqrt}}
\newcommand{\ksplitnorm}[1]{\Vert{#1}\Vert_{\kersplit}}
\newcommand{\ksplitprimenorm}[1]{\Vert{#1}\Vert_{\kersplit'}}

\newcommand{\dotsplitprime}[1]{\angles{#1}_{\kersplit'}}

\DeclareMathOperator{\mmd}{MMD}

\renewcommand{\l}{\ell}
\renewcommand{\j}{j}

\renewcommand{\r}{r}

\def\Matern{Mat\'ern\xspace}

\newcommand{\vareps}{\varepsilon}
\newcommand{\Linf}{L^\infty}

\newcommand{\cover}{\mc C}

\newcommand{\tail}[1][\kernel]{\tau_{#1}}
\newcommand{\klip}[1][\kernel]{L_{#1}}

\newcommand{\rttag}{\tgf{rt}}
\newcommand{\ksqrt}[1][\kernel]{{#1}_{\rttag}}

\newcommand{\funtwo}{g}

\newcommand{\ltwonorm}[1]{\norm{#1}_{2}}

\newcommand{\sgparam}[1][i]{\sigma_{#1}}
\newcommand{\vmax}[1][i]{\mathfrak{b}_{#1}}

\newcommand{\eventnotag}{\mc{E}}
\newcommand{\event}[1][]{\eventnotag_{#1}}

\newcommand{\wtil}[1]{\widetilde{#1}}

\renewcommand{\natural}{\mbb{N}}

\newcommand{\gaussparam}{\sigma}
\newcommand{\matone}{\nu}
\newcommand{\mattwo}{\gamma}
\newcommand{\splineparam}{\beta}
\newcommand{\sincparam}{\theta}
\newcommand{\bessel}[1][\matone]{{K}_{#1}}

\newcommand{\cnew}[1][i]{\mfk{a}_{#1}}

\newcommand{\wvprod}[1][i]{\alpha_{#1}}

\newcommand{\pseqxn}[1][n]{(\axi[i])_{i\geq 1}} %
\newcommand{\pseqxnn}[1][n]{(\axi[i])_{i=1}^n} %

\newcommand{\fourier}{\mc F}

\newcommand{\coveringnumber}{\mc N}

\newcommand{\coreset}[1][j]{\mathcal{S}^{(#1)}}

\SetKwInput{KwInit}{Init}

\newcommand{\brackets}[1]{\left[ #1 \right]}
\newcommand{\parenth}[1]{\left( #1 \right)}
\newcommand{\bigparenth}[1]{\big( #1 \big)}

\newcommand{\braces}[1]{\left\{ #1 \right \}}
\newcommand{\abss}[1]{\left| #1 \right |}
\newcommand{\sabss}[1]{\vert #1 \vert}
\newcommand{\angles}[1]{\left\langle #1 \right \rangle}

\newcommand{\tp}{^\top}
\newcommand{\inv}{^{-1}}
\newcommand{\real}{\ensuremath{\mathbb{R}}}
\newcommand{\Exs}{\ensuremath{{\mathbb{E}}}}

\newcommand{\para}[1]{\noindent\tbf{#1}\ \ }

\def\balign#1\ealign{\begin{align}#1\end{align}}
\def\baligns#1\ealigns{\begin{align*}#1\end{align*}}
\def\balignat#1\ealign{\begin{alignat}#1\end{alignat}}
\def\balignats#1\ealigns{\begin{alignat*}#1\end{alignat*}}
\def\bitemize#1\eitemize{\begin{itemize}#1\end{itemize}}
\def\benumerate#1\eenumerate{\begin{enumerate}#1\end{enumerate}}

\newenvironment{talign*}
 {\csname align*\endcsname}
 {\endalign}
\newenvironment{talign}
 {\csname align\endcsname}
 {\endalign}

\def\balignst#1\ealignst{\begin{talign*}#1\end{talign*}}
\def\balignt#1\ealignt{\begin{talign}#1\end{talign}}
\newcommand{\qtext}[1]{\quad\text{#1}\quad} 

\let\originalleft\left
\let\originalright\right
\renewcommand{\left}{\mathopen{}\mathclose\bgroup\originalleft}
\renewcommand{\right}{\aftergroup\egroup\originalright}

\def\Holder{H\"older\xspace}

\def\Matern{Mat\'ern\xspace}

\def\tinycitep*#1{{\tiny\citep*{#1}}}
\def\tinycitealt*#1{{\tiny\citealt*{#1}}}
\def\tinycite*#1{{\tiny\cite*{#1}}}
\def\smallcitep*#1{{\scriptsize\citep*{#1}}}
\def\smallcitealt*#1{{\scriptsize\citealt*{#1}}}
\def\smallcite*#1{{\scriptsize\cite*{#1}}}

\def\blue#1{\textcolor{blue}{{#1}}}

\def\mbi#1{\boldsymbol{#1}} %
\def\mbf#1{\mathbf{#1}}
\def\mbb#1{\mathbb{#1}}
\def\mc#1{\mathcal{#1}}
\def\mrm#1{\mathrm{#1}}
\def\trm#1{\textrm{#1}}
\def\tbf#1{\textbf{#1}}
\def\tsc#1{\textsc{#1}}
\def\tup#1{\textup{#1}}

\def\reals{\mathbb{R}} %
\def\R{\mathbb{R}}
\def\Q{\mathbb{Q}}
\def\naturals{\mathbb{N}} %
\def\N{\mathbb{N}}
\def\<{\left\langle} %
\def\>{\right\rangle}

\def\implies{\quad\Longrightarrow\quad}

\def\defeq{\triangleq} %
\def\half{\frac{1}{2}}
\def\quarter{\frac{1}{4}}

\newcommand{\textfrac}[2]{{\textstyle\frac{#1}{#2}}}
\newcommand{\floor}[1]{\lfloor{#1}\rfloor}
\newcommand{\ceil}[1]{\lceil{#1}\rceil}
\def\norm#1{\left\|{#1}\right\|} %
\newcommand{\twonorm}[1]{\norm{#1}_2} %
\newcommand{\infnorm}[1]{\norm{#1}_{\infty}} %
\newcommand{\inner}[2]{\langle{#1},{#2}\rangle} %
\def\what#1{\widehat{#1}}

\def\P{\mbb{P}} %

\newcommand{\iid}{\textrm{i.i.d.}\@\xspace}

\providecommand{\argmin}{\mathop\mathrm{arg min}}

\def\supp#1{\mathrm{supp}({#1})}

\ifdefined\nonewproofenvironments\else
\ifdefined\ispres\else
\newtheorem{theorem}{Theorem}

\newenvironment{proof}{\paragraph{Proof}}{\hfill$\square$\\}
\newenvironment{proofof}[1]{\paragraph{Proof of {#1}}}{\hfill$\square$\\}
\newenvironment{prooffor}[1]{\paragraph{Proof for {#1}}}{\hfill$\square$\\}

\newenvironment{proof-sketch}{\noindent\textbf{Proof Sketch}
  \hspace*{1em}}{\qed\bigskip\\}
\newenvironment{proof-idea}{\noindent\textbf{Proof Idea}
  \hspace*{1em}}{\qed\bigskip\\}
\newenvironment{proof-of-lemma}[1][{}]{\noindent\textbf{Proof of Lemma {#1}}
  \hspace*{1em}}{\qed\\}
\newenvironment{proof-of-theorem}[1][{}]{\noindent\textbf{Proof of Theorem {#1}}
  \hspace*{1em}}{\qed\\}
\newenvironment{proof-attempt}{\noindent\textbf{Proof Attempt}
  \hspace*{1em}}{\qed\bigskip\\}

\newcommand{\cset}{\mc{S}}
\newcommand{\inputcoreset}{\cset_{\tgf{in}}}
\newcommand{\outputcoreset}{\cset_{\tgf{out}}}
\newcommand{\ktcoreset}{\cset_{\tgf{KT}}}
\newcommand{\ktpluscoreset}{\cset_{\tgf{KT+}}}

\newcommand{\basecoreset}{\cset_{\tgf{base}}}

\newcommand{\rmin}{\mfk{R}}
\newcommand{\rminpn}[1][\inputcoreset]{\rmin_{#1}}

\newcommand{\err}{\mathfrak{M}}%

\newcommand{\ktsplit}{\hyperref[algo:ktsplit]{\color{black}{\textsc{kt-split}}}\xspace}
\newcommand{\ktsplitlink}{\hyperref[algo:ktsplit]{\textsc{kt-split}}\xspace}
\newcommand{\ktswap}{\hyperref[algo:ktswap]{\color{black}{\textsc{kt-swap}}}\xspace}
\newcommand{\ktswaplink}{\hyperref[algo:ktswap]{\textsc{kt-swap}}\xspace}

\newcommand{\compresspp}{\textsc{Compress++}\xspace}

\newcommand{\mcw}{\mc W}

\graphicspath{{code/figs/},{figs/}}

\iclrfinalcopy
\begin{document}
\etocdepthtag.toc{mtchapter}
\etocsettagdepth{mtchapter}{subsection}
\etocsettagdepth{mtappendix}{none}

\maketitle

\begin{abstract}%
The kernel thinning (KT) algorithm of Dwivedi and Mackey (2021) compresses a probability distribution more effectively than independent sampling by targeting a reproducing kernel Hilbert space (RKHS) and leveraging a less smooth square-root kernel. Here we provide four improvements. First, we show that KT applied directly to the target RKHS yields tighter, dimension-free guarantees for any kernel, any distribution, and any fixed function in the RKHS. Second, we show that, for analytic kernels like Gaussian, inverse multiquadric, and sinc, target KT admits maximum mean discrepancy (MMD) guarantees comparable to or better than those of square-root KT without making explicit use of a square-root kernel.  Third, we prove that KT with a fractional power kernel yields better-than-Monte-Carlo MMD guarantees for non-smooth kernels, like Laplace and Mat\'ern, that do not have square-roots. Fourth, we establish that KT applied to a sum of the target and power kernels (a procedure we call KT+) simultaneously inherits the improved MMD guarantees of power KT and the tighter individual function guarantees of target KT.  In our experiments with target KT and KT+, we witness significant improvements in integration error even in $100$ dimensions and when compressing challenging differential equation posteriors.
\end{abstract}

\section{Introduction}\label{sec:intro}
A core task in probabilistic inference is learning a compact representation of a probability distribution $\P$.
This problem is usually solved by sampling points $x_1, \dots, x_n$ independently from $\P$ or, if direct sampling is intractable, generating $n$ points from a Markov chain converging to $\P$.
The benefit of these approaches is that they provide asymptotically exact sample estimates $\pin f \defeq \frac1n \sum_{i=1}^n f(x_i)$ for intractable expectations $\P f \defeq \Exs_{X\sim \P}[f(X)]$.
However, they also suffer from a serious drawback: the learned representations are unnecessarily large, requiring $n$ points to achieve $\abss{\P\fun-\pin\fun} = \Theta(n^{-\half})$ integration error. %
These inefficient representations quickly become prohibitive for expensive downstream tasks and function evaluations: for example, in computational cardiology, each function evaluation $f(x_i)$ 
initiates a heart or tissue simulation that consumes 1000s of CPU hours \citep{niederer2011simulating,augustin2016anatomically,strocchi2020simulating}.

To reduce the downstream computational burden, 
a standard practice is to \emph{thin} the initial sample  by discarding every $t$-th sample point \citep{owen2017statistically}.
Unfortunately, standard thinning often results in a substantial loss of accuracy: for example, thinning an \iid or fast-mixing  Markov chain sample from $n$ points to $n^{\frac12}$ points increases integration error from $\Theta(n^{-\frac12})$ to $\Theta(n^{-\frac14})$.

The recent \emph{kernel thinning} (KT) algorithm of \citet{dwivedi2021kernel} addresses this issue by producing thinned coresets with better-than-\iid integration error in a reproducing kernel Hilbert space \citep[RKHS,][]{berlinet2011reproducing}.
 Given a target kernel\footnote{A kernel $\kernel$ is any function that yields positive semi-definite matrices $(\kernel(z_i,z_j))_{i,j=1}^l$  for all inputs $(z_i)_{i=1}^l$.} $\kernel$ and a suitable sequence of input points $\inputcoreset = \pseqxnn$ approximating $\P$, KT returns a subsequence $\outputcoreset$ of $\sqrt{n}$ points with better-than-\iid \emph{maximum mean discrepancy} \citep[MMD,][]{JMLR:v13:gretton12a},%
 \footnote{MMD is a metric for \emph{characteristic} $\kernel$, like those in \cref{tab:kernels}, and controls integration error for all bounded continuous $f$ when $\kernel$ \emph{determines convergence}, like each $\kernel$ in \cref{tab:kernels} except \sinc\, \citep{simon2020metrizing}.}
\begin{talign}
	\mmd_{\kernel}(\P, \pout)&\defeq \sup_{\knorm{\fun}\leq 1}\abss{\P\fun-\pout\fun}
	\qtext{for}
	\pout \defeq \frac{1}{\sqrt n} \sum_{x\in\outputcoreset}\dirac_{x},
	\label{eq:kernel_mmd_distance}
\end{talign}
where $\knorm{\cdot}$ denotes the norm for the RKHS $\rkhs$ associated with $\kernel$.
That is, the KT output admits $o(n^{-\quarter})$ worst-case integration error across the unit ball of $\rkhs$.

KT achieves its improvement with high probability using non-uniform randomness and a less smooth \emph{square-root kernel} $\ksqrt$ satisfying 
\begin{talign}\label{eq:sqrt-kernel}
    \kernel(x,y) = \int_{\Rd} \ksqrt(x, z) \ksqrt(z, y) dz.
\end{talign}
When the input points are sampled \iid or from a fast-mixing Markov chain on $\Rd$, \citeauthor{dwivedi2021kernel} prove that the KT output has, with high probability,   $\order_d(n^{-\frac12}\sqrt{\log n })$-$\mmd_{\kernel}$ error for $\P$ and $\ksqrt$ with bounded support,  $\order_d(n^{-\frac12}(\log^{d+1} n \log\log n)^{\half})$-$\mmd_{\kernel}$ error for $\P$ and $\ksqrt$ with light tails, and  $\order_d(n^{-\frac12+\frac{d}{2\htailparam}}\sqrt{\log n \log \log n})$-$\mmd_{\kernel}$ error for $\P$ and $\ksqrt^2$ with $\htailparam > 2d$ moments. 
Meanwhile, an \iid coreset of the same size suffers $\Omega(n^{-\quarter})$ $\mmd_{\kernel}$.
We refer to the original KT algorithm as \tsc{root KT} hereafter.

\para{Our contributions}
In this work, we offer four improvements over the original KT algorithm.
First, we show in \cref{sub:tkt_single_func} that a generalization of KT that uses only the target kernel $\kernel$ provides a tighter $\order(n^{-\half}\sqrt{\log n})$ integration error guarantee for each function $f$ in the RKHS. 
This \tkt guarantee (a) applies to \tbf{any kernel $\kernel$ on any domain} (even kernels that do not admit a square-root and kernels defined on non-Euclidean spaces), (b) applies to \tbf{any target distribution $\P$} (even heavy-tailed $\P$ not covered by \rkt guarantees), and (c) is \tbf{dimension-free}, eliminating the exponential dimension dependence and $(\log n)^{d/2}$ factors of prior \rkt guarantees.

Second, we prove in \cref{sub:tkt_mmd} that, for analytic kernels, like Gaussian,  inverse multiquadric (IMQ), and sinc, \tkt admits MMD guarantees comparable to or better than those of \citet{dwivedi2021kernel} without making explicit use of a square-root kernel.
Third, we establish in \cref{sub:power_kt} that generalized KT with a fractional $\aroot$-power kernel $\kroot$ yields improved MMD guarantees for kernels that do not admit a square-root, like Laplace and non-smooth \Matern.
Fourth, we show in \cref{sec:kt_plus} that, remarkably, applying generalized KT to a sum of $\kernel$ and $\kroot$---a procedure we call \emph{kernel thinning+} (KT+)---simultaneously inherits the improved MMD of \akt and the dimension-free individual function guarantees of \tkt.

In \cref{sec:experiments}, we use our new tools to generate substantially compressed representations of both \iid samples in dimensions $d=2$ through $100$ and Markov chain Monte Carlo samples targeting challenging differential equation posteriors. In line with our theory, we find that \tkt and KT+ significantly improve both single function integration error and MMD, even for kernels without fast-decaying square-roots.

\newcommand{\highlight}[1]{\blue{\mbi{#1}}}
\begin{table}[ht!]
    \centering
     \resizebox{\textwidth}{!}{
    {\renewcommand{\arraystretch}{1}
    \begin{tabular}{cccccccc}
    \toprule
    & \Centerstack{$\gauss(\gaussparam)$\\ $\gaussparam>0$}
    & \Centerstack{$\laplace(\gaussparam)$\\ $\gaussparam>0$}
    & \Centerstack{\maternk$(\matone, \mattwo)$\\ $\matone>\frac d2, \mattwo>0$ }
    & \Centerstack{\imq$(\matone, \mattwo)$\\ $\matone>0, \mattwo>0$ }
    & \Centerstack{\sinc$(\sincparam)$ \\ $\sincparam \neq 0$} 
    & \Centerstack{\bspline$(2\splineparam\!+\!1,\mattwo)$ \\ $\splineparam\in\natural$}
    \\[2mm]
        \midrule 
        & \Centerstack{$\exp\parenth{-\frac{\twonorm{\z}^2}{2\gaussparam^2}}$ }
        & \Centerstack{$\exp\parenth{-\frac{\twonorm{\z}}{\gaussparam}}$}
        & \Centerstack{$ c_{\matone-\frac{d}{2}}(\mattwo\twonorm{\z})^{\matone-\frac{d}{2}}$
         \\$\cdot \bessel[\matone-\frac{d}{2}](\mattwo\twonorm{\z})$}
        & $\frac{1}{(1+\twonorm{\z}^2/\mattwo^2)^{\matone}}$

        &  $\prod_{j=1}^d \frac{\sin(\sincparam z_j)}{\sincparam z_j}$
        
        & \Centerstack{$\mfk{B}_{2\splineparam+2}^{-d} \prod_{\j=1}^d h_{\splineparam}(\mattwo\z_{\j}) $}
          \\[1ex] \bottomrule \hline
    \end{tabular}
    }
    }
    \caption{\tbf{Common kernels $\kernel(x,y)$ on $\reals^d$  with $z=x-y$.}
    In each case, %
    $\infnorm{\kernel}\! =\! 1$. 
    Here,  $c_{a} \!\defeq \!\frac{2^{1\!-\!a}}{\Gamma(a)}$, $\bessel[a]$ is the modified Bessel function of the third kind of order $a$ \citep[Def.~5.10]{wendland2004scattered}, 
    $ h_\beta$ 
    is the recursive convolution of $2\splineparam+2$ copies of $\indicator_{[-\half, \half]}$, and
    $\mfk{B}_{\splineparam} \!\defeq\! \frac{1}{(\splineparam\!-\!1)!} \sum_{j=0}^{\lfloor\splineparam/2\rfloor} (\!-\!1)^j {\splineparam \choose j} (\frac{\splineparam}{2}\! -\! j)^{\splineparam\!-\!1}$.
    }
    \label{tab:kernels}
\end{table}

\para{Related work}
For bounded $\kernel$, both \iid samples~\citep[Prop.~A.1]{tolstikhin2017minimax} and thinned geometrically ergodic Markov chains~\cite[Prop.~1]{dwivedi2021kernel} deliver $n^{\half}$ points with $\order(n^{-\quarter})$ MMD with high probability.
The \emph{online Haar strategy} of \citet{dwivedi2019power} and low discrepancy \emph{quasi-Monte Carlo} methods \citep[see, e.g.,][]{hickernell1998generalized,novak2010tractability,dick2013high} provide improved  $\order_d(n^{-\frac{1}{2}}\log^d n)$ MMD guarantees but are tailored specifically to the uniform distribution on $[0,1]^d$.
Alternative coreset constructions for more general $\P$ include \emph{kernel herding}~\citep{chen2012super},  \emph{discrepancy herding}~\citep{harvey2014near},  \emph{super-sampling with a reservoir}~\citep{paige2016super}, \emph{support points convex-concave procedures}~\citep{mak2018support}, \emph{greedy sign selection}~\citep[Sec.~3.1]{karnin2019discrepancy}, \emph{Stein point MCMC}~\citep{chen2019stein}, and \emph{Stein thinning}~\citep{riabiz2020optimal}. 
While some admit better-than-\iid MMD guarantees for finite-dimensional kernels on $\reals^d$ \citep{chen2012super,harvey2014near}, none apart from KT are known to provide better-than-\iid MMD or integration error for the infinite-dimensional kernels covered in this work.
The lower bounds of \citet[Thm.~3.1]{phillips2020near} and \citet[Thm.~1]{tolstikhin2017minimax} respectively
establish that any procedure outputting $n^{\half}$-sized coresets and any procedure estimating $\P$ based only on $n$ \iid sample points must incur $\Omega(n^{-\half})$ MMD in the worst case.
Our guarantees in \cref{sec:kernel_thinning} match these lower bounds up to logarithmic factors.

\para{Notation} 
We define the norm 
$\infnorm{\kernel} = \sup_{x,y} |\kernel(x,y)|$ 
and the shorthand
$[n] \defeq \{1,\dots,n\}$, 
$\real_{+} \defeq \braces{x\in \real: x\geq 0}$, 
$\natural_0 \defeq \natural \cup \braces{0}$,
 $\ball_{\kernel}\defeq\braces{\fun\in\rkhs: \knorm{\fun} \leq 1}$,  
 and
$\ball_2(r)\defeq\braces{y\in\real^d: \twonorm{y}\leq r}$.
 We write $a\precsim b$ and $a\succsim b$ to mean $a = \order(b)$ and $a =\Omega(b)$, use $\precsim_d$ when masking constants dependent on $d$, and write $a = \order_p(b)$ to mean $a/b$ is bounded in probability.
 For any distribution $\Q$ and point sequences $\mc{S}, \mc{S}'$ with empirical distributions $\Q_n, \Q_n'$, we define $\mmd_{\kernel}(\Q, \mc{S}) \defeq \mmd_{\kernel}(\Q, \Q_n)$ and  $\mmd_{\kernel}(\mc{S}, \mc{S}') \defeq \mmd_{\kernel}(\Q_n, \Q_n')$. 
\newcommand{\mmdtablename}{Kernel thinning MMD guarantee under $\P$ tail decay and scaling of $\entropy$}
\newcommand{\explicitmmd}{Explicit MMD results for common kernels}
\newcommand{\tktcoreset}{\cset_{\mrm{tKT}}}
\newcommand{\mmdguaranteeresultname}{MMD guarantee for \tkt}

\section{Generalized Kernel Thinning} %
\label{sec:kernel_thinning}

Our generalized kernel thinning algorithm (\cref{algo:kernel_thinning}) for compressing an input point sequence $\inputcoreset = \pseqxnn$ proceeds in two steps: \ktsplitlink and \ktswaplink detailed in \cref{sec:algo}.
First, given a thinning parameter $m$ and an auxiliary kernel $\kersplit$, 
\ktsplit divides the input sequence into $2^m$ candidate coresets of size $n/2^m$ using non-uniform randomness. 
Next, given a target kernel $\kernel$, \ktswap selects a candidate coreset with smallest $\mmd_\kernel$ to $\inputcoreset$ and iteratively improves that coreset by exchanging coreset points for input points whenever the swap leads to reduced $\mmd_\kernel$. 
When $\kersplit$ is a square-root kernel $\ksqrt$ \cref{eq:sqrt-kernel} of $\kernel$, generalized KT recovers the original \rkt algorithm of \citeauthor{dwivedi2021kernel}. 
In this section, we establish performance guarantees for more general $\kersplit$ with special emphasis on the practical choice  $\kersplit=\kernel$.  
Like \rkt, for any $m$, generalized KT has time complexity dominated by $\order(n^2)$ evaluations of $\kersplit$ and $\kernel$ 
and $\order(n\min(d,n))$ space complexity from storing either $\inputcoreset$ or the kernel  matrices $(\kersplit(\x_i,\x_j))_{i,j=1}^n$ and $(\kernel(\x_i,\x_j))_{i,j=1}^n$. 
\vspace{-2mm}

\newcommand{\ksplitcoresets}{(\coreset[\m, \l])_{\l=1}^{2^{\m}}}

\begin{algorithm2e}[ht!]
\caption{Generalized Kernel Thinning\ --\ Return coreset of size $\floor{n/2^m}$ with small $\mmd_{\kernel}$} 
  \label{algo:kernel_thinning}
  \SetAlgoLined
  \DontPrintSemicolon
  \SetKwFunction{algo}{algo}
  \SetKwFunction{ksplit}{\textsc{KT-SPLIT}}
  \SetKwFunction{kswap}{{KT-SWAP}}
  \small
  {
  \KwIn{\textup{split kernel $\kersplit$, target kernel $\kernel$, point sequence $\inputcoreset = (\axi[i])_{i = 1}^n$, thinning parameter $\m \in \natural$, probabilities $(\delta_i)_{i = 1}^{\floor{n/2}}$}}
  \BlankLine
    $\ksplitcoresets \gets$ {\normalsize\ktsplitlink}\,$(\kersplit, \inputcoreset, \m, (\delta_i)_{i = 1}^{\floor{n/2}})$ \ // \textup{Split $\inputcoreset$ into $2^\m$ candidate coresets of size $\floor{\frac{n}{2^\m}}$} \\[2pt]
    \BlankLine
    $\ \ktcoreset \quad\quad\ \ \ \ \, \gets$ {\normalsize\ktswaplink}\,$(\kernel, \inputcoreset, \ksplitcoresets)$  \ \ \ \quad// \textup{Select best coreset and iteratively refine} \\
    \BlankLine
  \KwRet{\textup{coreset $\ktcoreset$ of size $\floor{n/2^\m}$}}
}
\end{algorithm2e} 
\vspace{-2mm}

\subsection{Single function guarantees for  \ktsplit} %
\label{sub:tkt_single_func}
We begin by analyzing the quality of the \ktsplit coresets. 
Our first main result, proved in \cref{sub:proof_of_theorem:single_function_guarantee}, 
bounds the \ktsplit integration error for any fixed function in the RKHS $ \rkhs_{\tgf{split}}$ generated by 
$\kersplit$. 

\newcommand{\singlefunctionguaranteeresultname}{Single function guarantees for \ktsplitlink}
\begin{theorem}[\singlefunctionguaranteeresultname]%
\label{theorem:single_function_guarantee}
Consider \ktsplit (\cref{algo:ktsplit}) with oblivious\footnote{Throughout, \emph{oblivious} indicates that a sequence is generated independently of any randomness in KT.} $\inputcoreset$ and $(\delta_i)_{i=1}^{n/2}$ and  $\delta^\star \defeq \min_i\delta_i$. If $\textfrac{n}{2^{\m}}\!\in\! \natural$, then, for any fixed $\fun \in \rkhs_{\trm{split}}$, index $\ell \in [2^\m]$, and scalar $\delta'\! \in\! (0, 1)$, the output coreset $\coreset[\m, \ell]$ with $\psplit^{(\ell)}\defeq\frac{1}{n/2^\m}\sum_{x\in\coreset[\m, \ell]}\dirac_x$
satisfies
    \begin{talign}
      |{\pin\fun - \psplit^{(\ell)}\fun}|
  &\leq  \norm{\fun}_{\kersplit}   \cdot \sigma_{m} \sqrt{ 2\log(\frac{2}{\delta'})} 
  \qtext{for}
  \sigma_m \defeq \frac{2}{\sqrt{3}} \frac{2^\m}{n}\sqrt{ \kinfsin[\kersplit] \cdot \log(\frac{6m}{2^m\delta^\star})}
\label{eq:single_function_guarantee_ktsplit}
       \end{talign}
with probability at least $\psg \!\defeq\! 1\!-\delta'-\!\sum_{j=1}^{\m}\frac{2^{j\!-\!1}}{m} \sum_{i=1}^{n/2^j}\delta_i\!$.
Here, $\kinfsin[\kersplit]\! \defeq \!\max_{x \in \inputcoreset}\kersplit(x,\! x)$.
\end{theorem}
    \begin{remark}[%
    \tbf{Guarantees for known and oblivious stopping times}%
    ]
    \label{rem:hpb_bound}
    By \citet[App.~D]{dwivedi2021kernel}, the success probability 
 $\psg$ is at least $1\!-\!\delta$ if we set $\delta'\!=\!\frac{\delta}{2}$ and $\delta_i\! =\! \frac{\delta}{n}$ for a stopping time $n$ known \emph{a priori} or $\delta_i\! =\! \frac{m\delta}{2^{m+2}(i\!+\!1)\log^2(i\!+\!1)}$ for an arbitrary oblivious stopping time~$n$. %
    \end{remark}

    When compressing heavily from $n$ to $\sqrt{n}$ points, \cref{theorem:single_function_guarantee} and \cref{rem:hpb_bound} guarantee $\order(n^{-\half}\sqrt{\log n})$ integration error with high probability for any fixed function $f\in \splitrkhs$.
    This represents a near-quadratic improvement over the $\Omega(n^{-\frac14})$ integration error of $\sqrt n$ \iid points. 
    Moreover, this guarantee applies to \tbf{any kernel} defined on any space including unbounded kernels on unbounded domains (e.g., energy distance \citep{sejdinovic2013equivalence} and Stein kernels \citep{oates2017control,chwialkowski2016kernel,LiuLeJo16,gorham2017measuring});
    kernels with slowly decaying square roots (e.g., sinc kernels); and non-smooth kernels without square roots (e.g., Laplace, \Matern with $\mattwo \in (\frac d2, d]$), and the compactly supported kernels of  \citet{wendland2004scattered} with $s < \half(d+1)$). 
    In contrast, the MMD guarantees of \citeauthor{dwivedi2021kernel} covered only bounded, smooth $\kernel$ on $\reals^d$ with bounded, Lipschitz, and rapidly-decaying square-roots. 
    In addition, for $\infnorm{\kernel}=1$ on $\reals^d$, the MMD bounds of \citeauthor{dwivedi2021kernel} feature exponential dimension dependence of the form $c^d$ or $(\log n)^{d/2}$ while the \cref{theorem:single_function_guarantee} guarantee is \tbf{dimension-free} and hence practically relevant even when $d$ is  large relative to $n$.
    \cref{theorem:single_function_guarantee} also guarantees better-than-\iid integration error for \tbf{any target distribution} with  $\sabss{\P\fun - \pin\fun} = o(n^{-\quarter})$.
    In contrast, the MMD improvements of \citet[cf.~Tab.~2]{dwivedi2021kernel} applied only to $\P$ with at least $2d$ moments. 
    Finally, when \ktsplit is applied with a square-root kernel $\kersplit = \ksqrt$, \cref{theorem:single_function_guarantee} still yields integration error bounds for $f\in \rkhs$, as $\rkhs \subseteq \rkhs_{\trm{split}}$.
    However, relative to target \ktsplit guarantees with $\kersplit = \kernel$, the error bounds are inflated by a multiplicative factor of $\sqrt{\frac{\kinfsin[\ksqrt]}{\kinfsin}  }\frac{\krtnorm{f}}{\knorm{f}}$.
    In \cref{ratio_of_kernel_norms}, we show that this inflation factor is at least $1$ for each kernel explicitly analyzed in \cite{dwivedi2021kernel} and  grows exponentially in dimension for Gaussian and \Matern kernels, unlike the dimension-free target \ktsplit bounds.%

Finally, if we run \ktsplit with the perturbed kernel $\kersplit'$ defined in \cref{l2pn_guarantee}, then we simultaneously obtain $\order(n^{-\half}\sqrt{\log n})$ integration error for $f\in \splitrkhs$, near-\iid $\order(n^{-\quarter}\sqrt{\log n})$ integration error for arbitrary bounded $f$ outside of $\splitrkhs$, and intermediate, better-than-\iid $o(n^{-\quarter})$ integration error for smoother $f$ outside of $\splitrkhs$ (by interpolation). We prove this guarantee in \cref{sec:proof_of_l2pn_guarantee}.

\begin{corollary}[Guarantees for functions outside of $\splitrkhs$]
\label{l2pn_guarantee}
Consider extending each input point $\x_i$ with the standard basis vector $e_i\in\reals^n$ 
and running \ktsplit (\cref{algo:ktsplit}) on  $\inputcoreset' = (\x_i,e_i)_{i=1}^n$ with 
$\kersplit'((\x,w),(\y,v))=\frac{\kersplit(\x,\y)}{\sinfnorm{\kersplit}} + \inner{w}{v}$ for $w, v, \in \real^n$.
Under the notation and assumptions of \cref{theorem:single_function_guarantee}, 
for any fixed index $\ell \in [2^\m]$, scalar $\delta'\! \in\! (0, 1)$, and $\fun$ defined on $\inputcoreset$, we have, with probability at least $\psg$,
\begin{talign}
      |{\pin\fun - \psplit^{(\ell)}\fun}|
  &\leq  \min(
  \sqrt{\frac{n}{2^m}} \kinfsin[\fun], 
  \sqrt{\infnorm{\kersplit}} \norm{\fun}_{\kersplit}) \frac{2^\m}{n}\sqrt{8 \log(\frac{2}{\delta'}) \cdot \log(\frac{8m}{2^m\delta^\star})}.
\label{eq:l2pn_guarantee}
\end{talign}

\end{corollary}

\subsection{MMD guarantee for target KT}
\label{sub:tkt_mmd}
Our second main result bounds the $\mmd_{\kernel}$ \cref{eq:kernel_mmd_distance}---the worst-case integration error across the unit ball of $\rkhs$---for  generalized KT applied to the target kernel, i.e., $\kersplit=\kernel$.
The proof of this result in \cref{sec:proof_of_theorem_mmd_guarantee} is based on \cref{theorem:single_function_guarantee} and an appropriate covering number for the unit ball $\kball$ of the $\kernel$ RKHS.
\begin{definition}[$\kernel$ covering number]
\label{rkhs_covering}
For a set $\set{A}$ and scalar $\vareps>0$, 
we define the $\kernel$ \emph{covering number} $\coveringnumber_{\kernel}(\set{A}, \vareps)$ 
with $\entropy(\set A, \vareps)\defeq\log\coveringnumber_{\kernel}(\set{A}, \vareps)$
as the minimum cardinality of a set $\cover \subset \kball$ satisfying
\begin{talign}
\label{eq:cover_ball}
\kball \subseteq \bigcup_{h \in \cover} \braces{g \in \kball: \sup_{x\in\set{A}}|h(x)-g(x)|\leq \varepsilon}.
\end{talign}
\end{definition}
\vspace{-5mm}
\begin{theorem}[\mmdguaranteeresultname]%
\label{theorem:mmd_guarantee}
Consider generalized KT (\cref{algo:kernel_thinning}) with $\kersplit=\kernel$, oblivious $\inputcoreset$ and $(\delta_i)_{i=1}^{\floor{n/2}}$, and $\delta^\star \defeq \min_i\delta_i$. If $\textfrac{n}{2^{\m}}\!\in\! \natural$, then for any $\delta'\! \in\! (0, 1)$, the output coreset $\ktcoreset$ is of size $\textfrac{n}{2^{\m}}$ and satisfies 
 \begin{align} 
	\mmd_{\kernel}(\inputcoreset, \ktcoreset)
	\leq \inf_{\vareps \in(0,1),\  \inputcoreset\subset\set{A}}\ 2\varepsilon + 
	\textstyle\frac{2^\m}{n}
	\cdot
	\sqrt{\frac{8}{3}\kinfsin  \log(\frac{6m}{2^m\delta^\star}) \cdot 
	\brackets{ \log(\frac{4}{\delta'})+\entropy(\set{A}, \vareps)}}
\label{eq:mmd_guarantee}
\end{align}
with probability at least $\psg$, where $\kinfsin$ and $\psg$ were defined in
\cref{theorem:single_function_guarantee}.
\end{theorem}

    When compressing heavily from $n$ to $\sqrt{n}$ points, \cref{theorem:mmd_guarantee} and \cref{rem:hpb_bound} with 
    $\vareps=\sqrt{\frac{\kinfsin}{n}}$ and $\set{A} = \ball_2(\rin)$ for $\rin \defeq \max_{x\in\inputcoreset}\twonorm{x}$
    guarantee 
    \begin{talign}
    \label{eq:mmd_simplified_bound}
     \mmd_{\kernel}(\inputcoreset, \ktcoreset)
        \precsim_{\delta} 
        \sqrt{\frac{\kinfsin \log n}{n} \cdot \entropy(\ball_2(\rin), \sqrt{\frac{\kinfsin}{n}})}
    \end{talign}
    with high probability. 
    Thus we immediately obtain an MMD guarantee for any kernel $\kernel$ with a covering number bound. 
    Furthermore, we readily obtain a comparable guarantee for $\P$  since $\mmd_{\kernel}(\P,\ktcoreset) \!\leq\! \mmd_{\kernel}(\P, \inputcoreset) \!+\! \mmd_{\kernel}(\inputcoreset,\ktcoreset)$. 
    Any of a variety of existing algorithms can be used to generate an input point sequence $\inputcoreset$ with $\mmd_{\kernel}(\P, \inputcoreset)$ no larger than the compression bound \cref{eq:mmd_simplified_bound}, including \iid sampling~\citep[Thm.~A.1]{tolstikhin2017minimax}, geometric MCMC~\citep[Prop.~1]{dwivedi2021kernel}, kernel herding \citep[Thm.~G.1]{lacoste2015sequential}, Stein points \citep[Thm.~2]{Chen2018SteinPoints}, Stein point MCMC \citep[Thm.~1]{chen2019stein}, greedy sign selection \citep[Sec.~3.1]{karnin2019discrepancy}, and Stein thinning \citep[Thm.~1]{riabiz2020optimal}.

\subsection{Consequences of \cref{theorem:mmd_guarantee}}
\cref{table:mmd_rates} summarizes the MMD guarantees of \cref{theorem:mmd_guarantee} 
under common growth conditions on the log covering number $\entropy$ and the input point radius $\rminpn[\inputcoreset]\defeq \max_{x\in\inputcoreset} \twonorm{x}$.
In \cref{general_rkhs_covering_bounds,rkhs_covering_numbers} of \cref{sec:bounds_on_rkhs_convering}, we show that analytic kernels, like Gaussian, inverse multiquadric (IMQ), and sinc, have \tbf{\loggrowth} $\entropy$ (i.e., $\entropy(\ball_2(r), \vareps) \precsim_d r^{d} \log^{\omega}(\frac{1}{\varepsilon})$) while finitely differentiable kernels (like \Matern and B-spline) have \tbf{\polygrowth} $\entropy$ (i.e., $\entropy(\ball_2(r), \vareps)\precsim_d r^d \varepsilon^{-\omega}$).

Our conditions on $\rminpn[\inputcoreset]$ arise from four forms of target distribution tail decay: (1) \tbf{\bndcase} ($\rminpn[\inputcoreset] \precsim_d 1$), (2) \tbf{\subgauss} ($\rminpn[\inputcoreset] \precsim_d \sqrt{\log n}$), (3) \tbf{\subexp} ($\rminpn[\inputcoreset] \precsim_d \log n$), and (4) \tbf{\heavytail} ($\rminpn[\inputcoreset] \precsim_d n^{1/\htailparam})$.
The first setting arises with a compactly supported $\P$ (e.g., on the unit cube $[0, 1]^d$), and the other three settings arise in expectation and with high probability when $\inputcoreset$ is generated \iid from~$\P$ with sub-Gaussian tails, sub-exponential tails, or $\rho$ moments respectively. 

Substituting these conditions into \cref{eq:mmd_simplified_bound} yields the eight entries of \cref{table:mmd_rates}. We find that, for \loggrowth $\entropy$, \tkt MMD is within log factors of the $\Omega(n^{-1/2})$ lower bounds of \cref{sec:intro} for light-tailed $\P$ and is  $o(n^{-1/4})$ (i.e., better than \iid) for any distribution with $\htailparam > 2d$ moments.
Meanwhile, for \polygrowth $\entropy$, \tkt MMD is $o(n^{-1/4})$ whenever $\omega<1$ for light-tailed $\P$ or whenever $\P$ has $\htailparam > 2d/(1 - \omega)$ moments. 
\begin{table}[ht!]
    \centering
\small
  {
    {
    \renewcommand{\arraystretch}{1}
    \begin{tabular}{ccccc}
        \toprule
        \Centerstack{ \\  %
        }
        
        & \Centerstack{ \bndcase\ $\P$\\$\rin \precsim_d $ $1$ 
        } 
        
        & \Centerstack{ \subgauss\ $\P$ \\ $\rin \precsim_d $ $\sqrt{\log n}$ 
        } 
        
        & \Centerstack{ \subexp\ $\P$ \\ $\rin \precsim_d $ $\log n$
        } 
       
        & \Centerstack{\heavytail\ $\P$ \\ $\rin \precsim_d $ $n^{1/\htailparam}$ 
        }   
        \\[2mm]
        \midrule 
        
         \Centerstack{ 
         \loggrowth $\entropy$ \\
         $\entropy(\ball_2(r), \vareps)$ 
         $\precsim_d r^{d} \log^{\omega}(\frac{1}{\varepsilon})$
         }
          & $\sqrt{\frac{(\log n)^{\omega+1}}{ n}}$
           & $\sqrt{\frac{(\log n)^{(d/2)+\omega+1}}{ n}}$
          & $\sqrt{\frac{(\log n)^{d+\omega+1}}{ n}}$
          & $\sqrt{\frac{(\log n)^{\omega+1}}{ n^{1-d/\htailparam}}}$
          \\[3mm]

        \Centerstack{ 
        \polygrowth $\entropy$ \\
        $\entropy(\ball_2(r), \vareps)$ $\precsim_d r^d \varepsilon^{-\omega}$
         }
         & $\sqrt{\frac{\log n}{n^{1-\omega/2}}}$
           & $\sqrt{\frac{(\log n)^{(d/2)+1}}{n^{1-\omega/2}}}$
          & $\sqrt{\frac{(\log n)^{d+1}}{n^{1-\omega/2}}}$
          & $\sqrt{\frac{\log n}{n^{1-\omega/2-d/\htailparam}}}$

        \\[1ex] \bottomrule \hline
    \end{tabular}
    }
    }
 \noindent\caption{
    \tbf{MMD guarantees for \tkt under $\entropy$~\cref{eq:cover_ball} growth and $\P$ tail decay.} 
    We report the $\mmd_{\kernel}(\inputcoreset,\ktcoreset)$ bound \cref{eq:mmd_simplified_bound} for target KT with $n$ input points and $\sqrt n$ output points, up to constants depending on $d$ and $\kinfsin$. Here $\rin \defeq \max_{x\in\inputcoreset} \twonorm{x}$.
    } \label{table:mmd_rates}
\end{table}

Next, for each of the popular convergence-determining kernels of \cref{tab:kernels}, we compare the \rkt MMD guarantees of \citet{dwivedi2021kernel} with the \tkt guarantees of \cref{theorem:mmd_guarantee} combined with  covering number bounds derived in \cref{sec:bounds_on_rkhs_convering,proof_of_table_explicit_mmd}.
We see in \cref{table:explicit_mmd} that \cref{theorem:mmd_guarantee}
provides better-than-\iid and better-than-\rkt guarantees for kernels with slowly decaying or non-existent square-roots 
(e.g., IMQ with $\nu < \frac{d}{2}$, sinc, %
and B-spline) and nearly matches known \rkt guarantees for analytic kernels like Gauss and IMQ with $\nu \geq \frac{d}{2}$, even though \tkt makes no explicit use of a square-root kernel. See \cref{proof_of_table_explicit_mmd} for the proofs related to \cref{table:explicit_mmd}.

\newcommand{\rowsep}{4mm}
\newcommand{\na}{N/A}
\newcommand{\cn}{c_n}
\begin{table}[ht!]
    \centering
  \resizebox{1\textwidth}{!}
  {
\small
  {
    \renewcommand{\arraystretch}{1}
    \begin{tabular}{cccc}
        \toprule
        \Centerstack{\bf Kernel $\kernel$}
        & \Centerstack{
        \tbf{\Tkt} %
        } 
         & \Centerstack{
        \tbf{\Rkt} %
        } 
        & \Centerstack{
        \tbf{KT+} %
        } 
        \\
        \midrule 
        
         \Centerstack{
      		$\gauss(\gaussparam)$
      		}
      	& \Centerstack{${\frac{(\log n)^{\frac{3d}{4}+1}}{\sqrt{n \cdot \cn^d}}}$}
    & \Centerstack{$\highlight{\frac{(\log n)^{\frac {d}{4}+\half} \sqrt{\cn}}{\sqrt{n}}}$}
    & \Centerstack{$\highlight{\frac{(\log n)^{\frac {d}{4}+\half} \sqrt{\cn}}{\sqrt{n}}}$}
        
          \\[\rowsep]
        
        \Centerstack{$\laplace(\gaussparam)$}
      	& \Centerstack{$n^{-\quarter}$}
      	& \Centerstack{\na} 
      	& \Centerstack{$\highlight{(\frac{\cn (\log n)^{1\!+\!2d(1\!-\!\aroot)}}{n})^{\frac{1}{4\aroot}}}$}
        
          \\[\rowsep]

        \Centerstack{\maternk$(\matone, \mattwo)$\\
        $\matone \in (\frac d2, d]$} 
        & \Centerstack{
        $n^{-\quarter}$
        }
        & \Centerstack{\na} 
        & \Centerstack{$\highlight{(\frac{\cn (\log n)^{1\!+\!2d(1\!-\!\aroot)}}{n})^{\frac{1}{4\aroot}}}$}
          \\[\rowsep]
          
          \Centerstack{\maternk$(\matone, \mattwo)$\\
        $\matone >d$} 
        & \Centerstack{
        $\min(n^{-\quarter},\frac{ (\log n)^{\frac{d+1}{2}}} {n^{{(\matone\!-\!d)}\!/\!{(2\matone\!-\!d)}}})$
        }

        & \Centerstack{${\highlight{\frac{(\log n)^{\frac{d+1}{2}} \sqrt{\cn}}{\sqrt{n}}}}$}

        & \Centerstack{${\highlight{\frac{(\log n)^{\frac{d+1}{2}} \sqrt{\cn}}{\sqrt{n}}}}$}
          \\[\rowsep]
          
         \Centerstack{\imq$(\matone, \mattwo)$\\
         $\matone<\frac d2$} 
        & \Centerstack{$\highlight{\frac{(\log n)^{d+1}}{\sqrt{n}}}$}
        & \Centerstack{$n^{-\quarter}$}
        & \Centerstack{$\highlight{\frac{(\log n)^{d+1}}{\sqrt{n}}}$}
          \\[\rowsep]
          
          \Centerstack{\imq$(\matone, \mattwo)$\\ $\matone\geq \frac d2$ } 
    
        & \Centerstack{${\frac{(\log n)^{d+1}}{\sqrt{n}}}$}
        & \Centerstack{${\highlight{\frac{(\log n)^{\frac{d+1}{2}} \sqrt{\cn}}{\sqrt{n}}}}$}
        & \Centerstack{$\highlight{\frac{(\log n)^{\frac{d+1}{2}} \sqrt{\cn}}{\sqrt{n}}}$}
          \\[\rowsep]
    
        \Centerstack{\sinc$(\sincparam)$} 
        & \Centerstack{$\highlight{\frac{(\log n)^{d/2+1}}{\sqrt n}}$}
        & \Centerstack{$n^{-\quarter}$}
        & \Centerstack{$\highlight{\frac{(\log n)^{d/2+1}}{\sqrt n}}$}
          \\[\rowsep]

          \Centerstack{
         $\bspline(2\splineparam+1, \mattwo)$\\
         $\splineparam\in2\natural$}
        & \Centerstack{$\min(n^{-\quarter},e_{n, d, \beta})$}
        & \Centerstack{\na}
        & \Centerstack{$\highlight{ \min(e_{n, d, \beta},  (\frac{\log n}{n})^{\frac{\beta+1}{2\beta+4}})}$}
         \\[\rowsep]
         
        \Centerstack{
         $\bspline(2\splineparam+1,\mattwo)$\\
         $\splineparam\in2\natural_0+1$}
        & \Centerstack{$\min(n^{-\quarter},e_{n, d, \beta} )$}
        & $\highlight{\sqrt{\frac{\log n}{n}}}$
        & $\highlight{\sqrt{\frac{\log n}{n}}}$
        \\[1ex] \bottomrule \hline
    \end{tabular}
    }
    }
 \noindent\caption{
    \tbf{$\mbi{\mmd_{\kernel}(\inputcoreset,\ktcoreset)}$ guarantees for commonly used kernels.}  For $n$ input and $\sqrt{n}$ output points, we report the MMD bounds of \cref{theorem:mmd_guarantee} for \tkt,
    of \citet[Thm.~1]{dwivedi2021kernel} for \rkt, and of \cref{theorem:mmd_guarantee_kt_plus} for KT+ (with $\aroot\! >\! \frac{d}{d+1}$ for \laplace, $\aroot \!>\! \frac{d}{2\matone}$ for \maternk, $\aroot\!=\!\frac{\beta+2}{2\beta+2}$ for \bspline\ with even $\beta$, and $\aroot\!=\!\half$ for all other kernels). We assume a \subgauss\ $\P$ for the \gauss\ kernel, a \bndcase\ $\P$ for the \bspline\  kernel, and a \subexp\ $\P$ for all other $\kernel$ (see \cref{table:mmd_rates} for a definition of each $\P$ class). Here, $c_n \defeq \log\log n$, 
    $e_{n, d, \beta}\defeq \frac{\sqrt{\log n}}{n^{(2\beta-d)/4\beta}}$, 
    $\delta_i = \frac{\delta}{n}$, $\delta'=\frac{\delta}{2}$, and error is reported up to constants depending on $(\kernel, d, \delta,\aroot)$. 
    The \tkt guarantee for \maternk\ with  $\matone>3d/2$ assumes $\matone\!-\!d/2 \in \natural$ to simplify the presentation (see \cref{eq:matern_k_explicit} for the general case).
    The best rate is highlighted in \tbf{\blue{blue}}. 
    See \cref{proof_of_table_explicit_mmd} for  details of the derivation.
    }
     \label{table:explicit_mmd}
\end{table}

\section{Kernel Thinning+}
\label{sec:kt_plus}
We next introduce and analyze two new generalized KT variants: (i) \akt which leverages a {power kernel} $\kroot$ that interpolates between $\kernel$ and $\ksqrt$ to improve upon the MMD guarantees of target KT even when $\ksqrt$ is not available and (ii) KT+ which uses a sum of $\kernel$ and $\kroot$ to retain both the improved MMD guarantee of $\kroot$ and the superior single function guarantees of $\kernel$.

\para{Power kernel thinning}
\label{sub:power_kt}
First, we generalize the square-root kernel \cref{eq:sqrt-kernel} definition for shift-invariant $\kernel$ using the order $0$ generalized Fourier transform~\citep[GFT,][Def.~8.9]{wendland2004scattered}  $\what{f}$ of $\fun:\Rd\to\R$.
\begin{definition}[$\aroot$-power kernel]
\label{def:root}
Define $\kroot[1] \defeq \kernel$.
We say a kernel $\kroot[\half]$ is a \emph{$\half$-power kernel} for $\kernel$ if $\kernel(x,y) = (2\pi)^{-d/2} \int_{\Rd} \kroot[\half](x, z) \kroot[\half](z, y) dz$.
For $\alpha\in(\half,1)$, a kernel $\kroot(x, y)\! = \!\kappaaroot(x\!-\!y)$  on $\reals^d$ is an \emph{$\aroot$-power kernel} for  $\kernel(x, y)\! =\! \kappa(x\!-\!y)$ if $\what{\kappaaroot} = \what{\kappa}^{\aroot}$.
\end{definition}

By design,  
$\kroot[\half]$ matches $\ksqrt$ \cref{eq:sqrt-kernel} up to an immaterial constant rescaling.
Given a power kernel $\kroot$ we define \akt as generalized KT with $\kersplit = \kroot$.
Our next result (with proof in \cref{proof_of_aroot}) provides an MMD guarantee for \akt.
\newcommand{\aktcoreset}[1][\aroot]{\cset_{#1\mrm{KT}}}
\newcommand{\powerktresultname}{MMD guarantee for \akt}
\begin{theorem}[\tbf{\powerktresultname}]
\label{thm:a_root_kt}
Consider generalized KT (\cref{algo:kernel_thinning}) with $\kersplit=\kroot$ for some $\alpha \in [\half, 1]$, oblivious sequences $\inputcoreset$ and $(\delta_i)_{i=1}^{\floor{n/2}}$, and $\delta^\star \defeq \min_i\delta_i$. If $\textfrac{n}{2^{\m}}\!\in\! \natural$, then for any $\delta'\! \in\! (0, 1)$, the output coreset $\ktcoreset$ is of size $\textfrac{n}{2^{\m}}$ and satisfies 
    \begin{talign} 
	\mmd_{\kernel}(\inputcoreset, \ktcoreset)
    &\leq \parenth{\frac{2^m}{n} \sinfnorm{\kroot}}^{\frac{1}{2\aroot}} (2\cdot\wtil{\err}_{\aroot})^{1-\frac{1}{2\aroot}} 
    \big({2
    \!+ \!
    \sqrt{\frac{(4\pi)^{d/2}}{\Gamma(\frac{d}{2}\!+\!1)}}
    \!\cdot \rmin_{\max}^{\frac{d}{2}} \cdot  \wtil{\err}_{\aroot}   }\big)^{\frac1\aroot-1},
    \label{eq:mmd_guarantee_aroot_kt}
    \end{talign}
with probability at least $\psg$ (defined in  \cref{theorem:single_function_guarantee}).
The parameters $\wtil{\err}_{\aroot}$ and $\rmin_{\max}$ are defined in \cref{proof_of_aroot} and satisfy  $\wtil{\err}_{\aroot} = \order_{d}(\sqrt{\log n})$ and $\rmin_{\max} = \order_{d}(1)$ for compactly supported $\P$ and $\kroot$   and $\wtil{\err}_{\aroot} = \order_{d}(\sqrt{\log n \log\log n})$ and $\rmin_{\max} = \order_{d}(\log n)$ for subexponential $\P$ and $\kroot$, when $\delta^\star = \frac{\delta'}{n}$.
\end{theorem}
\cref{thm:a_root_kt} reproduces the \rkt guarantee of \citet[Thm.~1]{dwivedi2021kernel} when  $\aroot=\half$ and more generally accommodates any power kernel via an MMD interpolation result (\cref{mmd_sandwich}) that may be of independent interest. 
This generalization is especially valuable for less-smooth kernels like \laplace\ and \maternk$(\matone, \gamma)$ with $\matone \in (\frac d2, d]$ that have no square-root kernel. 
Our \tkt MMD guarantees are no better than \iid for these kernels, but, as shown in \cref{proof_of_table_explicit_mmd}, these kernels have \maternk\ kernels as $\aroot$-power kernels, which yield $o(n^{-\quarter})$ MMD in conjunction with \cref{thm:a_root_kt}.

\newcommand{\mmdkt}{\overline{\mbf{M}}_{\mrm{targetKT}}(\kernel)}
\newcommand{\mmdakt}[1][\kroot]{\overline{\mbf{M}}_{\mrm{powerKT}}(#1)}
\para{Kernel thinning+} %
\label{sub:ktplus_bounds}
Our final KT variant, \emph{kernel thinning+}, runs \ktsplit with a scaled sum of the target and power kernels, $\kplus \defeq {\kernel}/{\infnorm{\kernel}} + {\kroot}/{\infnorm{\kroot}}$.\footnote{When $\inputcoreset$ is known in advance, one can alternatively choose $\kplus \defeq {\kernel}/{\kinfsin[\kernel]} + {\kroot}/{\kinfsin[\kroot]}$.}
Remarkably, this choice simultaneously provides the improved MMD guarantees of \cref{thm:a_root_kt} and the dimension-free single function guarantees of \cref{theorem:single_function_guarantee} (see \cref{sec:proof_of_ktplus} for the proof).

\newcommand{\ktplusresultname}{Single function \& MMD guarantees for KT+}
\begin{theorem}[\tbf{\ktplusresultname}]
\label{theorem:mmd_guarantee_kt_plus}
Consider generalized KT (\cref{algo:kernel_thinning})  with $\kersplit=\kplus$, oblivious $\inputcoreset$ and $(\delta_i)_{i=1}^{\floor{n/2}}$, $\delta^\star \defeq \min_i\delta_i$, and $\textfrac{n}{2^{\m}}\!\in\! \natural$.
    For any fixed function $\fun \in \rkhs$, 
    index $\l \in [2^\m]$, and scalar $\delta'\! \in\! (0, 1)$, the \ktsplit coreset $\coreset[\m, \ell]$ satisfies
    \begin{talign}
          |{\pin\fun - \psplit^{(\ell)}\fun}|
      \leq 
      \frac{2^\m}{n}
      \cdot 
    	\sqrt{\frac{16}{3}\log(\frac{6m}{2^m\delta^\star}) \log(\frac{2}{\delta'})}
    	\norm{\fun}_{\kernel}  \sqrt{\sinfnorm{\kernel}},
    \label{eq:single_function_guarantee_ktplus}
    \end{talign}
    with probability at least  $\psg$ (for $\psg$ and $\psplit^{(\ell)}$ defined in \cref{theorem:single_function_guarantee}).
    Moreover, 
    \begin{talign} 
	\mmd_{\kernel}(\inputcoreset, \ktcoreset)
	\leq 
	\min\big[ 
	\sqrt{2} \cdot \mmdkt, 
	\quad 
	 2^{\frac{1}{2\aroot}} \cdot \mmdakt\big]
    \label{eq:mmd_guarantee_kt_plus}
\end{talign}
with probability at least  $\psg$, where $\mmdkt$ denotes the right hand side of \cref{eq:mmd_guarantee} with $\kinfsin$ replaced by $\sinfnorm{\kernel}$, and $\mmdakt$ denotes the right hand side of~\cref{eq:mmd_guarantee_aroot_kt}. %
\end{theorem}

As shown in \cref{table:explicit_mmd}, KT+ provides better-than-\iid MMD guarantees for every kernel in \cref{tab:kernels}---even the Laplace, non-smooth \Matern, and odd B-spline kernels neglected by prior analyses---while matching or improving upon the guarantees of \tkt and \rkt in each case.

\section{Experiments}
\label{sec:experiments}

\begin{figure}[tb!]
    \centering
    \begin{tabular}{c}
    \includegraphics[width=\linewidth]{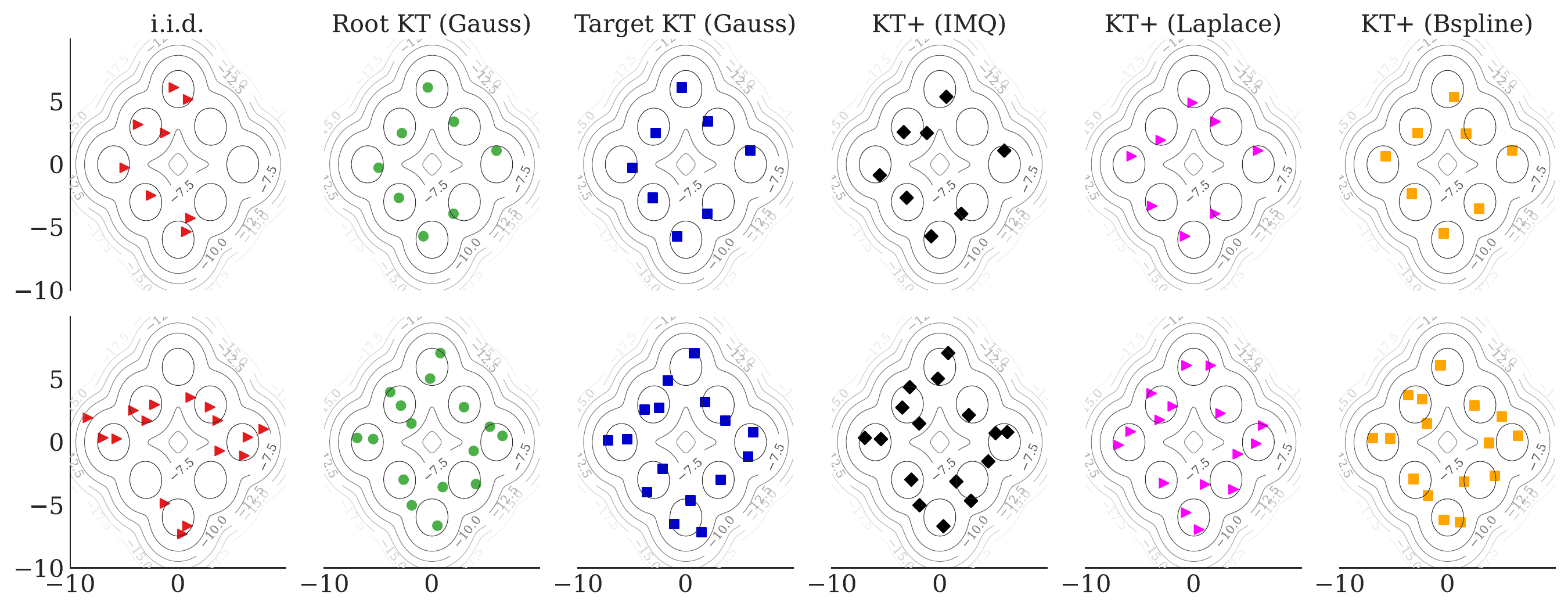}\\
    \hline
    \includegraphics[width=\linewidth]{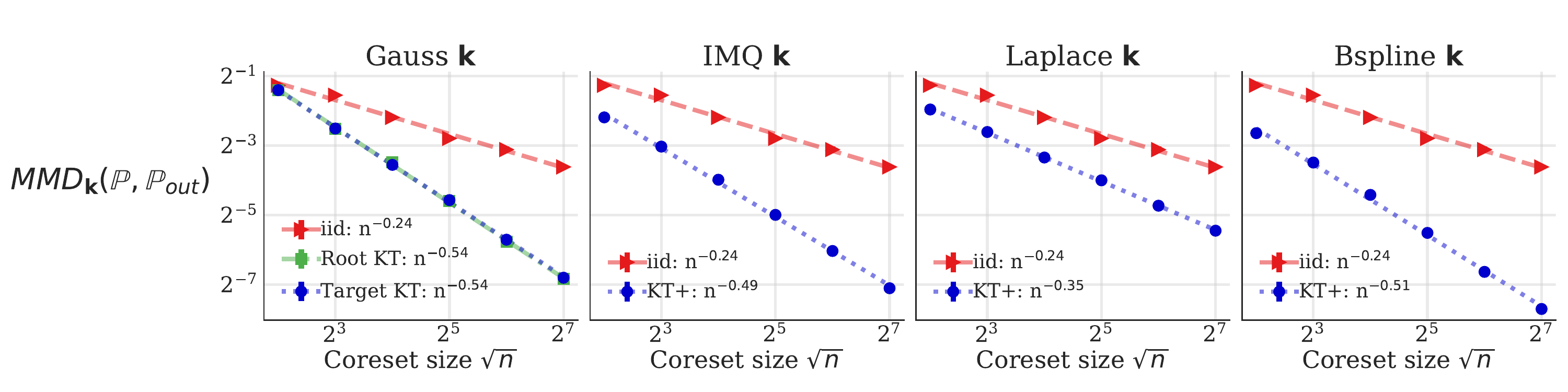}
    \end{tabular}
    \caption{\tbf{Generalized kernel thinning (KT) vs \iid sampling} for an 8-component mixture of Gaussians target $\P$. For kernels $\kernel$ without fast-decaying square-roots, KT+ offers visible and quantifiable improvements over \iid sampling. For Gaussian $\kernel$,  \tkt closely mimics \rkt. %
    }
    \label{fig:mog_scatter}
\end{figure}

\citet{dwivedi2021kernel} illustrated the MMD benefits of \rkt over \iid sampling and standard MCMC thinning with a series of vignettes focused on the Gaussian kernel.
We revisit those vignettes with the broader range of kernels covered by generalized KT and demonstrate significant improvements in both MMD and single-function integration error.
We focus on coresets of size $\sqrt{n}$ produced from $n$ inputs with $\delta_i\!=\!\frac{1}{2n}$, let $\P_{\mrm{out}}$ denote the empirical distribution of each output coreset, and report mean error ($\pm 1$ standard error) over $10$ independent replicates of each experiment.

\para{Target distributions and kernel bandwidths}
We consider three classes of target distributions on $\reals^d$: (i) mixture of Gaussians $\P = \frac{1}{M}\sum_{j=1}^{M}\mc{N}(\mu_j, \mbf{I}_2)$ with $M$ component means $\mu_j\in\reals^2$ defined in \cref{sec:vignettes_supplement}, (ii) Gaussian $\P = \mathcal {N}(0, \mbf{I}_d)$, %
and (iii) the posteriors of four distinct coupled ordinary differential equation models: the \emph{\citet{goodwin1965oscillatory} model} of oscillatory enzymatic control ($d=4$), the \emph{\citet{lotka1925elements} model} of oscillatory predator-prey evolution ($d=4$), the  \emph{\citet{hinch2004simplified} model} of calcium signalling in cardiac cells ($d=38$), and a tempered Hinch posterior.
For settings (i) and (ii), we use an \iid input sequence $\inputcoreset$ from $\P$ and kernel bandwidths   $\gaussparam={1}{/\mattwo}=\sqrt{2d}$. 
For setting (iii), we use MCMC input sequences $\inputcoreset$ from 12 posterior inference experiments of \citet{riabiz2020optimal} and set the bandwidths 
$\gaussparam={1}{/\mattwo}$ as specified by \citet[Sec. K.2]{dwivedi2021kernel}.

\begin{figure}[htb!]
    \centering
    \includegraphics[width=\linewidth]{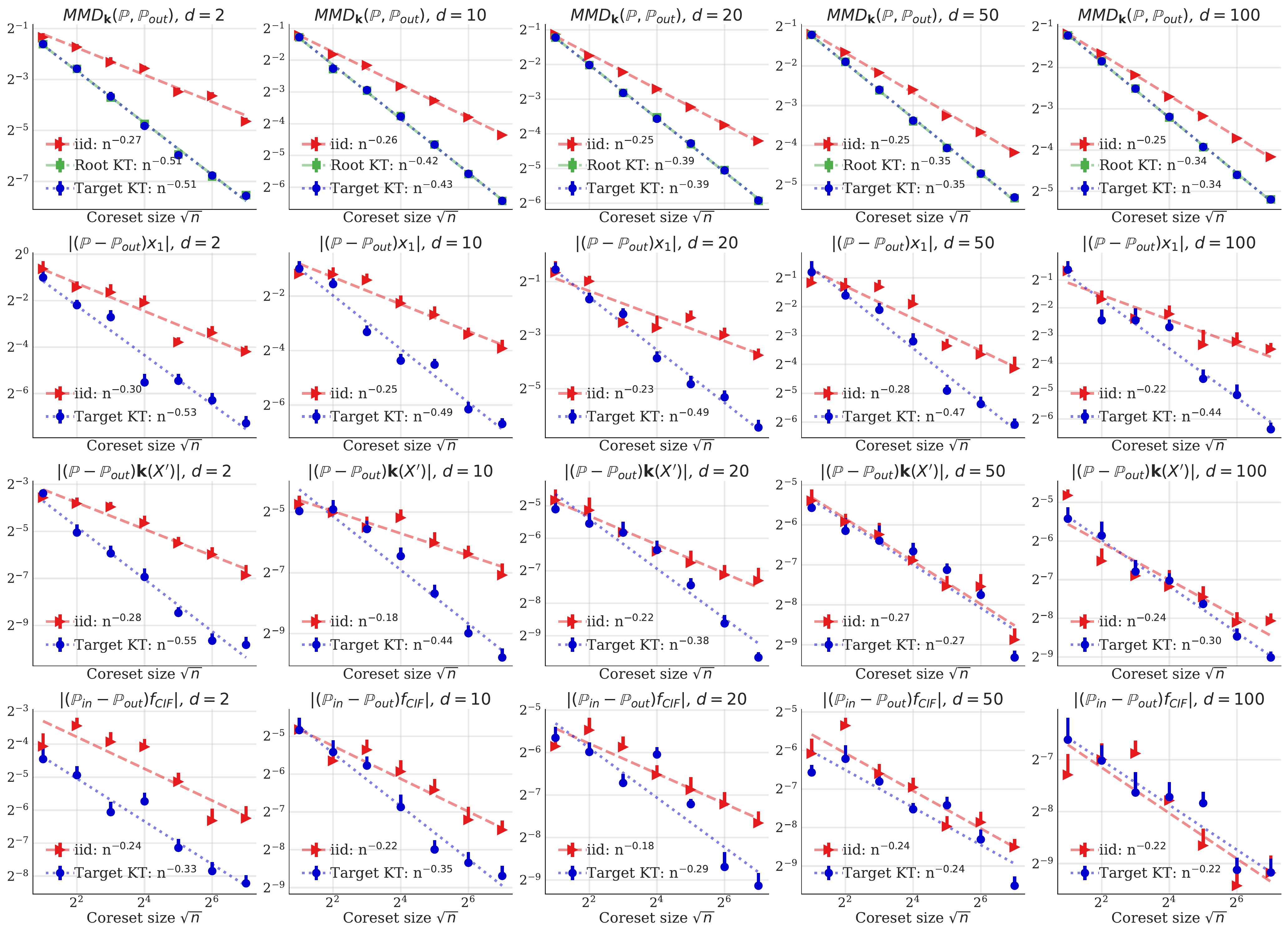}
    \caption{\tbf{MMD and single-function integration error for Gaussian $\kernel$ and standard Gaussian $\P$ in $\reals^d$.} Without using a square-root kernel, \tkt matches the MMD performance of \rkt and improves upon \iid MMD and single-function integration error, even in $d=100$ dimensions.}
    \label{fig:gauss}
\end{figure}

\para{Function testbed}
To evaluate the ability of generalized KT to improve integration both inside and outside of $\rkhs$, we evaluate integration error for (a) a random element of the target kernel RKHS ($f(x) = \kernel(X', x)$ described in \cref{sec:vignettes_supplement}), (b) moments ($f(x) = x_1$ and $f(x)=x_1^2$), and (c) a standard numerical integration benchmark test function from the \emph{continuous integrand family} \citep[CIF,][]{genz1984testing}, $f_{\mrm{CIF}}(x) =  \exp(-\frac{1}{d} \sum_{j=1}^d \abss{x_j-u_j})$ for $u_j$ drawn \iid and uniformly from $[0, 1]$. 

\para{Generalized KT coresets} 
For an $8$-component mixture of Gaussians target $\P$, the top row of \cref{fig:mog_scatter} highlights the visual differences between \iid coresets and coresets generated using generalized KT.
We consider \rkt with \gauss\ $\kernel$, \tkt with \gauss\ $\kernel$, and KT+ ($\alpha=0.7$) with \laplace\ $\kernel$, KT+ ($\alpha=\half$) with \imq\ $\kernel$ ($\nu=0.5$), and KT+($\aroot = \frac23$) with \bspline(5) $\kernel$, and note that the \bspline(5) (i.e., $\beta=2$) and \laplace\ $\kernel$ do not admit square-root kernels. In each case, even for small $n$, generalized KT provides a more even distribution of points across components with fewer within-component gaps and clumps.
Moreover, as suggested by our theory, \tkt and \rkt coresets for \gauss\ $\kernel$ have similar quality despite \tkt making no explicit use of a square-root kernel. The MMD error plots in the bottom row of \cref{fig:mog_scatter} provide a similar conclusion quantitatively, where we observe that for both variants of KT, the MMD error decays as $n^{-\half}$, a significant improvement over the $n^{-\quarter}$ rate of \iid\ sampling.
We also observe that the majority of the empirical MMD decay rates are in close agreement with the rates guaranteed by our theory in \cref{table:explicit_mmd} ($n^{-\half}$ for \gauss\ and \imq\ and $n^{-\frac1{4\aroot}} = n^{-0.36}$ for \laplace).
We provide additional visualizations and results in \cref{fig:scatter_plot_additional,fig:mog_mmd} of \cref{sec:vignettes_supplement}, including MMD errors for $M=4$ and $M= 6$ component mixture targets.
The conclusions remain consistent with those drawn from \cref{fig:mog_scatter}.

\para{\Tkt vs.\ \iid sampling} 
For Gaussian $\P$ and Gaussian $\kernel$, \cref{fig:gauss} quantifies the improvements in distributional approximation obtained when using \tkt in place of a more typical \iid summary.
Remarkably, \tkt significantly improves the rate of decay and order of magnitude of mean $\mmd_{\kernel}(\P,\pout)$, even in $d=100$ dimensions with as few as $4$ output points.
Moreover, in line with our theory, \tkt MMD closely tracks that of \rkt without using $\ksqrt$. 
Finally, \tkt delivers improved single-function integration error, both of functions in the RKHS (like $\kernel(X',\cdot)$) and those outside (like the first moment and CIF benchmark function), even with large $d$ and relatively small sample sizes.

\begin{figure}[htb!]
    \centering
    \includegraphics[width=\linewidth]{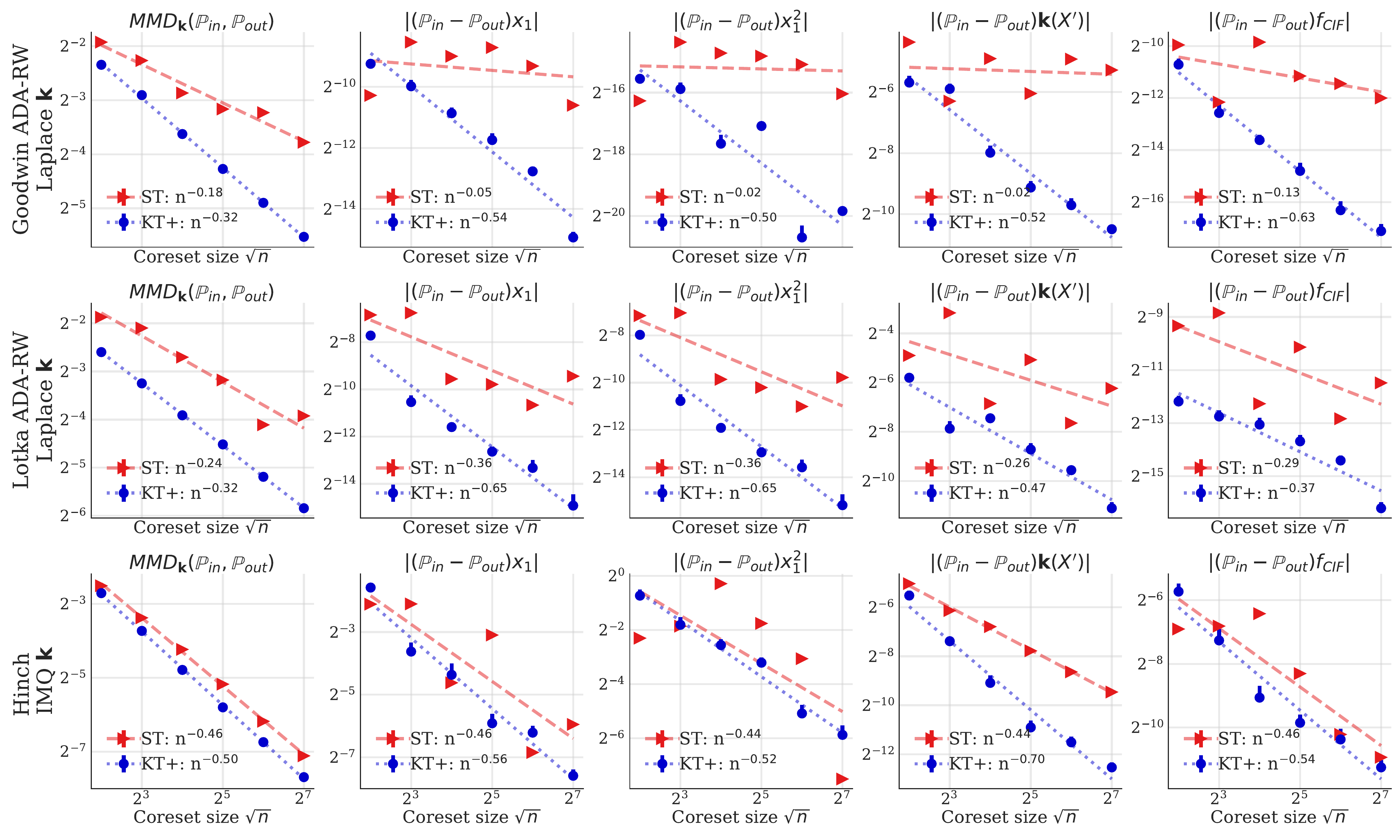}
    \caption{\tbf{Kernel thinning+ (KT+) vs. standard MCMC thinning (ST).} For kernels without fast-decaying square-roots, KT+ improves MMD and integration error decay rates in each posterior inference task.}
    \label{fig:mcmc}
\end{figure}

\para{KT+ vs.\ standard MCMC thinning} For the MCMC targets, we measure error with respect to the input distribution $\pin$ (consistent with our guarantees), as exact integration under each posterior $\P$ is intractable. We employ KT+ ($\alpha = 0.81$) with  \laplace\ $\kernel$ for  Goodwin and Lotka-Volterra and KT+ ($\alpha = 0.5$) with \imq\ $\kernel$ ($\nu=0.5$) for Hinch. Notably, neither kernel has a square-root with fast-decaying tails.
 In \cref{fig:mcmc}, we evaluate thinning results from one chain targeting each of the Goodwin, Lotka-Volterra, and Hinch posteriors and observe that  KT+ uniformly improves upon the MMD error of standard thinning (ST), even when ST exhibits better-than-\iid accuracy. Furthermore, KT+ provides significantly smaller integration error for functions inside of the RKHS (like $\kernel(X',\cdot)$) and outside of the RKHS (like the first and second moments and the benchmark CIF function) in nearly every setting. See \cref{fig:mcmc_supplement} of \cref{sec:vignettes_supplement} for plots of the other 9 MCMC settings.

\section{Discussion and Conclusions}

In this work, we introduced three new generalizations of the \rkt algorithm of \citet{dwivedi2021kernel}
with broader applicability and strengthened guarantees for generating compact representations of a probability distribution.
\textsc{Target} \ktsplit provides $\sqrt{n}$-point summaries with $\order(\sqrt{\log n / n})$ integration error guarantees for any kernel, any target distribution, and any function in the RKHS; 
\akt yields improved better-than-\iid MMD guarantees even when a square-root kernel is unavailable; and KT+ simultaneously inherits the guarantees of \tkt and \akt.
While we have focused on unweighted coreset quality we highlight that the same MMD guarantees extend to any improved reweighting of the coreset points. 
For example, for downstream tasks that support weights, 
one can optimally reweight $\pout$ to approximate $\pin$ in $\order(n^{\frac32})$ time by directly minimizing $\mmd_{\kernel}$. 
Finally, one can combine generalized KT with the \compresspp meta-algorithm of \citet{shetty2022distribution} to obtain coresets of comparable quality in near-linear time.  
\clearpage\newpage
\subsubsection*{Reproducibility Statement}
{See \cref{sec:vignettes_supplement}
for supplementary experimental details and results
and the \texttt{goodpoints} Python package 
\begin{center}
    \url{https://github.com/microsoft/goodpoints}
\end{center}
for Python code reproducing all experiments.}
{\subsubsection*{Acknowledgments}
We thank Lucas Janson and Boaz Barak for their valuable feedback on this work. RD acknowledges the support by the National Science Foundation under Grant No. DMS-2023528 for the Foundations of Data Science Institute (FODSI).}

\bibliographystyle{iclr2022_conference}
{\small \bibliography{refs}}
\appendix
\etoctocstyle{1}{Appendix}
\etocdepthtag.toc{mtappendix}
\etocsettagdepth{mtchapter}{none}
\etocsettagdepth{mtappendix}{section}
\newpage
{\small\tableofcontents}

\section{Details of \ktsplit and \ktswap}
\label{sec:algo}

\setcounter{algocf}{0}
\renewcommand{\thealgocf}{\arabic{algocf}a}
\begin{algorithm2e}[H]
  \SetKwFunction{proctwo}{\texttt{get\_swap\_params}}
\caption{{\large\textsc{kt-split}\ --\ } Divide points into candidate coresets of size $\floor{n/2^\m}$} 
  \label{algo:ktsplit}
  \SetAlgoLined\DontPrintSemicolon
  \small
  {
  \KwIn{kernel $\kersplit$, point sequence $\inputcoreset = (\axi[i])_{i = 1}^n$, thinning parameter $\m \in \natural$, probabilities $(\delta_i)_{i = 1}^{\floor{\frac{n}{2}}}$ 
  }
  \BlankLine
  {$\coreset[j,\ell] \gets \braces{}$ 
  for $0\leq j\leq \m$ and $1\leq \ell \leq 2^j$} 
  \quad// Empty coresets: $\coreset[j,\ell]$ 
   has size $\floor{\frac{i}{2^j}}$ 
  after round $i$\\ 
  \BlankLine
  {$\sgparam[j,\ell] \gets 0$ 
  for $1\leq j\leq \m$ and $1\leq \ell \leq 2^{j-1}$}   
 \quad // Swapping parameters
\\
  \BlankLine
  \For{$i=1, \ldots, \floor{n/2}$}
	  {
        $\coreset[0,1]\texttt{.append}(\x_i); \coreset[0,1]\texttt{.append}(\x_{2i})$ \\[2pt]
        // Every $2^j$ rounds, add one point from parent coreset $\coreset[j\!-\!1,\ell]$
			to each child coreset $\coreset[j,2\ell\!-\!1]$,  $\coreset[j,2\ell]$\\[1pt]
		\For{\textup{($j = 1;
		\ j \leq m\ \textbf{and}\ i / 2^{j-1} \in \natural;
		\ j = j + 1$)}}
			{
			\For{$\ell=1, \ldots, 2^{j-1}$}
			{
			$(\mc{S},\mc{S}') \gets (\coreset[j-1,\ell], \coreset[j,2\ell-1])$;
			\quad
			$(\x, \tilde{\x}) 
			    \gets
			   \texttt{get\_last\_two\_points}(\mc{S})$\\[2pt]
            // Compute swapping threshold $\cnew[]$\\[1pt] %
			$ \cnew[], \sigma_{j,\l} \gets $\texttt{get\_swap\_params}($\sigma_{j, \l}, \vmax[], \delta_{|\cset|/2} \frac{2^{j-1}}{m}$\!)
     for $\vmax[]^2\! \!
      =\! \kersplit(\x,\x)\!+\!\kersplit(\tilde{\x},\tilde{\x})\!-\!2\kersplit(x,\tilde{\x})$\\[2pt]
      
            // Assign one point to each child after probabilistic swapping\\[1pt]
$\wvprod[]\gets \kersplit(\tilde{\x},\tilde{\x})-\kersplit(\x, \x)
		  +\Sigma_{\y\in\mc{S}}(\kersplit
			    (\y, \x)-\kersplit(\y,\tilde{\x})) 
			     - 2\Sigma_{\z\in\mc{S}'}(\kersplit(\z, \x)-\kersplit(\z,\tilde{\x}))$
			 \\[3pt]
			 
		    $(x, \tilde{\x}) \gets (\tilde{\x}, x)$ \textit{ with probability }
		    $\min(1, \half (1-\frac{\wvprod[]}{\cnew[]})_+)$
		    \\[2pt]
          $\coreset[j,2\ell-1]\texttt{.append}(\x); 
		        \quad \coreset[j,2\ell]\texttt{.append}(\tilde{\x})$
			}
			}
	  }
    \KwRet{$\ksplitcoresets$\textup{, candidate coresets of size $\floor{n/2^\m}$}}\\
    \hrulefill\\
    \SetKwProg{myproc}{function}{}{}
     \myproc{\proctwo{$\sigma, \vmax[], \delta$}:}{
     $
			\cnew[] 
			    \gets \max(\vmax[] \sigma\sqrt{\smash[b]{2\log(2/\delta)}}, \vmax[]^2)$ \\
     $\sigma^2 \gets \sigma^2
	        \!+\! \vmax[]^2(1 \!+\! ({\vmax[]^2}{}\! - \!2\cnew[]){\sigma^2}{/\cnew[]^2})_+$\\
     }
     \KwRet{$(\cnew[], \sigma)$}\;
  }
\end{algorithm2e}

\setcounter{algocf}{0}
\renewcommand{\thealgocf}{\arabic{algocf}b}
\begin{algorithm2e}[ht!]
\caption{\large\textsc{kt-swap}\ --\ \normalsize Identify and refine the best candidate coreset} 
  \label{algo:ktswap}
 \SetAlgoLined\DontPrintSemicolon
\footnotesize
{
    \KwIn{kernel $\kernel$, point sequence $\inputcoreset = (\axi[i])_{i = 1}^n$, candidate coresets $\ksplitcoresets$}
        \BlankLine
    $\coreset[\m,0] 
	    \gets \texttt{baseline\_thinning}(\inputcoreset, \texttt{size}=\floor{n/2^\m})
	 $  \quad// Compare to baseline (e.g., standard thinning)%
	 \BlankLine
	 $\ktcoreset \gets \coreset[\m, \ell^\star]
	 \text{ for }
	 \ell^\star 
	    \gets \argmin_{\ell \in \braces{0, 1, \ldots, 2^\m}} \mmd_{\kernel}(\inputcoreset, \coreset[\m,\ell])$ 
	    // \textup{Select best candidate coreset} \\
	    \BlankLine
	   // Swap out each point in $\ktcoreset$ for best alternative in $\inputcoreset$ \\[1pt]
	   \For{$i=1, \ldots, \floor{n/2^\m}$}{
	   \BlankLine
	    $\ktcoreset[i] \gets \argmin_{z\in\inputcoreset}\mmd_{\kernel}(\inputcoreset, \ktcoreset \text{ with } \ktcoreset[i] = z)$ 
	   }
    \KwRet{
	 $\ktcoreset$\textup{, refined coreset of size $\floor{n/2^\m}$}
	}
}
\end{algorithm2e}
\renewcommand{\thealgocf}{\arabic{algocf}}

\newcommand{\kdiff}{\wtil{\mc W}_{\m}}

\section{Proof of \cref{theorem:single_function_guarantee}: \singlefunctionguaranteeresultname}
\label{sub:proof_of_theorem:single_function_guarantee}

The proof is identical for each index $\ell$, so, without loss of generality, we prove the result for the case $\ell=1$. Define
\begin{talign}
\kdiff \defeq \kdiffgen[1] = \pin\kersplit - \pout^{(1)}\kersplit
= \frac1n\sum_{x\in\inputcoreset}\kersplit(x, \cdot)-
\frac1{n/2^{\m}}\sum_{x\in\coreset[\m, 1]}\kersplit(x, \cdot).
\end{talign}
Next, we use the results about an intermediate algorithm, kernel halving \citep[Alg.~3]{dwivedi2021kernel} that was introduced for the analysis of kernel thinning. Using the arguments from \citet[Sec.~5.2]{dwivedi2021kernel}, we conclude that \ktsplit with $\kersplit$ and thinning parameter $\m$, is equivalent to repeated kernel halving with kernel $\kersplit$ for $\m$ rounds (with no \texttt{Failure} in any rounds of kernel halving). 
On this event of equivalence, denoted by  $\eventequi$, \citet[Eqns.~(50, 51)]{dwivedi2021kernel} imply that the function $\kdiff \in \rkhs_{\trm{split}}$ is equal in distribution to another random function $\mc W_{m}$, where $\mc W_m$ is unconditionally  sub-Gaussian with parameter
\begin{talign}
\label{eq:sigma_m}
\sigma_{m}= \frac{2}{\sqrt{3}} \frac{2^\m}{n}\sqrt{\infnorm{\kersplit} \log(\frac{6m}{2^{m}\delta^\star})},
\end{talign}
that is,
\begin{talign}
\Exs[\exp(\langle{\mc W_m, \fun}\rangle_{\kersplit})] 
\leq \exp(\half \sigma_{m}^2\norm{\fun}_{\kersplit}^2)
\qtext{for all} \fun \in \rkhs_{\trm{split}},
\label{eq:wm_temp}
\end{talign}
where we note that the analysis of \cite{dwivedi2021kernel} remains unaffected when we replace $\infnorm{\kersplit}$ by $\kinfsin[\kersplit]$ in all the arguments.
Applying the sub-Gaussian Hoeffding inequality \citep[Prop.~2.5]{wainwright2019high} along with \cref{eq:wm_temp}, we obtain that 
\begin{talign}
\P[\abss{\langle{\mc W_m, \fun}\rangle_{\kersplit}}>t] 
\leq 2\exp(-\half t^2/( \sigma_{m}^2\norm{\fun}_{\kersplit}^2))
\leq \delta' \text{ for } t \defeq \sigma_{m}\norm{\fun}_{\kersplit} \sqrt{2\log(\frac{2}{\delta'})}.
\label{eq:fun_subgauss_bound}
\end{talign}
Call this event $\event[\textup{sg}]$.
As noted above, conditional to the event $\eventequi$, we also have
\begin{talign}
\label{eq:wm_wtilm}
\mc W_m \seq{d} \kdiff \implies 
\langle{\mc W_m, \fun}\rangle_{\kersplit} 
\seq{d} 
\pin\fun-\pout^{(1)}\fun,
\end{talign}
where $\seq{d}$ denotes equality in distribution. Furthermore, \citet[Eqn.~48]{dwivedi2021kernel} implies that
\begin{talign}
\label{eq:equievent}
\P(\eventequi) \geq 1\!-\!\sum_{j=1}^{\m}\frac{2^{j-1}}{m}\sum_{i=1}^{n/2^j}\delta_{i}.
\end{talign}
Putting the pieces together, we have
\begin{talign}
\P[|\pin\fun-\pout^{(1)}\fun| \leq t] 
\geq \P(\eventequi \cap \event[\textup{sg}]^c)
\geq \P(\eventequi) - \P(\event[\textup{sg}])
\geq 1\!-\!\sum_{j=1}^{\m}\frac{2^{j-1}}{m}\sum_{i=1}^{n/2^j}\delta_{i} \!-\! \delta' = \psg,
\label{eq:fun_subgauss_bound}
\end{talign}
as claimed. The proof is now complete.
\section{Proof of \cref{l2pn_guarantee}: Guarantees for functions outside of $\splitrkhs$}
\label{sec:proof_of_l2pn_guarantee}
Fix any index $\ell \in [2^\m]$, scalar $\delta'\! \in\! (0, 1)$, and $\fun$ defined on $\inputcoreset$, 
and consider the associated vector
$g\in\reals^n$ with $g_i = f(\x_i)$ for each $i\in[n]$.
We define two extended functions by appending the domain by $\real^n$ as follows: For any $w \in \real^n$, define $\fun_1((x,w)) = f(x)$ and $\fun_2((x,w)) = \angles{g,w}$ (the Euclidean inner product). Then we note that these functions replicate the values of $f$ on $\inputcoreset$, since $\fun_1((x_i,w)) = f(x_i)$ for arbitrary $w \in \real^{n}$, and $\fun_2((x_i,e_i)) = \angles{g,e_i} = g_i = f(x_i)$, where $e_i$ denotes the $i$-th basis vector in $\real^n$. Thus we can write
\begin{talign}
\label{eq:psplit_variants}
\pin\fun - \psplit^{(\ell)}\fun
=
\pin'\fun_1 - {\P'}^{(\ell)}_{\tgf{split}}\fun_1
=
\pin'\fun_2 - {\P'}^{(\ell)}_{\tgf{split}}\fun_2
\end{talign}
for the extended empirical distributions $\pin' = \frac{1}{n}\sum_{i=1}^n \delta_{\x_i, e_i}$ and ${\P'}^{(\ell)}_{\tgf{split}}$, defined analogously.
Notably, any function of the form $\tilde{f}((x,w)) = \inner{\tilde{g}}{w}$ belongs to the RKHS of $\kersplit'$ with \begin{talign}\label{eq:vector-rkhs-bound}
\snorm{\tilde{f}}_{\kersplit'}\leq \twonorm{\tilde{g}}
\end{talign}
by \citet[Thm.~5]{berlinet2011reproducing}.

By the repeated halving interpretation of kernel thinning \citep[Sec.~5.2]{dwivedi2021kernel}, on an event $\event$ of probability at least $\psg + \delta'$ we may write 
\begin{talign}
\pin'\fun_2 - {\P'}^{(\ell)}_{\tgf{split}}\fun_2
=
\sum_{j=1}^m  \dotsplitprime{\mcw_j,\fun_2} 
=
\sum_{j=1}^m  \dotsplitprime{\mcw_j,\fun_{2,j}} 
\end{talign}
where $\mcw_j$ denotes suitable random functions in the RKHS of $\kersplit'$, and each $\fun_{2,j}((x,w)) = \inner{g^{(j)}}{w}$ for $g^{(j)} \in \reals^n$ a sparsification of $g$ with at most $\frac{n}{2^{j-1}}$ non-zero entries that satisfy
\begin{talign}
\Exs[\exp(\langle{\mcw_j, \fun_{2,j}}\rangle_{\kersplit'})\mid \mcw_{j-1}] 
\leq \exp( \frac{\sigma_{j}^2}{2}\ksplitprimenorm{\fun_{2,j}}^2)
\sless{\cref{eq:vector-rkhs-bound}} \exp( \frac{\sigma_{j}^2}{2}\twonorm{g^{(j)}}^2)
\leq \exp( \frac{\sigma_{j}^2}{2}\frac{n}{2^{j-1}}\kinfsin[f]^2)
\end{talign}
for $\mcw_0\defeq0$
and
\begin{talign}
\sigma_j^2 = 4(\frac{2^{j-1}}{n})^2 \kinfsin[\kersplit']\log(\frac{4m}{2^j\delta^\star})
\leq 2 \cdot \frac{4^{j}}{n^2} \log(\frac{4m}{2^j\delta^\star}),
\end{talign}
since by definition $\kinfsin[\kersplit'] \leq 2$.
Hence, by sub-Gaussian additivity \citep[see, e.g.,][Lem.~8]{dwivedi2021kernel}, 
$\pin\fun_2 - {\P}^{(\ell)}_{\tgf{split}}\fun_2$ is $\wtil{\sigma}_2$ sub-Gaussian with 
\begin{talign}
\wtil{\sigma}_2^2 
    \leq \frac{4}{n} \kinfsin[f]^2 \cdot \sum_{j=1}^m 2^j \log(\frac{4m}{2^j\delta^\star})
    &\seq{(i)} \frac{4}{n} \kinfsin[f]^2 \cdot  2 \parenth{(2^{m}-1) \log(\frac{4m}{\delta^\star})
    - \parenth{(2^{m}-1)(m-1)+m} \log 2}\\
    &= \frac{4}{n} \kinfsin[f]^2 \cdot  2 \parenth{(2^{m}-1) \log(\frac{4m \cdot 2}{2^{m}\delta^\star})
    - m \log 2}\\
    &\leq 8 \cdot \frac{2^{m}}{n}  \kinfsin[f]^2 \cdot \log(\frac{8m}{2^{m}\delta^\star}), 
\end{talign}
i.e., 
\begin{talign}
\label{eq:sigma_2}
    \wtil{\sigma}_2 \leq \sqrt{\frac{2^{m}}{n}} \cdot \kinfsin[f] \cdot \sqrt{ 8 \log(\frac{8m}{2^{m}\delta^\star})}
\end{talign}
on the event $\event$, where step~(i) makes use of the following expressions:
\begin{talign}
    \sum_{j=1}^m 2^j = 2(2^{m}-1)
    \qtext{and}
    \sum_{j=1}^m j2^j = 2((m-1)(2^m-1)+ m).
\end{talign}

Moreover, when $f\in\splitrkhs$, we additionally have $f_1$ in the RKHS of $\kersplit'$ with \begin{talign}\label{eq:vector-rkhs-bound-2}
\snorm{f_1}_{\kersplit'}\leq \ksplitnorm{f}\sqrt{\infnorm{\kersplit}},
\end{talign}
as argued in the proof of \cref{eq:rkhs_scaling_norms}.
The proof of \cref{theorem:single_function_guarantee} then implies that 
$\pin'\fun_1 - {\P'}^{(\ell)}_{\tgf{split}}\fun_1$ is $\wtil{\sigma}_1$ sub-Gaussian with
\begin{talign}
\label{eq:sigma_1}
\wtil{\sigma}_1
    &\leq
    \ksplitprimenorm{f_a}
    \frac{2}{\sqrt{3}}\frac{2^m}{n}
    \sqrt{ \kinfsin[\kersplit'] \cdot \log(\frac{6m}{2^m\delta^\star})}
    \leq \frac{2^m}{n} \cdot \ksplitnorm{f} \sqrt{\infnorm{\kersplit}} \cdot 
    \sqrt{\frac83\log(\frac{6m}{2^m\delta^\star})},
\end{talign}
on the very same event $\event$.

Recalling \cref{eq:psplit_variants} and putting the pieces together with the definitions~\cref{eq:sigma_1,eq:sigma_2}, we conclude that on the event $\event$, the random variable $\pin\fun - \psplit^{(\ell)}\fun$ is $\wtil{\sigma}$ sub-Gaussian for
\begin{talign}
    \wtil{\sigma} \defeq \min(\wtil{\sigma}_1, \wtil{\sigma}_2)
    \sless{\cref{eq:sigma_2},\cref{eq:sigma_1}}
    \min\parenth{\sqrt{\frac{n}{2^{m}}} \kinfsin[\fun], \ksplitnorm{f} \sqrt{\infnorm{\kersplit}} } \cdot
    \frac{2^{m}}{n}\sqrt{8\log(\frac{8m}{2^m\delta^\star})}.
\end{talign}
The advertised high-probability bound~\cref{eq:l2pn_guarantee} now follows from the $\wtil{\sigma}$ sub-Gaussianity on $\event$ exactly as in the proof of \cref{theorem:single_function_guarantee}.
\section{Proof of \cref{theorem:mmd_guarantee}: \mmdguaranteeresultname}
\label{sec:proof_of_theorem_mmd_guarantee}
First, we note that by design, \ktswap ensures
\begin{talign}
     \mmd_{\kernel}(\inputcoreset, \ktcoreset) \leq 
     \mmd_{\kernel}(\inputcoreset, \coreset[\m, 1]),
     \label{eq:mmd_proof_step1}
\end{talign}
where $\coreset[\m, 1]$ denotes the first coreset returned by \ktsplit.
Thus it suffices to show that $\mmd_{\kernel}(\inputcoreset, \coreset[\m, 1])$ is bounded by the term stated on the right hand side of \cref{eq:mmd_guarantee}. Let $\pout^{(1)} \defeq \frac{1}{n/2^{\m}} \sum_{x\in\coreset[\m, 1]}\dirac_x$.
 By design of \ktsplit, $\supp{\poutone} \subseteq \supp{\pin}$. Recall the set $\set{A}$ is such that $\supp{\pin} \subseteq \set{A}$.

\paragraph{Proof of \cref{eq:mmd_guarantee}}
Let $\cover \defeq \kcover(\set{A})$ denote the cover of minimum cardinality satisfying \cref{eq:cover_ball}. Fix any $\fun\in\kball$.
By the triangle inequality and the covering property~\cref{eq:cover_ball} of $\cover$, we have 
\begin{talign}
\abss{(\pin-\poutone)\fun} 
&\leq 
\inf_{\funtwo\in\cover}\abss{(\pin-\poutone)(\fun-\funtwo)} + \abss{(\pin-\poutone)(\funtwo)} \\
&\leq 
\inf_{\funtwo\in\cover}\abss{\pin(\fun-\funtwo)}+\abss{\poutone(\fun-\funtwo)} + \sup_{\funtwo \in \cover}\abss{(\pin-\poutone)(\funtwo)} \\
&\leq 
\inf_{\funtwo\in\cover}2\sup_{x\in\set{A}}|\fun(\x)-\funtwo(\x)| + \sup_{\funtwo \in \cover}\abss{(\pin-\poutone)(\funtwo)} \\
&\leq 2\varepsilon  + \sup_{\funtwo\in\cover}\abss{(\pin-\poutone)(\funtwo)}.
\label{eq:mmd_inf_sup_decomposition}
\end{talign}
Applying \cref{theorem:single_function_guarantee},
we have
\begin{talign}
\label{eq:single_function_explicit_bound}
\abss{(\pin-\poutone)(\funtwo)}
    \leq 
\frac{2^\m}{n} \knorm{\funtwo} \sqrt{\frac83 \kinfsin \cdot \log(\frac{4}{\delta^\star}) \log(\frac{4}{\delta'})} 
\end{talign}
with probability at least $1\!-\!\delta'\!-\!\sum_{j=1}^{\m}\frac{2^{j-1}}{m} \sum_{i=1}^{n/2^j}\delta_i=\psg-\delta'$.
A standard union bound then yields that
\begin{talign}
\sup_{\funtwo\in\cover}\abss{(\pin-\poutone)(\funtwo)}
    \leq 
\frac{2^\m}{n} \sup_{\funtwo\in\cover}\knorm{\funtwo} \sqrt{\frac83 \kinfsin \cdot \log(\frac{4}{\delta^\star}) \brackets{\log |\cover| + \log(\frac{4}{\delta'})} }
\end{talign}
probability at least $\psg-\delta'$.
Since $f\in\kball$ was arbitrary, and $\cover \subset \kball$ and thus $\sup_{\funtwo\in\cover}\knorm{\funtwo} \leq 1$, we therefore have
\begin{talign}
\mmd_{\kernel}(\inputcoreset,\coreset[\m, 1]) 
&= \sup_{\knorm{\fun}\leq 1} \abss{(\pin\!-\!\poutone)\fun} 
\sless{\cref{eq:mmd_inf_sup_decomposition}}2\varepsilon \! +\! \sup_{\funtwo\in\cover}\abss{(\pin\!-\!\poutone)(\funtwo)} \\
&\leq2\varepsilon\!+\!
\sqrt{\frac{8\sinfnorm{\kernel}}{3}} \cdot  \frac{2^\m}{n}\sqrt{\log(\frac{4}{\delta^\star}) \brackets{\log |\cover| + \log(\frac{4}{\delta'})} },
\end{talign}
with probability at least $\psg-\delta'$ as claimed.

\newcommand{\asplitcoreset}{\cset_{\aroot}^{(\m, 1)}}
\newcommand{\pasplit}{\P_{\aroot}^{(\m, 1)}}

\section{Proof of \cref{thm:a_root_kt}: \powerktresultname}
\label{proof_of_aroot}
\paragraph{Definition of $\wtil{\err}_{\aroot}$ and $\rmin_{\max}$}
Define the $\kroot$ tail radii,
\begin{talign}
	\label{eq:tail_k_p}
	\rmin_{\kroot,n}^{\dagger}\!&\defeq\! \min\{r:\tail[\kroot](r)\! \leq\! \frac{\infnorm{\kroot}}{n}\},
	\qtext{where}
	\tail[\kroot](R) \defeq (\sup_{x} \int_{\ltwonorm{y}\geq R} \kroot^2(\x, \x-\y)d\y)^{\frac{1}{2}}, \\
    \rmin_{\kroot, n} 
    \!&\defeq\! \min\{r:\!
    \sup_{\ltwonorm{\x-\y}\geq r} \abss{\kroot(\x,\y)}\!\leq\! \frac{\infnorm{\kroot}}{n}\},  \quad
     \label{eq:rmin_k} 
\end{talign}
and the $\inputcoreset$ tail radii
\begin{talign}
 &\rminpn[\inputcoreset]\! \defeq\! \max_{\x\in \inputcoreset}\ltwonorm{\x},
    \qtext{and}
     \rminnew[\inputcoreset, \kroot, n] \!\defeq\! \min\big(\rminpn[\inputcoreset], n^{1+\frac1d}\rmin_{\kroot, n}+ n^{\frac1d} {\sinfnorm{\kroot}}/{\klip[\kroot]} \big).
     \label{eq:rmin_P}
\end{talign} 
Furthermore, define the inflation factor
    \begin{talign}
    \label{eq:err_simple_defn}  
    &\err_{\kroot}(n,\!m, d,\! \delta,\! \delta',\! R)
     \defeq
     37\sqrt{ \log\parenth{\frac{6m}{2^{m}\delta}}} \brackets{ \sqrt{\log\parenth{\frac{4}{\delta'}}} + 5 \sqrt{{d\log (2+2\frac{\klip[\kroot]}{\infnorm{\kroot}}\big(\rmin_{\kroot, n} + R\big) ) }} },
\end{talign}
where $\klip[\kroot]$ denotes a Lipschitz constant satisfying
$\abss{\kroot(\x, \y)-\kroot(\x, \z)} \leq \klip[\kroot] \ltwonorm{\y-\z}$ for all $x, y, z \in \Rd$. With the notations in place, we can define the quantities appearing in \cref{thm:a_root_kt}:
\begin{talign}
\label{eq:err_simplified}
\wtil{\err}_{\aroot} \defeq
\err_{\kroot}(n,\! m, d,\! \delta^\star, \!\delta',\!\rminnew[\inputcoreset, \kroot, n])
\qtext{and}
 \rmin_{\max} \defeq \max(\rminpn[\inputcoreset],\rmin_{\kroot, n/2^m}^\dagger).
\end{talign}
The scaling of these two parameters depends on the tail behavior of $\kroot$ and the growth of the radii $\rmin_{\inputcoreset}$ (which in turn would typically depend on the tail behavior of $\P$). The scaling of $\wtil{\err}_{\aroot}$ and $\rmin_{\max}$ stated in \cref{thm:a_root_kt} under the compactly supported or subexponential tail conditions follows directly from \citet[Tab.~2, App.~I]{dwivedi2021kernel}.

\paragraph{Proof of \cref{thm:a_root_kt}}
The \ktswap step ensures that
\begin{talign}
    \mmd_{\kernel}(\inputcoreset, \cset_{\aroot\mrm{KT}}) \leq
    \mmd_{\kernel}(\inputcoreset, \asplitcoreset),
\end{talign}
where $\asplitcoreset$ denotes the first coreset output by \ktsplit with $\kersplit=\kroot$.
Next, we state a key interpolation result for $\mmd_{\kernel}$ that relates it to the MMD of its power kernels~(\cref{def:root}) (see \cref{proof_of_mmd_sandwich} for the proof).
\newcommand{\mmdinterpolationresultname}{An interpolation result for MMD}
\begin{proposition}[\tbf{\mmdinterpolationresultname}]
\label{mmd_sandwich}
Consider a shift-invariant kernel $\kernel$ that admits valid $\aroot$ and $2\aroot$-power kernels $\kroot$ and $\kroot[2\aroot]$ respectively for some $\aroot \in [\frac12, 1]$. Then for any two discrete measures $\P$ and $\Q$ supported on finitely many points, we have
\begin{talign}
\label{eq:mmd_sandwich}
\mmd_{\kernel}(\P, \Q) \leq 
(\mmd_{\kroot}(\P, \Q))^{2-\frac{1}{\aroot}}
\cdot (\mmd_{\kroot[2\aroot]}(\P, \Q))^{\frac{1}{\aroot}-1}.
\end{talign}
\end{proposition}

Given \cref{mmd_sandwich}, it remains to establish suitable upper bounds on MMDs of $\kroot$ and $\kroot[2\aroot]$. To this end, first we note that for any reproducing kernel $\kernel$ and any two distributions $\P$ and $\Q$, \Holder's inequality implies that
\begin{talign}
\label{eq:mmd_holder}
    \mmd_{\kernel}^2(\P, \Q) = \knorm{(\P-\Q)\kernel}^2 = (\P-\Q)(\P-\Q)\kernel
    &\leq \norm{\P-\Q}_1 \sinfnorm{(\P-\Q)\kernel} \\
    &\leq 2  \sinfnorm{(\P-\Q)\kernel}.
\end{talign}
Now, let $\pin$ and $\pasplit$ denote the empirical distributions of $\inputcoreset$ and $\asplitcoreset$. We find that
\begin{talign}
\label{eq:linf_root_bound}
    \mmd_{\kroot}(\inputcoreset, \asplitcoreset)
    \sless{\cref{eq:mmd_holder}} \sqrt{2 \kinfsin[(\pin-\pasplit)\kroot]}
    \sless{(i)} \sqrt{2  \cdot \frac{2^{\m}}{n} \kinfsin[\kroot]  \wtil{\err}_{\kroot}}
\end{talign}
with probability $\psg-\delta'$, where $\wtil{\err}_{\kroot}$ was defined in \cref{eq:err_simplified}, and step~(i) follows from \citet[Thm.~4(b)]{dwivedi2021kernel}.
We note that while \citet[Thm.~4(b)]{dwivedi2021kernel} uses $\sinfnorm{\kroot}$ in their bounds, we can replace it by $\kinfsin[\kroot]$, and verifying that all the steps of the proof continue to be valid (noting that $\kinfsin[\kroot]$ is deterministic given $\inputcoreset$).
Putting \cref{eq:linf_root_bound}(i) together with \citet[Thm.~2]{dwivedi2021kernel} yields that
\begin{talign}
\label{eq:mmd_root_bound}
    \mmd_{\kroot[2\aroot]}(\inputcoreset, \asplitcoreset)
    \leq  \frac{2^{\m}}{n} \kinfsin[\kroot]
	\parenth{2\!+ \! \sqrt{\frac{(4\pi)^{d/2}}{\Gamma(\frac{d}{2}\!+\!1)}}
    \!\cdot \rmin_{\max}^{\frac{d}{2}} \cdot  \wtil{\err}_{\aroot}},
\end{talign}
with probability $\psg-\delta'$, where we have once again replaced the term $\sinfnorm{\kroot}$ with $\kinfsin[\kroot]$ for the same reasons as stated above.  We note that the two bounds~\cref{eq:linf_root_bound,eq:mmd_root_bound} apply under the same high probability event as noted in \citet[proof of Thm.~1, eqn.~(18)]{dwivedi2021kernel}. 
Putting together the pieces, we find that
\begin{talign}
        \mmd_{\kernel}(\inputcoreset, \asplitcoreset)
    &\  \ \sless{\cref{eq:mmd_sandwich}} 
    (\mmd_{\kroot}(\inputcoreset, \asplitcoreset)^{2-\frac{1}{\aroot}}
    \cdot (\mmd_{\kroot[2\aroot]}(\inputcoreset, \asplitcoreset))^{\frac{1}{\aroot}-1} \\
    &\sless{(\ref{eq:linf_root_bound},\ref{eq:mmd_root_bound})}
    \brackets{2 \cdot \frac{2^{\m}}{n} \kinfsin[\kroot]  \wtil{\err}_{\aroot} }^{1-\frac1{2\aroot}}
	\brackets{\frac{2^{\m}}{n} \kinfsin[\kroot]
	\parenth{2
    \!+ \!
    \sqrt{\frac{(4\pi)^{d/2}}{\Gamma(\frac{d}{2}\!+\!1)}}
    \!\cdot \rmin_{\max}^{\frac{d}{2}} \cdot  \wtil{\err}_{\aroot}   } }^{\frac1\aroot-1} \\
    & \ \ \ = \parenth{\frac{2^m}{n} \kinfsin[\kroot]}^{\frac{1}{2\aroot}} (2\cdot\wtil{\err}_{\aroot})^{1-\frac{1}{2\aroot}} 
    \parenth{2
    \!+ \!
    \sqrt{\frac{(4\pi)^{d/2}}{\Gamma(\frac{d}{2}\!+\!1)}}
    \!\cdot \rmin_{\max}^{\frac{d}{2}} \cdot  \wtil{\err}_{\aroot}   }^{\frac1\aroot-1},
\end{talign}
as claimed. The proof is now complete.

\section{Proof of \cref{theorem:mmd_guarantee_kt_plus}: \ktplusresultname}
\label{sec:proof_of_ktplus}
\begin{proofof}{\cref{eq:single_function_guarantee_ktplus}}
First, we note that the RKHS $\rkhs$ of $\kernel$ is contained in the RKHS $\rkhs^\dagger$ of $\kplus$ \citet[Thm.~5]{berlinet2011reproducing}.
Now, applying \cref{theorem:single_function_guarantee} with $\kersplit=\kplus$ for any fixed function $\fun\in\rkhs\subset\rkhs^\dagger$ and $\delta'\in(0,1)$, we obtain that
\begin{talign}
\label{eq:f_kplus}
      \abss{\pin\fun - \psplit^{(\ell)}\fun} 
  &\leq  \norm{\fun}_{\kplus}   \cdot 
  \frac{2}{\sqrt{3}} \frac{2^\m}{n}\sqrt{2\kinfsin[\kplus] \cdot \log(\frac{6m}{2^m\delta^\star})}
  \sqrt{ 2\log(\frac{2}{\delta'})}\\
  &\sless{(i)} \norm{\fun}_{\kplus}   \cdot 
   \frac{2^\m}{n}\sqrt{\frac{16}{3}\log(\frac{6m}{2^m\delta^\star})
  \log(\frac{2}{\delta'})},
    \\
&\sless{(ii)} \norm{\fun}_{\kernel}   \cdot 
  \frac{2^\m}{n}\sqrt{\frac{16}{3} \sinfnorm{\kernel} \log(\frac{6m}{2^m\delta^\star})
  \log(\frac{2}{\delta'})},    
    \end{talign}
with probability at least  $\psg$. Here step~(i) follows from the inequality $\sinfnorm{\kplus}\!\leq\! 2$, and step~(ii) follows from the inequality $\norm{\fun}_{\kplus} \!\leq\! \sqrt{\sinfnorm{\kernel}} \knorm{\fun}$, which in turn follows from the standard facts that 
\begin{talign}
\label{eq:rkhs_scaling_norms}
    \norm{\fun}_{\lambda \kernel} \seq{(iii)} \frac{\knorm{\fun}}{\sqrt{\lambda}},
    \qtext{and}
    \norm{\fun}_{\lambda\kernel + \kroot} 
    \sless{(iv)} \norm{\fun}_{\lambda\kernel}
    \qtext{for} \lambda>0, \fun \in \rkhs,
\end{talign}
see, e.g., \citet[Proof of Prop.~2.5]{zhang2013inclusion} for a proof of step~(iii), \citet[Thm.~5]{berlinet2011reproducing} for step~(iv).
The proof for the bound \cref{eq:single_function_guarantee_ktplus} is now complete.
\end{proofof}

\begin{proofof}{\cref{eq:mmd_guarantee_kt_plus}}
Repeating the proof of \cref{theorem:mmd_guarantee} with the bound~\cref{eq:single_function_explicit_bound} replaced by \cref{eq:single_function_guarantee_ktplus} yields that
\begin{talign}
    \mmd_{\kernel}(\inputcoreset, \ktpluscoreset) 
    &\leq
    \inf_{\vareps, \inputcoreset\subset \set A} 2\varepsilon \!+\! 
	\frac{2^\m}{n}
	\sqrt{\frac{16}{3}\sinfnorm{\kernel}  \log(\frac{6m}{2^m\delta^\star}) \cdot
	\brackets{ \log(\frac{4}{\delta'})\!+\!\entropy(\set{A}, \vareps)}}\\
	&\leq \sqrt 2 \cdot \mmdkt
	\label{eq:bound_1}
\end{talign}
with probability at least $\psg$. Let us denote this event by $\event[1]$.

\newcommand{\ktpsplitcoreset}{\cset_{\mrm{KT+}}^{(m, 1)}}
\newcommand{\pktpsplit}{\P_{\mrm{KT+}}^{(m, 1)}}
To establish the other bound, first we note that 
\ktswap step ensures that
\begin{talign}
\label{eq:mmd_ktplus_ktsplit}
    \mmd_{\kernel}(\inputcoreset, \ktpluscoreset) \leq
    \mmd_{\kernel}(\inputcoreset, \ktpsplitcoreset),
\end{talign}
where $\ktpsplitcoreset$ denotes the first coreset output by \ktsplit with $\kersplit=\kplus$. Thus for this case the suitable analog of the sub-Gaussian parameter (in \cref{eq:sigma_m}) is given by
\begin{talign}
\label{eq:sigma_kplus}
     \sigma_m = \frac{2}{\sqrt{3}} \frac{2^{m}}{n} \sqrt{\sinfnorm{\kplus} \log(\frac{6m}{2^m\delta^\star})}
     \qtext{where} \sinfnorm{\kplus} \leq 2.
\end{talign}
Next we note that $\kroot(x, \cdot)$ belongs to the RKHS of $\kplus$ with
\begin{talign}
\label{eq:kroot_kplus_norm}
     \Vert \kroot(x,\cdot) \Vert_{\kplus} \sless{\cref{eq:rkhs_scaling_norms}} \sqrt{\sinfnorm{\kroot}} \Vert \kroot(x,\cdot) \Vert_{\kroot}
     = \sqrt{\sinfnorm{\kroot}} \sqrt{\kroot(x, x)}
     \leq \sinfnorm{\kroot}.
\end{talign}
Now we are ready to adapt the arguments from \citet[Proof of Thm.~4]{dwivedi2021kernel} with $\sinfnorm{\kernel}$ by replacing $\sinfnorm{\kplus}$ (which in turn we bound by $2$) in \citet[Eqn.~35]{dwivedi2021kernel} due to \cref{eq:sigma_kplus}, and replacing $\kernel, \sinfnorm{\kernel}$ by $\kroot, \sinfnorm{\kroot}$ respectively in \citet[Lem.~(5, 6, 7)]{dwivedi2021kernel} due to \cref{eq:kroot_kplus_norm}. Overall these substitutions imply that we can repeat the proof of \cref{thm:a_root_kt} from \cref{proof_of_aroot}
with $\kinfsin[\kroot]$ replaced by $2\sinfnorm{\kroot}$.\footnote{This adaptation is also analogous to those used in the proofs of \cref{theorem:single_function_guarantee,l2pn_guarantee} albeit with different kernel and applied to different functions; and consequently all the arguments also go through if we replace $\sinfnorm{\kroot}$ by $\kinfsin[\kroot]$.} Putting it together with \cref{eq:mmd_ktplus_ktsplit}, we conclude that
\begin{talign}
     \mmd_{\kernel}(\inputcoreset, \ktpluscoreset)
    &\leq \parenth{\frac{2^m}{n} 2 \sinfnorm{\kroot}}^{\frac{1}{2\aroot}} (2\wtil{\err}_{\kroot})^{1-\frac{1}{2\aroot}} 
    \parenth{2
    \!+ \!
    \sqrt{\frac{(4\pi)^{d/2}}{\Gamma(\frac{d}{2}\!+\!1)}}
    \!\cdot \rmin_{\max}^{\frac{d}{2}} \cdot  \wtil{\err}_{\kroot}   }^{\frac1\aroot-1} \\
    &= 2^{\frac{1}{2\aroot}} \cdot \mmdakt,
    \label{eq:bound_2}
\end{talign}
with probability at least $\psg$. Let us denote this event by $\event[2]$.

Note that the quantities on the right hand side of the bounds~\cref{eq:bound_1,eq:bound_2} are deterministic given $\inputcoreset$ and thus can be computed a priori. Consequently, we apply the high probability bound only for one of the two events $\event[1]$ or $\event[2]$ depending on which of the two quantities (deterministically) attains the minimum. Thus, the bound~\cref{eq:mmd_guarantee_kt_plus} holds with probability at least $\psg$ as claimed.
\end{proofof}

\newcommand{\measure}{\mbb{D}}
\section{Proof of \cref{mmd_sandwich}: \mmdinterpolationresultname}
\label{proof_of_mmd_sandwich}

For two arbitrary distributions $\P$ and $\Q$, and any reproducing kernel $\kernel$, \citet[Lem.~4]{JMLR:v13:gretton12a} yields that
\begin{talign}
\label{eq:mmd_pq}
\mmd_{\kernel}^2(\P, \Q) = \knorm{(\P-\Q)\kernel}^2.
\end{talign}
Let $\fourier$ denote the generalized Fourier transform (GFT) operator (\citet[Def.~8.9]{wendland2004scattered}).
Since $\kernel(x, y) = \kappa(x-y)$, \citet[Thm.~10.21]{wendland2004scattered} yields that
\begin{talign}
\label{eq:mmd_fourier}
\knorm{\fun}^2 = \frac{1}{(2\pi)^{d/2}}\int_{\Rd}\frac{(\fourier(\fun)(\omega))^2}{\fourier(\kappa)(\omega)} d\omega,
\qtext{for} \fun \in \rkhs.
\end{talign}
Let $\what{\kappa}\defeq \fourier(\kappa)$, and 
consider a discrete measure $\measure = \sum_{i=1}^n w_i \dirac_{x_i}$ supported on finitely many points, and let
$\measure\kernel(x) \defeq \sum w_i \kernel(x, x_i) = \sum w_i \kappa(x-x_i)$.
Now using the linearity of the GFT operator $\fourier$, we find that for any $\omega\in\Rd$,
\begin{talign}
\fourier(\measure \kernel)(\omega) 
= \fourier(  \sum_{i=1}^n w_i \kappa(\cdot\!-\!x_i))
= \sum_{i=1}^n w_i \fourier(\kappa(\cdot\!-\!x_i)
&= (\sum_{i=1}^{n} w_i  e^{-\angles{\omega, x_i}}) \cdot \what{\kappa}(\omega) 
\\
&= \what{D}(\omega) \what{\kappa}(\omega)
\label{eq:fourier_relation}
\end{talign}
where we used the time-shifting property of GFT  that $\fourier(\kappa(\cdot\!-\!x_i))(\omega)\!=\! e^{-\angles{\omega, x_i}} \what{\kappa}(\omega)$ (proven for completeness in \cref{lem:gft_shift}),
and used the shorthand $\what{D}(\omega) \defeq (\sum_{i=1}^{n} w_i  e^{-\angles{\omega, x_i}})$ in the last step.
Putting together \cref{eq:mmd_pq,eq:mmd_fourier,eq:fourier_relation} with $\measure=\P-\Q$, we find that
\begin{talign}
\mmd_{\kernel}^2(\P, \Q) 
&= 
\frac{1}{(2\pi)^{d/2}} \int_{\Rd} \what{D}^2(\omega) \what{\kappa}(\omega) d\omega
\label{eq:mmd_pq_fourier}
\\
&= \frac{1}{(2\pi)^{d/2}} \int_{\Rd} \what{D}^2(\omega) \what{\kappa}^{\aroot}(\omega) (\what{\kappa}^{\aroot}(\omega))^{\frac{1-\aroot}{\aroot}}  d\omega \\
&= \frac{1}{(2\pi)^{d/2}} 
 \int_{\Rd} \what{D}^2(\omega') \what{\kappa}^{\aroot}(\omega')d\omega'
\int_{\Rd} \frac{\what{D}^2(\omega) \what{\kappa}^{\aroot}(\omega)}{\int_{\Rd} \what{D}^2(\omega') \what{\kappa}^{\aroot}(\omega')d\omega'} (\what{\kappa}^{\aroot}(\omega))^{\frac{1-\aroot}{\aroot}}  d\omega \\
&\sless{(i)} 
\frac{1}{(2\pi)^{d/2}}
 \int_{\Rd} \what{D}^2(\omega') \what{\kappa}^{\aroot}(\omega')d\omega'
\parenth{\int_{\Rd} \frac{\what{D}^2(\omega) \what{\kappa}^{\aroot}(\omega)}{\int_{\Rd} \what{D}^2(\omega') \what{\kappa}^{\aroot}(\omega')d\omega'} \what{\kappa}^{\aroot}(\omega)  d\omega}^{\frac{1-\aroot}{\aroot}} \\
&=
\frac{1}{(2\pi)^{d/2}}
 \parenth{\int_{\Rd} \what{D}^2(\omega') \what{\kappa}^{\aroot}(\omega')d\omega'}^{2-\frac1{\aroot}}
\parenth{\int_{\Rd} \frac{\what{D}^2(\omega) \what{\kappa}^{2\aroot}(\omega)} d\omega}^{\frac{1-\aroot}{\aroot}} \\
&=
 \parenth{\frac{1}{(2\pi)^{d/2}} \int_{\Rd} \what{D}^2(\omega') \what{\kappa}^{\aroot}(\omega')d\omega'}^{2-\frac1{\aroot}}
\parenth{\frac{1}{(2\pi)^{d/2}} \int_{\Rd} \frac{\what{D}^2(\omega) \what{\kappa}^{2\aroot}(\omega)} d\omega}^{\frac{1-\aroot}{\aroot}}\\
&\seq{(ii)} (\mmd^2_{\kroot}(\P, \Q))^{2-\frac1{\aroot}}
\cdot (\mmd^2_{\kroot[2\aroot]}(\P, \Q))^{\frac1{\aroot}-1},
\end{talign}
where step~(i) makes use of Jensen's inequality and the fact that the function $t\mapsto t^{\frac{1-\aroot}{\aroot}}$ for $t\geq 0$ is concave for $\aroot \in [\half, 1]$, and step~(ii) follows by applying \cref{eq:mmd_pq_fourier} for kernels $\kroot$ and $\kroot[2\aroot]$ and noting that by definition $\fourier(\kroot) = \what{\kappa}^{\aroot}$, and $\fourier(\kroot[2\aroot]) = \what{\kappa}^{2\aroot}$. Noting $\mmd$ is a non-negative quantity, and taking square-root establishes the claim~\cref{eq:mmd_sandwich}.

\begin{lemma}[Shifting property of the generalized Fourier transform]
\label{lem:gft_shift}
If $\what{\kappa}$ denotes the generalized Fourier transform (GFT)~\citep[Def.~8.9]{wendland2004scattered} of the function $\kappa:\real^d \to \real$, then $e^{-\angles{\cdot, x_i}}\what{\kappa}$ denotes the GFT of the shifted function $\kappa(\cdot - x_i)$, for any $x_i \in \real^d$.
\end{lemma}
\begin{proof}
Note that by definition of the GFT $\what{\kappa}$~\citep[Def.~8.9]{wendland2004scattered}, we have
\begin{talign}
\label{eq:gft_condition}
\int \kappa(x) \what{\gamma}(x) dx
=
\int \what{\kappa} (\omega) \gamma(\omega) d\omega, 
\end{talign}
for all suitable Schwartz functions $\gamma$~\citep[Def.~5.17]{wendland2004scattered}, where $\what{\gamma}$ denotes the Fourier transform~\citep[Def.~5.15]{wendland2004scattered} of $\gamma$ since GFT and FT coincide for these functions (as noted in the discussion after \citet[Def.~8.9]{wendland2004scattered}). Thus to prove the lemma, we need to verify that
\begin{talign}
\label{eq:equality_of_ft}
\int \kappa(x-x_i) \what{\gamma}(x) dx
=
\int e^{-\angles{\omega, x_i}}\what{\kappa} (\omega) \gamma(\omega) d\omega,
\end{talign}
for all suitable Schwartz functions $\gamma$.
Starting with the right hand side of the display~\cref{eq:equality_of_ft}, we have
\begin{talign}
\int e^{-\angles{\omega, x_i}}\what{\kappa} (\omega) \gamma(\omega) d\omega 
= \int \what{\kappa} (\omega) (e^{-\angles{\omega, x_i}}\gamma(\omega)) d\omega 
\seq{(i)} \int \kappa(x) \what{\gamma}(x+x_i) dx
\seq{(ii)}\int \kappa(z-x_i) \what{\gamma}(z) dz,
\end{talign}
where step~(i) follows from the shifting property of the FT~\citep[Thm.~5.16(4)]{wendland2004scattered}, and the fact that the GFT condition~\cref{eq:gft_condition} holds for the shifted function $\gamma(\cdot + x_i)$ function as well since it is still a Schwartz function (recall that $\what \gamma$ is the FT), and step~(ii) follows from a change of variable. We have thus established \cref{eq:equality_of_ft}, and the proof is complete.
\end{proof}

\section{Sub-optimality of single function guarantees with root KT}
\label{ratio_of_kernel_norms}
Define $\wtil{\kernel}_{\mrm{rt}}$ as the scaled version of $\ksqrt$, i.e.,   $\wtil{\kernel}_{\mrm{rt}} \defeq \ksqrt / \infnorm{\ksqrt}$ that is bounded by $1$.
Then \citet[Proof of Prop.~2.3]{zhang2013inclusion} implies that
\begin{talign}
\label{eq:root_norm_rescale}
    \krtnorm{f} = \frac{1}{\sqrt{\infnorm{\ksqrt}}} \norm{f}_{\wtil{\kernel}_{\mrm{rt}}}.
\end{talign}
And thus we also have $\rkhs_{\trm{rt}}  = \wtil{\rkhs}_{\trm{rt}}$ where $\rkhs_{\trm{rt}}$ and $ \wtil{\rkhs}_{\trm{rt}}$ respectively denote the RKHSs of $\ksqrt$ and $\wtil{\kernel}_{\mrm{rt}}$.

Next, we note that for any two kernels $\kernel_1$ and $\kernel_2$ with corresponding RKHSs $\rkhs_1$ and $\rkhs_2$ with $\rkhs_1 \subset \rkhs_2$, in the convention of \citet[Lem.~2.2, Prop.~2.3]{zhang2013inclusion}, we have
\begin{talign}
\label{eq:ratio_norms}
    \frac{\norm{f}_{\kernel_2}}{\norm{f}_{\kernel_1}}
    \leq \beta(\rkhs_1, \rkhs_2) \leq \sqrt{\lambda(\rkhs_1, \rkhs_2)}
    \qtext{for} f \in \rkhs.
\end{talign}

Consequently, we have
\begin{talign}
\label{eq:ratio_norm_final_bound}
    \sqrt{\max_{\x\in \inputcoreset} \ksqrt(\x,\x)} \frac{\krtnorm{f}}{\knorm{f}}
    \leq \sqrt{\infnorm{\ksqrt}} \frac{\krtnorm{f}}{\knorm{f}}
    \seq{\cref{eq:root_norm_rescale}} \frac{\norm{f}_{\wtil{\kernel}_{\mrm{rt}}}}{\knorm{\fun}}
    \leq \sqrt{\lambda(\rkhs, \wtil{\rkhs}_{\mrm{rt}})},
\end{talign}
where in the last step, we have applied the bound~\cref{eq:ratio_norms} with $(\kernel_1, \rkhs_1) \gets (\kernel, \rkhs)$ and $(\kernel_2, \rkhs_2)\gets (\wtil{\kernel}_{\mrm{rt}}, \wtil{\kernel}_{\mrm{rt}})$ since $\rkhs \subset \rkhs_{\trm{rt}} = \wtil{\kernel}_{\mrm{rt}}$.

Next, we use \cref{eq:ratio_norm_final_bound} to the kernels studied in \cite{dwivedi2021kernel} where we note that all the kernels in that work were scaled to ensure $\sinfnorm{\kernel}=1$ and in fact satisfied $\kernel(x, x) = 1$. Consequently, the multiplicative factor stated in the discussion after \cref{theorem:single_function_guarantee}, namely, $\sqrt{\frac{\kinfsin[\ksqrt]}{\kinfsin}  }\frac{\krtnorm{f}}{\knorm{f}} $ can be bounded by $ \sqrt{\lambda(\rkhs, \wtil{\rkhs}_{\mrm{rt}})}$ given the arguments above.

For $\kernel=\tbf{Gauss}(\sigma)$ kernels, \citet[Prop.~3.5(1)]{zhang2013inclusion} yields that
\begin{talign}
    \lambda(\rkhs, \wtil{\rkhs}_{\mrm{rt}}) = 2^{d/2}.
\end{talign}
For $\kernel=\tbf{B-spline}(2\beta+1, \mattwo)$ with $\beta \in 2\natural+1$, \citet[Prop.~3.5(1)]{zhang2013inclusion} yields that
\begin{talign}
    \lambda(\rkhs, \wtil{\rkhs}_{\mrm{rt}}) = 1.
\end{talign}
For $\kernel=$\tbf{\Matern}$(\matone, \mattwo)$ with $\matone > d$, some algebra along with \citet[Prop~3.1]{zhang2013inclusion} yields that
\begin{talign}
    \lambda(\rkhs, \wtil{\rkhs}_{\mrm{rt}}) = \frac{\Gamma(\matone)\Gamma((\matone-d)/2)}{\Gamma(\matone-d/2)\Gamma(\matone/2)}
    \geq 1.
\end{talign}

\section{Additional experimental results}
\label{sec:vignettes_supplement}
This section provides additional experimental details and results deferred from \cref{sec:experiments}.

\para{Common settings and error computation} To obtain an output coreset of size $n^{\half}$ with $n$ input points, we (a) take every $n^{\half}$-th point for standard thinning (ST) and (b) run KT with $\m=\half \log_2 n$ using an ST coreset as the base coreset in \ktswap. For Gaussian and MoG target we use \iid points as input, and for MCMC targets we use an ST coreset after burn-in as the input (see \cref{sec:vignettes_supplement} for more details).  We compute errors with respect to $\P$ whenever available in closed form and otherwise use $\pin$.  For each input sample size $n \in \braces{2^{4}, 2^{6}, \ldots, 2^{14}}$ with  $\delta_i = \frac{1}{2n}$, we report the mean MMD or function integration error $\pm 1$ standard error across $10$ independent replications of the experiment (the standard errors are too small to be visible in all experiments). 
We also plot the ordinary least squares fit (for log mean error vs log coreset size), with the slope of the fit denoted as the empirical decay rate, e.g., for an OLS fit with slope $-0.25$, we display the decay rate of $n^{-0.25}$.

\para{Details of test functions}
We note the following:
(a) For Gaussian targets, the error with CIF function and \iid input is measured across the sample mean over the $n$ input points and $\sqrt n$ output points obtained by standard thinning the input sequence, since $\P \fun_{\mrm{CIF}}$ does not admit a closed form. %
(b) To define the function $\fun:x\mapsto \kernel(X', x)$, first we draw a sample $X\sim \P$, independent of the input, and then set $X' = 2X$. For the MCMC targets, we draw a point uniformly from a held out data point not used as input for KT.  For each target, the sample is drawn exactly once and then fixed throughout all sample sizes and repetions. 
\subsection{Mixture of Gaussians Experiments}
\begin{figure}[ht!]
    \centering
    \includegraphics[width=\linewidth]{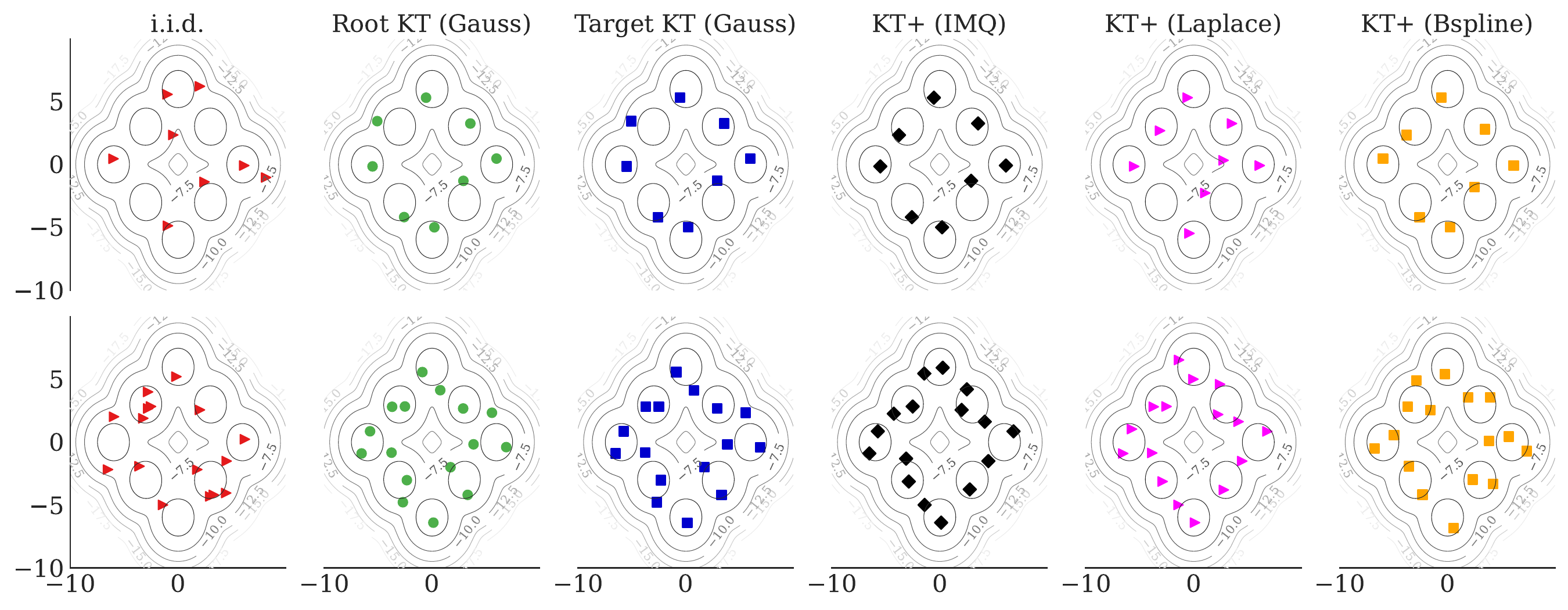}
    \\
    \includegraphics[width=\linewidth]{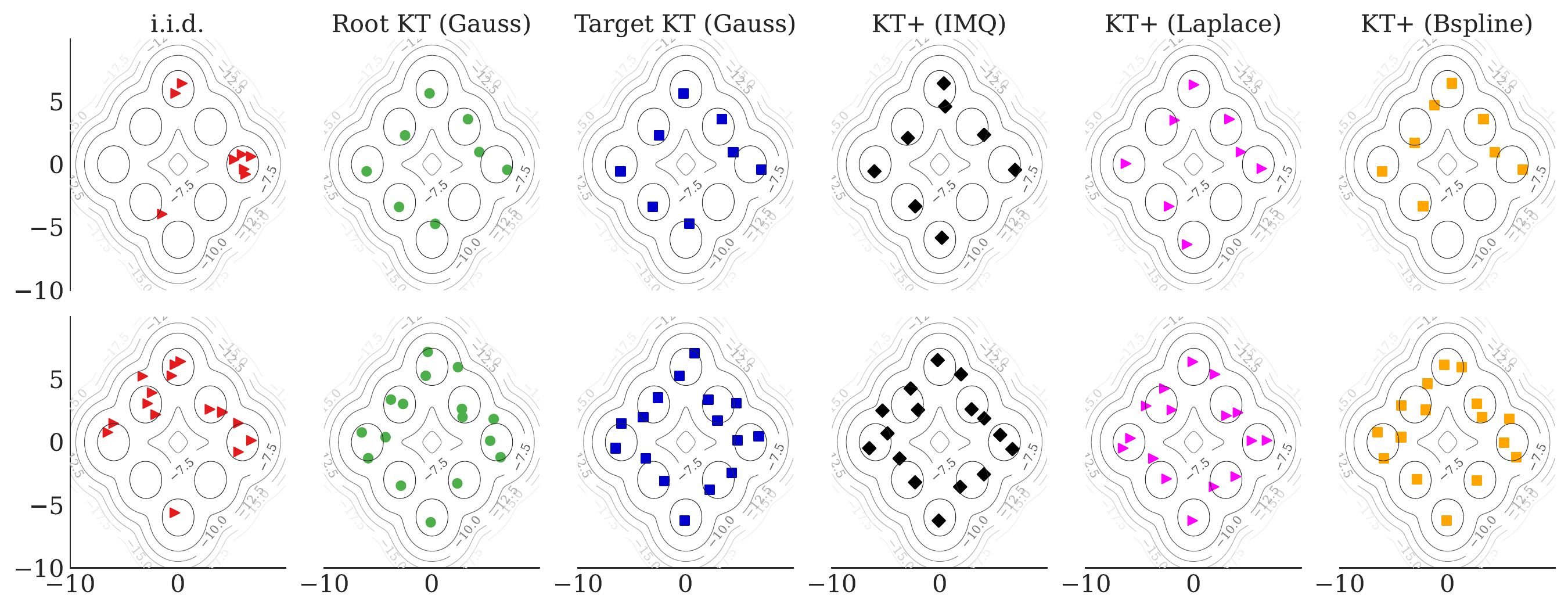}
    \caption{\tbf{Generalized kernel thinning (KT) and \iid coresets} for various kernels $\kernel$ (in parentheses) and an 8-component mixture of Gaussian target $\P$ with equidensity contours underlaid. These plots are independent replicates of  \cref{fig:mog_scatter}. 
    See \cref{sec:experiments} for more details.
    }
    \label{fig:scatter_plot_additional}
\end{figure}

\begin{figure}[ht!]
    \centering
    \includegraphics[width=\linewidth]{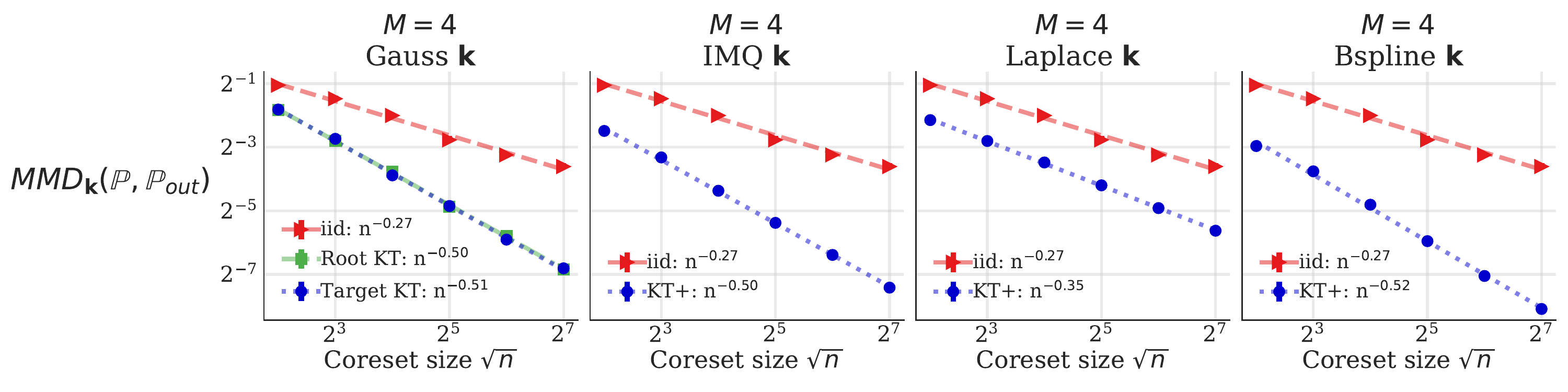}\\
    \includegraphics[width=\linewidth]{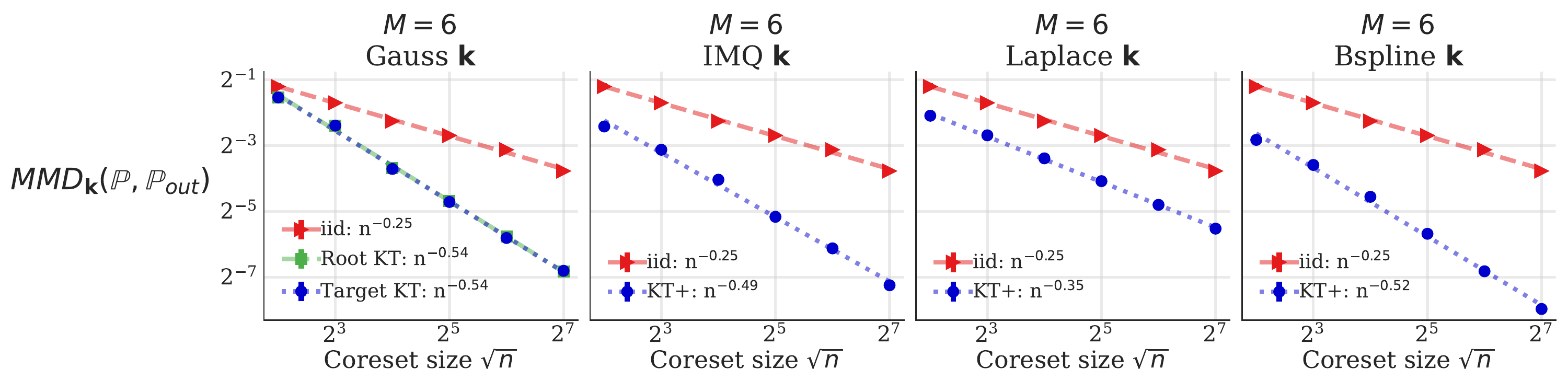}
    \caption{\tbf{Kernel thinning versus \iid sampling.} 
    For mixture of Gaussians $\P$ with $M \in \braces{4, 6}$ components and the kernel choices of \cref{sec:experiments}, the \tkt with \gauss\ $\kernel$ provides comparable $\mmd_{\kernel}(\P, \pout)$ error to the \rkt, and both provide an $n^{-\half}$ decay rate improving significantly over the $n^{-\quarter}$ decay rate from \iid\ sampling. For the other kernels, KT+ provides a decay rate close to $n^{-\half}$ for \imq\ and \bspline\ $\kernel$, and $n^{-0.35}$ for \laplace\ $\kernel$.
    See \cref{sec:experiments} for further discussion.
    }
    \label{fig:mog_mmd}
\end{figure}

Our mixture of Gaussians target is given by 
 $\P = \frac{1}{M}\sum_{j=1}^{M}\mc{N}(\mu_j, \mbf{I}_d)$ for $M \in \braces{4, 6, 8}$ where
    \begin{talign}
    \label{eq:mog_description}
    \mu_1 &= [-3, 3]\tp,  \quad \mu_2 = [-3, 3]\tp,  \quad  \mu_3 = [-3, -3]\tp,  \quad  \mu_4 = [3, -3]\tp,\\
    \mu_5 &= [0, 6]\tp, \qquad \mu_6 = [-6, 0]\tp,  \quad  \mu_7 = [6, 0]\tp,  \qquad  \mu_8 = [0, -6]\tp.
    \end{talign}
Two independent replicates of \cref{fig:mog_scatter} can be found in \cref{fig:scatter_plot_additional}.
Finally,we display mean MMD ($\pm 1$ standard error across ten independent experiment replicates) as a function of coreset size in \cref{fig:mog_mmd} for $M=4, 6$  component MoG targets. The conclusions from \cref{fig:mog_mmd} are identical to those from the bottom row of \cref{fig:mog_scatter}: \tkt and \rkt provide similar MMD errors with \gauss\ $\kernel$, and all variants of KT provide a significant improvement over \iid sampling both in terms of  magnitude and decay rate with input size. Morever the observed decay rates for KT+ closely match the rates guaranteed by our theory in \cref{table:explicit_mmd}.%

\subsection{MCMC experiments}
\label{sec:mcmc_supplement}
Our set-up for MCMC experiments follows closely that of \cite{dwivedi2021kernel}.
For complete details on the targets and sampling algorithms we refer the reader to \citet[Sec. 4]{riabiz2020optimal}.

\paragraph{Goodwin and Lotka-Volterra experiments} From \citet{DVN/MDKNWM_2020}, we use the output of four distinct MCMC procedures targeting each of two $d=4$-dimensional posterior distributions  $\P$: (1) a posterior over the parameters of the \emph{Goodwin model} of oscillatory enzymatic control \citep{goodwin1965oscillatory} and (2) a posterior over the parameters of the \emph{Lotka-Volterra model} of oscillatory predator-prey evolution \citep{lotka1925elements,volterra1926variazioni}.
For each of these targets, \citet{DVN/MDKNWM_2020} provide $2\times 10^6$ sample points from the following four MCMC algorithms: Gaussian random walk (RW), adaptive Gaussian random walk \citep[adaRW,][]{haario1999adaptive}, Metropolis-adjusted Langevin algorithm \citep[MALA,][]{roberts1996exponential}, and pre-conditioned MALA \citep[pMALA,][]{girolami2011riemann}. 

\paragraph{Hinch experiments}
\citet{DVN/MDKNWM_2020} also provide the output of two independent Gaussian random walk MCMC chains targeting each of two $d=38$-dimensional posterior distributions $\P$: (1) a posterior over the parameters of the Hinch model of calcium signalling in cardiac cells \citep{hinch2004simplified} and (2) a tempered version of the same posterior, as defined by \citet[App.~S5.4]{riabiz2020optimal}.

\para{Burn-in and standard thinning} We discard the initial burn-in points of each chain using the maximum burn-in period reported in \citet[Tabs. S4 \& S6, App.~S5.4]{riabiz2020optimal}. 
Furthermore, we also normalize each Hinch chain by subtracting the post-burn-in sample mean and dividing each coordinate by its post-burn-in sample standard deviation.
To obtain an input sequence $\inputcoreset$ of length $n$ to be fed into a thinning algorithm, we downsample the remaining even indices of points using standard thinning (odd indices are held out).
When applying standard thinning to any Markov chain output, we adopt the convention of keeping the final sample point.

The selected burn-in periods for the Goodwin task were 820,000 for RW; 824,000 for adaRW; 1,615,000 for MALA; and 1,475,000 for pMALA. The respective numbers for the Lotka-Volterra task were 1,512,000 for RW; 1,797,000 for adaRW; 1,573,000 for MALA; and 1,251,000 for pMALA.

\para{Additional remarks on \cref{fig:mcmc}}
When a Markov chain is fast mixing (as in the Goodwin and Lotka-Volterra examples), we expect standard thinning to have $\Omega(n^{-\quarter})$ error.  However, when the chain is slow mixing, standard thinning can enjoy a faster rate of decay due to a certain degeneracy of the chain that leads it to lie close to a one-dimensional curve. 
In the Hinch figures, we observe these better-than-\iid  rates of decay for standard thinning, but, remarkably, KT+ still offers improvements in both MMD and integration error. Moreover, in this setting, every additional point discarded via improved compression translates into thousands of CPU hours saved in downstream heart-model simulations.

\begin{figure}[ht!]
    \centering
    \includegraphics[width=0.835\linewidth]{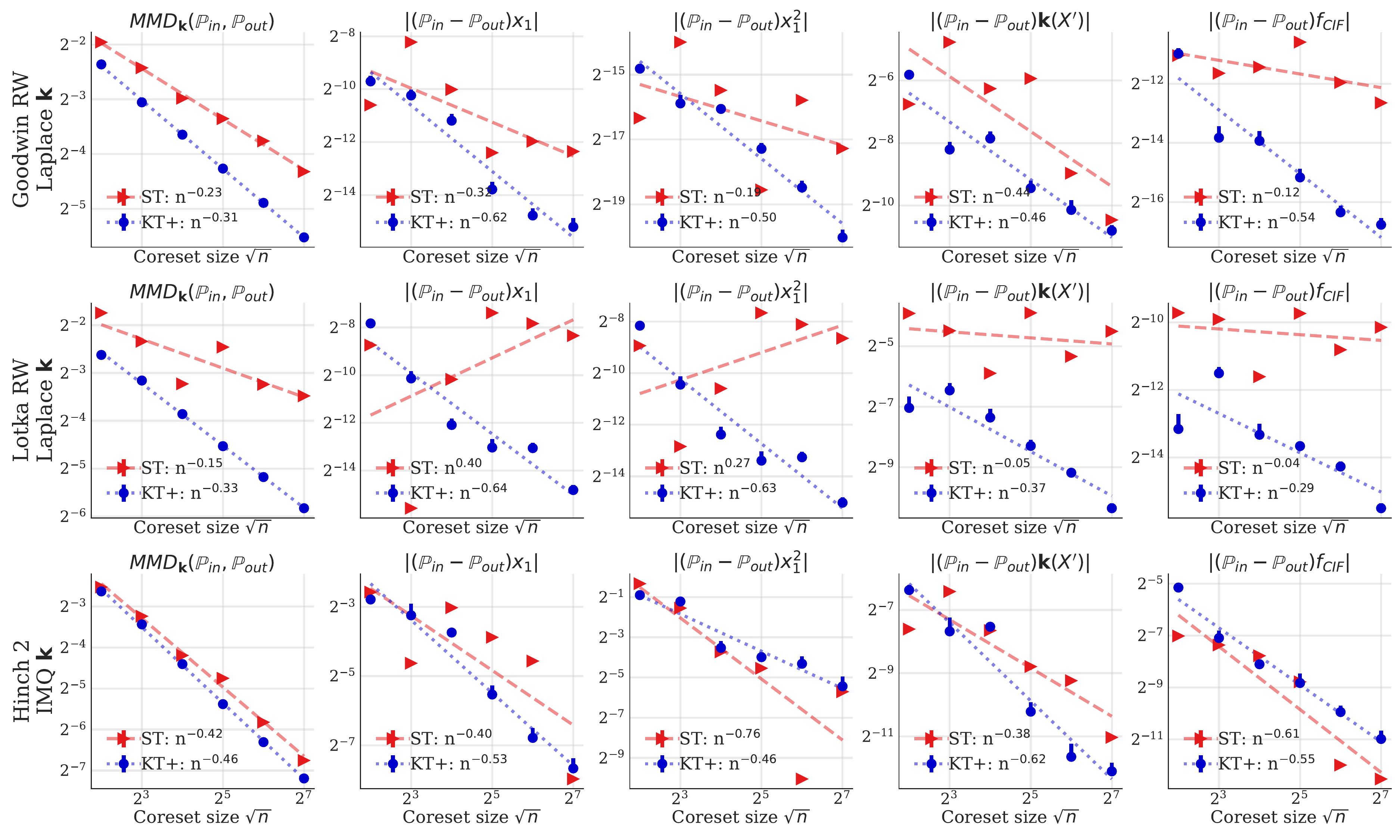}
    \includegraphics[width=0.835\linewidth]{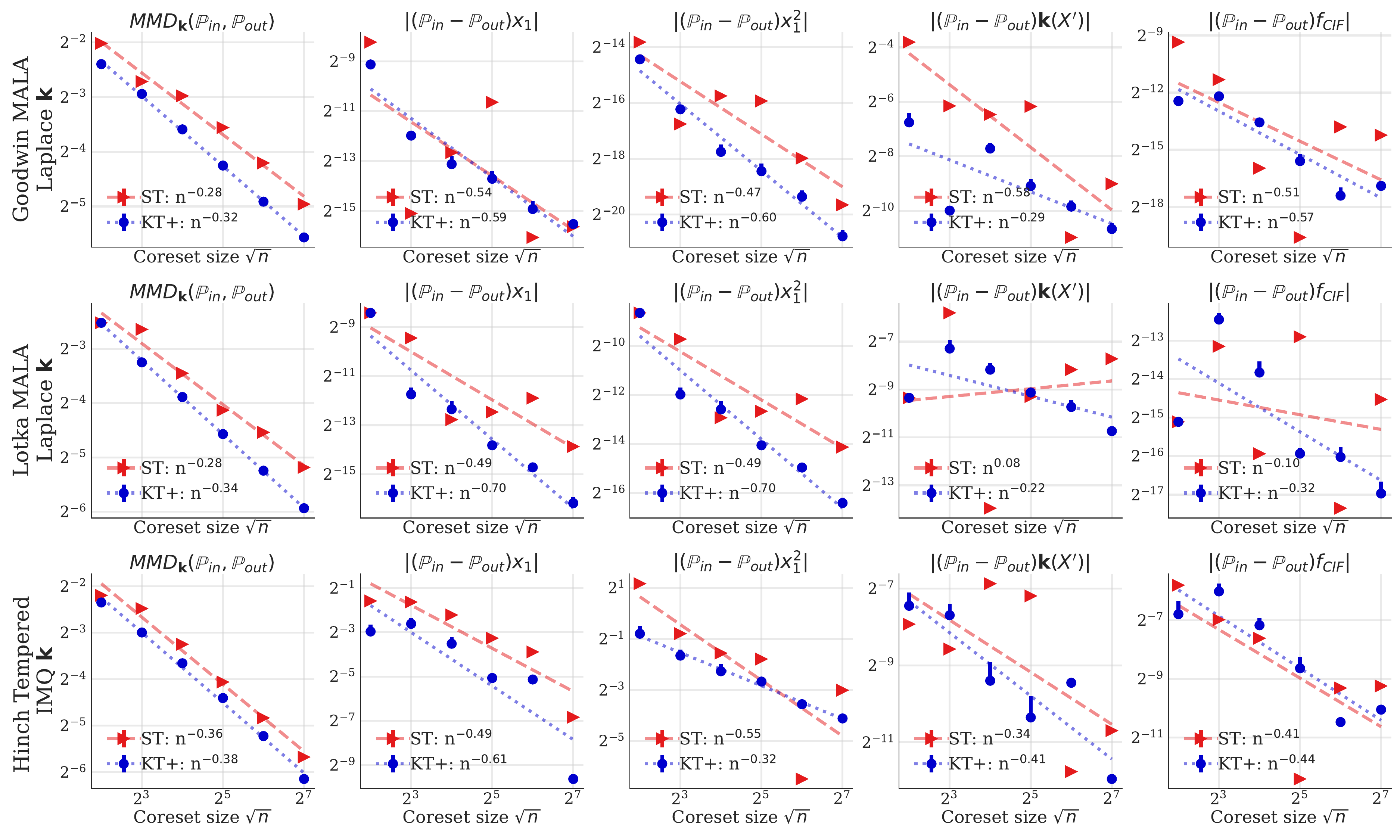}
    \includegraphics[width=0.835\linewidth]{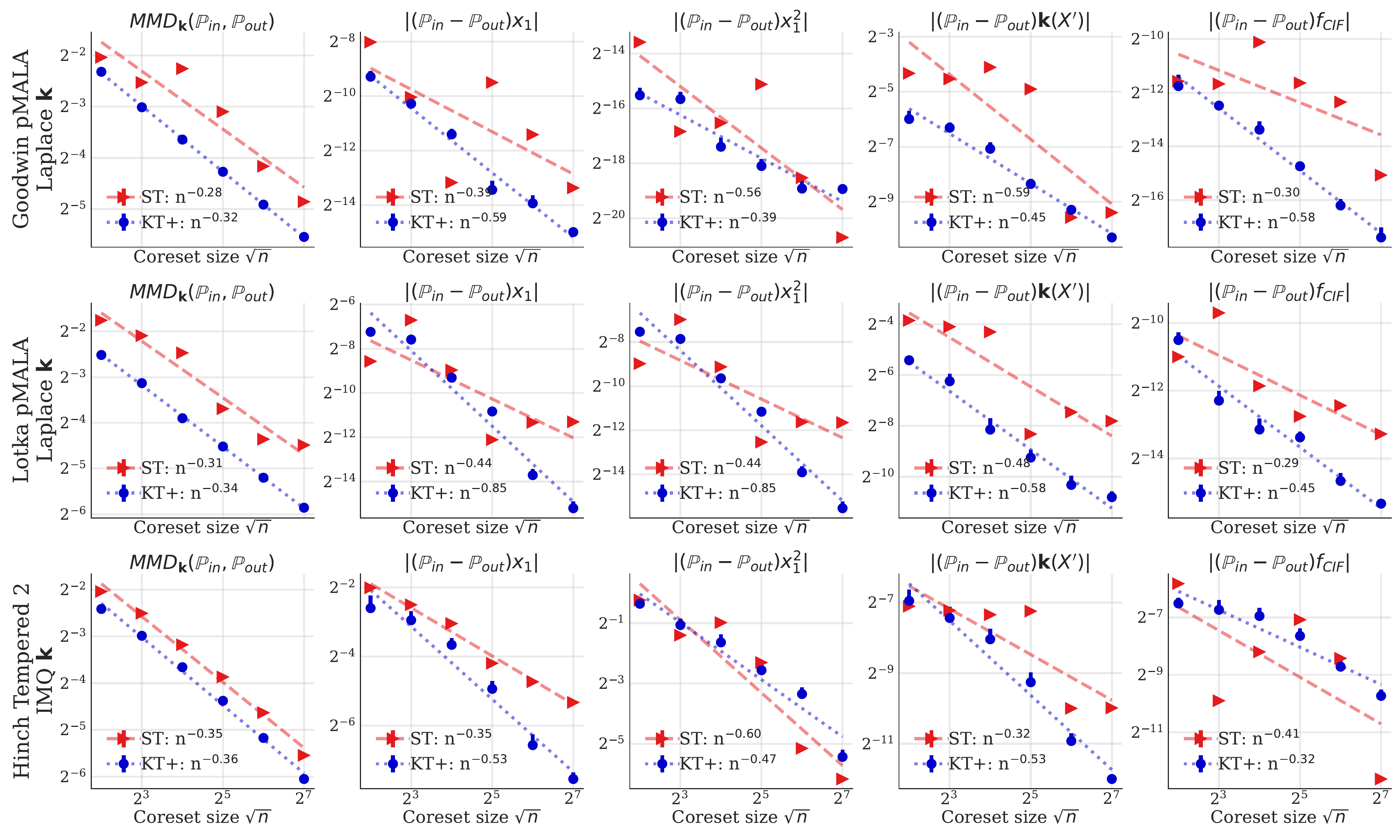}
    \caption{\tbf{Kernel thinning+ (KT+) vs. standard MCMC thinning (ST).}  For kernels without fast-decaying square-roots, KT+ improves MMD and integration error decay rates in each posterior inference task. 
    }
    \label{fig:mcmc_supplement}
\end{figure}

\section{Upper bounds on RKHS covering numbers}
\label{sec:bounds_on_rkhs_convering}
In this section, we state several results on covering bounds for RKHSes for both generic and specific kernels. We then use these bounds with \cref{theorem:mmd_guarantee} (or \cref{table:mmd_rates}) to establish MMD guarantees for the output of generalized kernel thinning as summarized in \cref{table:explicit_mmd}.

We first state covering number bounds for RKHS associated with generic kernels, that are either (a) analytic, or (b) finitely many times differentiable. These results follow essentially from \cite{sun2008reproducing,steinwart2008support}, but we provide a proof in \cref{sec:proof_of_general_rkhs_covering_bounds} for completeness.
\newcommand{\generalcoveringresultname}{Covering numbers for analytic and differentiable kernels}
\begin{proposition}[\tbf{\generalcoveringresultname}]
\label{general_rkhs_covering_bounds}
    The following results hold true.
    \begin{enumerate}[label=(\alph*),leftmargin=*]
        \item\label{item:analytic_kernel} \tbf{Analytic kernels:} Suppose that $\kernel(\x,\y) = \kappa(\twonorm{\x-\y}^2)$ for $\kappa:\real_{+}\to\real$ real-analytic with convergence radius $R_{\kappa}$, that is,
        \begin{talign}
        \label{eq:analytic}
            \abss{\frac{1}{j!}\kappa_+^{(j)}(0)} \leq C_{\kappa} \parenth{2/R_{\kappa}}^{j}
            \qtext{for all}
            j \in \naturals_0
        \end{talign}
        for some constant $C_{\kappa}$, where $\kappa_+^{(j)}$ denotes the right-sided $j$-th derivative of $\kappa$. Then for any set $\set A \subset \Rd$ and any $\vareps \in (0, \half)$, we have
        \begin{talign}
        \label{eq:analytic_cover_bound}
             \entropy(\set{A}, \vareps) &\leq \coveringnumber_{2}(\set{A}, r^{\dagger}/2) \cdot \parenth{4\log\parenth{1/{\vareps}} + 2 + 4 \log(16\sqrt{C_{\kappa}}+1) }^{d+1},\\ 
             \text{where } r^{\dagger} &\defeq \min\parenth{\frac{\sqrt{R_{\kappa}}}{2d}, \sqrt{R_{\kappa}+D^2_{\set A}} - D_{\set{A}}}, \text{ and }
             D_{\set A} \defeq \max_{x,y\in\set A} \twonorm{x-y}.
             \label{eq:special_radius}
        \end{talign}     
    \item\label{item:differentiable_kernel} \tbf{Differentiable kernels:}  Suppose that for $\mc X \subset \Rd$, the kernel $\kernel:\mc X \times \mc X \to \real$ is $s$-times continuously differentiable, i.e., all partial derivatives $\partial^{\alpha, \alpha}\kernel: \mc X \times \mc X \to \real$ exist and are continuous for all multi-indices $\alpha \in \natural_0^d$ with $\abss{\alpha}\leq s$. Then, for any closed Euclidean ball $\bar{\ball}_2(r)$ contained in $\mc X$ and any $\vareps>0$, we have
        \begin{talign}
        \label{eq:differentiable_kernel_cover_bound}
            \entropy(\bar{\ball}_2(r), \vareps) \leq c_{s, d, \kernel} \cdot r^{d} \cdot (1/\vareps)^{d/s},
        \end{talign}
        for some constant $c_{s, d, \kernel}$ that depends only on on $s, d$ and $\kernel$.
    \end{enumerate}
\end{proposition}

Next, we state several explicit bounds on covering numbers for several popular kernels. See \cref{sec:proof_of_rkhs_covering_numbers} for the proof.

\newcommand{\specificcoveringresultname}{Covering numbers for specific kernels}
\begin{proposition}[\tbf{\specificcoveringresultname}]
\label{rkhs_covering_numbers}
 The following statements hold true.
    \begin{enumerate}[label=(\alph*),leftmargin=*]
        \item\label{item:gauss_cover} When $\kernel = \gauss(\gaussparam)$, we have
        \begin{talign}
            \entropy(\ball_2(r), \vareps) &\leq C_{\mrm{Gauss},d} \cdot  \parenth{\frac{\log(4/\vareps)}{\log\log(4/\vareps)} }^d  \log(4/\vareps) \cdot
            \begin{cases}
            1 \qquad\ \ \quad\text{ when } r \leq %
            {\sqrt 2\gaussparam}, \\
            (3r/(\sqrt 2 \gaussparam))^d \text{ otherwise, }
            \end{cases}
            \label{eq:gauss_cover}\\
            \text{where }
            C_{\gauss,d} &\defeq {4e+d \choose d} e^{-d} \leq \begin{cases}
                4.3679 \ \ \qquad\text{ for } d=1 \\
                0.05 \cdot d^{4e}e^{-d}\text{ for } d\geq 2 
            \end{cases}
            \leq 30 \text{ for all } d \geq 1.
            \label{eq:gauss_const}
        \end{talign}
        \item\label{item:matern_kernel} When $\kernel = \maternk(\matone, \mattwo)$, $\matone\geq \frac d 2+1$, then for some constant  $ C_{\maternk, \matone, \mattwo, d}$,  we have
        \begin{talign}
        \label{eq:matern_cover}
        \entropy(\ball_2(r), \vareps) &\leq C_{\maternk, \matone, \mattwo, d} \cdot  r^d \cdot  (1/\vareps)^{d/\floor{\matone-\frac d2}}.
        \end{talign}
        \item\label{item:imq_cover} When $\kernel = \imq(\matone, \mattwo)$, we have
        \begin{talign}
        \label{eq:imq_cover}
            \entropy(\ball_2(r), \vareps) &\leq 
           (1+\frac{4r}{\wtil{r}})^d \cdot 
            \parenth{4\log(1/\vareps)+2+C_{\imq, \matone, \mattwo} }^{d+1}, \\
            \label{eq:imq_const}
            \text{where }
            C_{\imq, \matone, \mattwo} &\defeq 4\log\parenth{16 \frac{(2\matone+1)^{\matone+1}}{\mattwo^{2\matone}} \!+ \!1}, 
            \text{ and }
            \wtil{r} \defeq \min\parenth{\frac{ \mattwo}{2d}, \sqrt{ \mattwo^2+4r^2}\!-\!2r}.
        \end{talign}
        \item\label{item:sinc_cover}When $\kernel = \sinc(\sincparam)$, then for $\vareps \in (0, \half)$, we have
        \begin{talign}
        \label{eq:sinc_cover}
            \entropy([-r, \r]^{d}, \vareps) 
    &\leq 
    2d\log 2 \cdot (g_{\sinc, r\theta, d}(\vareps)+1)^{d} \bigparenth{g_{\sinc, r\theta, d}(\vareps)  + \log(\frac{16}{\vareps})}
         \\ 
        \text{where } g_{\sinc, r\theta, d}(\vareps) &\defeq \max\braces{1,  \ \ \ceil{2r\theta}, \ \ \log((\frac{3}{2})^d \cdot \frac{32d}{3\vareps^2})}.
        \end{talign}
        \item\label{item:spline_kernel} When $\kernel = \bspline(2\splineparam+1, \mattwo)$, then for some constant $ C_{\bspline, \beta,\gamma,  d}$,  we have
        \begin{talign}
        \label{eq:spline_cover}
        \entropy(\ball_2(r), \vareps) &\leq  C_{\bspline, \beta, \gamma, d} \cdot r^{d} \cdot (1/\vareps)^{d/\beta}.
        \end{talign}
    \end{enumerate}
\end{proposition}

\subsection{Auxiliary results about RKHS and Euclidean covering numbers}

In this section, we collect several results regarding the covering numbers of Euclidean and RKHS spaces that come in handy for our proofs. These results can also be of independent interest.

We start by defining the notion of restricted kernel and its unit ball~(\citet[Prop.~8]{rudi2020finding}). For $\set X \subset \Rd$, let $\restrict[\set{X}]$ denotes the restriction operator. That is, for any function $\fun:\real^d\to\real$, we have $\fun\restrict[\set{X}] : \set{X} \to \real$ such that $\fun\restrict[\set{A}](x) = \fun(x)$ for $x \in \set{X}$.
\begin{definition}[Restricted kernel and its RKHS]
\label{def:restricted_kernel}
Consider a kernel $\kernel$ defined on $\Rd \times \Rd$ with the corresponding RKHS $\rkhs$, any set $\set{X} \subset \Rd$. The restricted kernel $\kernel_{\vert\set X}$ is defined as
\begin{talign}
\label{eq:restrict_kernel}
\kernel\restrict[\set{X}]: \set{X}\times \set{X} \to \real
\qtext{such that} \kernel\restrict[\set{X}](x, y) \defeq \kernel\restrict[\set{X} \times \set{X}](x, y) =  \kernel(x, y)
\qtext{for all} x, y \in \set{X},
\end{talign}
and $\rkhs\restrict[\set{X}]$ denotes its RKHS. For $\fun\in\rkhs\restrict[\set{X}]$, the restricted RKHS norm is defined as follows:
\begin{talign}
\label{eq:restrict_norm}
\norm{\fun}_{\kernel\restrict[\set{X}]} = \inf_{h\in\rkhs}\knorm{h}
\qtext{such that} h\restrict[\set{X}] = \fun.
\end{talign}
Furthermore, we use $\kball[\vert \set{X}] \defeq \{ \fun \in \rkhs\vert_{\set{X}} : \norm{\fun}_{\kernel\vert_{\set{X}}} \leq 1\}$ to denote the unit ball of the RKHS corresponding to this restricted kernel.
\end{definition}

In this notation, the unit ball of unrestricted kernel satisfies $\kball \defeq \kball[\vert\Rd]$. Now, recall the RKHS covering number definition from \cref{rkhs_covering}. In the sequel, we also use the covering number of the restricted kernel defined as follows:
\begin{talign}
\label{eq:covering_number_restrict}
\coveringnumber_{\kernel}^\dagger(\set{X}, \vareps) = \coveringnumber_{\kernel\restrict[\set X]}(\set{X}, \vareps),
\end{talign}
that is  
$ \coveringnumber_{\kernel}^\dagger(\set{X}, \vareps)$  denotes the minimum cardinality over all possible covers $\cover \subset \kball[\vert \set{X}]$ that satisfy
	\begin{talign}
	\label{eq:restricted_cover}
		\kball[\vert \set{X}] \subset \bigcup_{h \in \cover } \braces{g \!\in\! \kball[\vert \set{X}]\!:\! \sup_{x\in\set{\set{X}}}|h(x)\!-\!g(x)|\!\leq\! \varepsilon}.
	\end{talign}

With this notation in place, we now state a result that relates the covering numbers $\coveringnumber^{\dagger}$~\cref{eq:covering_number_restrict} and $\coveringnumber$~\cref{rkhs_covering}.
\begin{lemma}[Relation between restricted and unrestricted RKHS covering numbers]
    \label{rkhs_cover_restriction_domain}
    We have
    \begin{talign}
    \coveringnumber_{\kernel, \vareps}(\set{X}) \leq
        \coveringnumber_{\kernel, \vareps}^\dagger(\set{X})
    \end{talign}
\end{lemma}
\begin{proof}
\citet[Prop.~8(d,f)]{rudi2020finding} imply that there exists a bounded linear extension operator $E:\rkhs\vert_{\set X}\to\rkhs$ with operator norm bounded by $1$, which when combined with \citet[eqns.~(A.38), (A.39)]{steinwart2008support} yields the claim.
\end{proof}

Next, we state results that relate RKHS covering numbers for a change of domain for a shift-invariant kernel. 
We use $\ball_{\norm{\cdot}}(x; r) \defeq \braces{y\in\real^d: \norm{x-y}\leq r}$ to denote the $r$ radius ball in $\Rd$ defined by the metric induced by a norm $\norm{\cdot}$.

\begin{definition}[Euclidean covering numbers]
Given a set $\set{X} \subset \Rd$, a norm $\norm{\cdot}$, and a scalar $\vareps>0$, we use $\coveringnumber_{\norm{\cdot}}(\set{X}, \vareps)$ to denote the $\vareps$-covering number of $\set{X}$ with respect to $\norm{\cdot}$-norm. That is, $\coveringnumber_{\norm{\cdot}}(\set{X}, \vareps)$ denotes the minimum cardinality over all possible covers $\cover \subset \set{X}$  that satisfy
\begin{talign}
    \set{X} \subset \cup_{z\in\cover} \ball_{\norm{\cdot}}(z; \vareps).
\end{talign}
When $\norm{\cdot} = \norm{\cdot}_{q}$ for some $q \in [1, \infty]$, we use the shorthand $\coveringnumber_{q} \defeq \coveringnumber_{\norm{\cdot}_q}$.
\end{definition}

\begin{lemma}[Relation between RKHS covering numbers on different domains]
\label{rkhs_cover_change_domain}
Given a shift-invariant kernel $\kernel$, a norm $\norm{\cdot}$ on $\Rd$, and any set $\set{X} \subset \Rd$, we have
    \begin{talign}
     \coveringnumber_{\kernel}^\dagger(\set{X}, \vareps)
     \leq  \brackets{\coveringnumber_{\kernel}^\dagger(\ball_{\norm{\cdot}}, \vareps)}^{\coveringnumber_{\norm{\cdot}}(\set{X}, 1)}.
    \end{talign}
\end{lemma}
\begin{proof}
Let $\cover \subset \set X$ denote the cover of minimum cardinality such that
\begin{talign}
	\set{X} \subseteq \bigcup_{z\in\cover} \ball_{\norm{\cdot}}(z, 1).
\end{talign}
We then have
\begin{talign}
 	\coveringnumber^\dagger_{\kernel}(\set{X}, \vareps) \sless{(i)} 
 	\prod_{z\in \cover} \coveringnumber^\dagger_{\kernel}(\ball_{\norm{\cdot}}(z, 1), \vareps)
 	\sless{(ii)} \prod_{z\in \cover} \coveringnumber^\dagger_{\kernel}(\ball_{\norm{\cdot}}, \vareps)
 	\leq \brackets{\coveringnumber^\dagger_{\kernel}(\ball_{\norm{\cdot}}, \vareps)}^{\abss{\cover}},
 \end{talign} 
 where step~(i) follows by applying \citet[Lem.~3.11]{steinwart2021closer},\footnote{\citet[Lem.~3.11]{steinwart2021closer} is stated for disjoint partition of $\set{X}$ in two sets, but the argument can be repeated for any finite cover of $\set X$.} and step~(ii) follows by applying \citet[Lem.~3.10]{steinwart2021closer}. The claim follows by noting that $\cover$ denotes a cover of minimum cardinality, and hence by definition $\abss{\cover} = \coveringnumber_{\norm{\cdot}}(\set{X}, 1)$.
\end{proof}

\begin{lemma}[Covering number for shift-invariant kernels with compactly supported spectral density]
\label{lem:compact_covering}
    Suppose $\kappa : \Rd\to\reals$ denotes the Fourier transform \begin{talign}
    \kappa(z) = \frac{1}{(2\pi)^d} \int_{\real^d} \what\kappa(\xi) e^{-iz\xi} d\xi\end{talign} 
    of a bounded nonnegative function $\what\kappa$ supported on $[-a, a]^d$ for a finite $a>0$. 
    Then the shift-invariant kernel $\kernel(x, y) = \kappa(x-y)$ satisfies
    \begin{talign}
        \entropy([0, 1]^d, \vareps)
        &\leq   2d\log 2 \cdot (N_{\kappa, a, d}+1)^{d} \bigparenth{N_{\kappa, a, d}  + \log(\frac{16\sqrt{\kappa(0)}}{\vareps})}
        \label{eq:entropy_bound_zhou}
         \\ 
        \qtext{where} N_{\kappa, a, d} &\defeq \max\braces{1,  \ \ \ceil{2a}, \ \ \log((\frac{3a}{2\pi})^d \cdot \frac{32d\sinfnorm{\what\kappa}}{3\vareps^2})}.
    \end{talign}
\end{lemma}
\begin{proof}
    Our proof makes use of \citet[Thm.~2]{zhou2002covering}.\footnote{While stated differently, the proof of \citet[Thm.~2]{zhou2002covering} only makes use of the fact that $\kappa$ is the Fourier transform of a non-negative function $\what\kappa$.} In that result, the author bounds the \emph{external} covering number of the balls $\{f\in \rkhs:\norm{f}_{\kernel}\leq R\}$ in RKHS using centers from the class of continuous functions in $\norm{\cdot}_{\infty}$-norm. Notably, given an $\vareps$-cover $\mc C = \{f_1, \ldots, f_k\}$ of smallest size that comprises of continuous functions for the unit RKHS ball $\ball_{\kernel}$, we can immediately identify an \emph{internal} $2\vareps$-cover $\{g_1, g_2, \ldots, g_k\}$ with $g_j \in \ball_{\kernel}$ for each $j\in[k]$. To see this claim, for each $f_j \in \mc C$, choose an arbitrary $g_j \in \ball_{\kernel}$ in the $\vareps$-ball centered around $f_j$. Note that such a $g_j$ exists since $\mc C$ is a cover of smallest size. Now for any $g \in \ball_{\kernel}$, there exists an $f_j \in \mc C$ such that $\sinfnorm{g-f_j}\leq \vareps$ by the definition of cover, and consequently $\sinfnorm{g-g_j}\leq \sinfnorm{g-f_j} + \sinfnorm{f_j-g_j} \leq 2\vareps$ by triangle inequality and the definition of $g_j$. Our claim then follows.

    Using this claim and substituting $n\gets d$, $R\gets1$, and $\eta \gets \vareps/2$ in \citet[Thm.~1]{zhou2002covering} we find that the righthand side of \citet[(4.5)]{zhou2002covering} is a valid upper bound on $\entropy([0,1]^{d}, \vareps)$ in our notation:
    \begin{talign}
        \entropy([0,1]^{d}, \vareps)
        &\leq (N+1)^d \log\brackets{8\sqrt{\kappa(0)} (N+1)^{d/2} (N2^N)^d \frac{2}{\vareps}} \\ 
        &\leq (N+1)^{d+1} \cdot d \log 2 + (N+1)^d \brackets{\frac{3d}{2} \log (N+1) + \log(\frac{16\sqrt{\kappa(0)}}{\vareps})} \\
        &\sless{(i)} 2d\log 2 \cdot (N+1)^{d+1} +  (N+1)^d \log(\frac{16\sqrt{\kappa(0)}}{\vareps}),
        \label{eq:cover_bound_N}
    \end{talign}
    for any positive integer $N$ satisfying $\lambda_k(N) \leq (\frac{(\vareps/2)}{(2\cdot 1)})^2 = \frac{\vareps^2}{16}$, 
    where
    \begin{talign}
        \lambda_k(N) &\defeq  \frac{d(1+2^{-N})^{d-1}}{(2\pi)^d} \max_{j\in[d]}\int_{\xi\in[-N/2,N/2]^d}\what{\kappa}(\xi) \frac{\abss{\xi_j}^N}{N^N} d\xi
        \\ 
        &\qquad+ \frac{(1+(N2^N)^d)^2 }{(2\pi)^d} \int_{\xi\not\in[-N/2,N/2]^d}\what{\kappa}(\xi)  d\xi.
        \label{eq:lam_k_N}
    \end{talign}
    In the display~\cref{eq:cover_bound_N}, step~(i) follows from the fact that $3\log x \leq 2x\log 2$ for all $x\geq 2$ and $N+1\geq 2$.

    Now for any $N \geq \ceil{2a}$, the second term in the display~\cref{eq:lam_k_N} is zero. For any such $N$, we find that
    \begin{talign}
    \max_{j\in[d]}\int_{\xi\in[-N/2,N/2]^d}\what{\kappa}(\xi) \frac{\abss{\xi_j}^N}{N^N} d\xi
        &=
        \max_{j\in[d]}\int_{\xi\in[-a,a]^d}\what{\kappa}(\xi) \frac{\abss{\xi_j}^N}{N^N} d\xi \\
        &\leq \frac{\sinfnorm{\what\kappa}}{N^N} \cdot \int_{\xi\in[-a,a]^d} \frac{\abss{\xi_1}^N}{N^N} d\xi    \\ 
        &= \frac{\sinfnorm{\what\kappa} (2a)^{d-1}}{N^N} \cdot \int_{\xi_1\in[-a,a]} 
        \abss{\xi_1}^Nd\xi_1\\
        &= \frac{\sinfnorm{\what\kappa} (2a)^{d-1}}{N^N} \cdot \frac{2a^{N+1}}{N+1} \\ 
        &= \frac{\sinfnorm{\what\kappa} 2^d a^{d+N}}{N^{N+1}} \cdot (1+N\inv)\inv.
    \end{talign}
    Now to achieve,
    \begin{talign}
         \lambda_{\kappa}(N) \leq \frac{d(1+2^{-N})^{d-1}}{(2\pi)^d}  \cdot \frac{\sinfnorm{\what\kappa} 2^d a^{d+N}}{N^{N+1}} \cdot (1+N\inv)\inv
         \leq \frac{\vareps^2}{16},
    \end{talign}
    noting that for any $N\geq 1 \vee \ceil{2a}$,
    \begin{talign}
        \frac{d(1+2^{-N})^{d-1}}{(2\pi)^d} \cdot \frac{\sinfnorm{\what\kappa} 2^d a^{d+N}}{N^{N+1}}  \cdot (1+N\inv)\inv
        \leq \frac{2d\sinfnorm{\what\kappa}}{3}
        \frac{(a(1+2^{-N})/\pi)^d}{(N/a)^{N}},
    \end{talign}
    it suffices to choose
    \begin{talign}
        \frac{N}{a}\log(\frac{N}{a}) \geq \frac{1}{a} \log( (\frac{3a}{2\pi})^d\cdot \frac{32d\sinfnorm{\what\kappa}}{3\vareps^2}),
    \end{talign}
    for which it suffices to choose
    \begin{talign}
        N \geq 1  \vee \ceil{2a} \vee \parenth{\log((\frac{3a}{2\pi})^d \cdot \frac{32d\sinfnorm{\what\kappa}}{3\vareps^2})}.
        \label{eq:N_value}
    \end{talign}
    Substituting the choice~\cref{eq:N_value} into \cref{eq:cover_bound_N} yields the claimed bound in \cref{eq:entropy_bound_zhou}.
\end{proof}

\begin{lemma}[Relation between Euclidean covering numbers]
\label{euclidean_covering_numbers}
    We have
    \begin{talign}
    \coveringnumber_{\infty}(\ball_2(r), 1)
        \leq \frac{1}{\sqrt{\pi d}} \cdot \brackets{(1+\frac{2r}{\sqrt d}) \sqrt{2\pi e}}^d 
        \qtext{for all} d \geq 1.
    \end{talign}
\end{lemma}
\begin{proof}
We apply \citet[Lem.~5.7]{wainwright2019high} with $\ball = \ball_2(r)$ and $\ball' = \ball_{\infty}(1)$ to conclude that
\begin{talign}
	\coveringnumber_{\infty}(\ball_2(r), 1)
	\leq \frac{\vol(2\ball_2(r)+\ball_{\infty}(1))}{\vol(\ball_{\infty}(1))} \leq \vol(\ball_2(2r+\sqrt d))
	\leq \frac{\pi^{d/2}}{\Gamma(\frac d2+1)} \cdot (2r+\sqrt d)^{d},
\end{talign}
where $\vol(\set X)$ denotes the $d$-dimensional Euclidean volume of $\set X \subset \Rd$, and $\Gamma(a)$ denotes the Gamma function.
Next, we apply the following bounds on the Gamma function from \citet[Thm.~2.2]{batir2017bounds}:
\begin{talign} 
\label{eq:stirling_approx} 
\Gamma(b+1) \geq (b/e)^{b} \sqrt{2\pi b} 
\text{ for any } b \geq 1,
\qtext{and}
\Gamma(b+1)  \leq (b/e)^{b} \sqrt{e^2 b}
\text{ for any } b \geq 1.1.
\end{talign}
Thus, we have
\begin{talign}
	\coveringnumber_{\infty}(\ball_2(r), 1)
	\leq  \frac{\pi^{d/2} }{\sqrt{2\pi d} (\frac d{2e})^{d/2}} \cdot (2r+\sqrt d)^{d}
	\leq \frac{1}{\sqrt {\pi d}} \cdot \brackets{(1+\frac{2r}{\sqrt d})  \sqrt{2e\pi} }^d,
\end{talign}
as claimed, and we are done.
\end{proof}

\subsection{Proof of \cref{general_rkhs_covering_bounds}: \generalcoveringresultname}
\label{sec:proof_of_general_rkhs_covering_bounds}
First we apply \cref{rkhs_cover_restriction_domain} so that it remains to establish the stated bounds simply on $\log \coveringnumber^\dagger_{\kernel}(\mc X, \vareps)$.

\begin{proofof}{bound~\cref{eq:analytic_cover_bound} in part~\cref{item:analytic_kernel}} %
 The bound~\cref{eq:analytic_cover_bound} for the real-analytic kernel is a restatement of \citet[Thm.~2]{sun2008reproducing} in our notation (in particular, after making the following substitutions in their notation: $R\gets 1, C_0 \gets C_{\kappa}, r\gets R_{\kappa}, \mc X \gets \set A, \wtil{r} \gets r^{\dagger}, \eta \gets \vareps, D \gets D_{\set A}^2, n \gets d$).
\end{proofof}
 
\begin{proofof}{bound~\cref{eq:differentiable_kernel_cover_bound} for part~\cref{item:differentiable_kernel}:} Under these assumptions, \citet[Thm.~6.26]{steinwart2008support} states that the $i$-th dyadic entropy number~\citet[Def.~6.20]{steinwart2008support} of the identity inclusion mapping from $\rkhs\vert_{\bar{\ball}_2(r)}$ to $\Linf_{\bar{\ball}_2(r)}$ is bounded by $c_{s, d, \kernel}'\cdot r^{s} i^{-s/d}$ for some constant $c'_{s, d, \kernel}$ independent of $\vareps$ and $r$. Given this bound on the entropy number, and applying \citet[Lem.~6.21]{steinwart2008support}, we conclude that the log-covering number $\log \coveringnumber^\dagger_{\kernel}(\bar{\ball}_2(r), \vareps)$ is bounded by $\ln 4 \cdot (c_{s, d, \kernel}' r^{s}/\vareps)^{d/s} = c_{s, d, \kernel} r^{d} \cdot (1/\vareps)^{d/s}$ as claimed.
\end{proofof}

\subsection{Proof of \cref{rkhs_covering_numbers}: \specificcoveringresultname} %
\label{sec:proof_of_rkhs_covering_numbers}

First we apply \cref{rkhs_cover_restriction_domain} so that it remains to establish the stated bounds in each part on the corresponding $\log \coveringnumber_{\kernel}$. 

\begin{prooffor}{\gauss\ kernel: Part~\cref{item:gauss_cover}} The bound~\cref{eq:gauss_cover} for the Gaussian kernel follows directly from \citet[Eqn.~11]{steinwart2021closer} along with the discussion stated just before it. Furthermore, the bound~\cref{eq:gauss_const} for $C_{\mrm{Gauss},d}$ are established in \citet[Eqn.~6]{steinwart2021closer}, and in the discussion around it. 
\end{prooffor}

\begin{prooffor}{\maternk\ kernel: Part~\cref{item:matern_kernel}}
We claim that $\maternk(\matone, \mattwo)$ is $\floor{\matone-\frac d2}$-times continuously differentiable which immediately implies the bound~\cref{eq:matern_cover} using \cref{general_rkhs_covering_bounds}\cref{item:differentiable_kernel}.

To prove the differentiability, we use Fourier transform of \Matern kernels. 
For $\kernel = \maternk(\matone, \mattwo)$, let $\kappa:\real^d\to\real$ denote the function such that noting that $\kernel(x, y) = \kappa(x-y)$. Then using the Fourier transform of $\kappa$ from \citet[Thm~8.15]{wendland2004scattered}, and noting that $\kappa$ is real-valued, we can write
\begin{talign}
\kernel(x, y) = c_{\kernel, d} \int  \cos(\omega\tp(x-y)) (\mattwo^2 + \twonorm{\omega}^2)^{-\matone} d\omega
\end{talign}
for some constant $c_{\kernel, d}$ depending only on the kernel parameter, and $d$ (due to the normalization of the kernel, and the Fourier transform convention). Next, for any multi-index $a \in \natural_0^d$, we have
\begin{talign}
\abss{\partial^{a, a}\cos(\omega\tp(x-y)) (\mattwo^2 + \twonorm{\omega}^2)^{-\matone}}
\leq \prod_{j=1}^{d} \omega_j^{2a_j} (\mattwo^2 + \twonorm{\omega}^2)^{-\matone}
\leq \frac{\twonorm{\omega}^{2\sum_{j=1}^d a_j}} {(\mattwo^2 + \twonorm{\omega}^2)^{\matone}},
\end{talign}
where $\partial^{a, a}$ denotes the partial derivative of order $a$. Moreover, we have
\begin{talign}
\int\frac{\twonorm{\omega}^{2\sum_{j=1}^d a_j}} {(\mattwo^2 + \twonorm{\omega}^2)^{\matone}}  d\omega
= c_d \int_{r >0} r^{d-1} \frac{r^{2\sum_{j=1}^d a_j}}{(\mattwo^2 + r^2)^{\matone}} dr 
\leq c_d \int_{r>0} r^{d-1 + 2\sum_{j=1}^d a_j-2\matone } \stackrel{(i)}{<} \infty,
\end{talign}
where step~(i) holds whenever
\begin{talign}
d-1 + 2\sum_{j=1}^d a_j-2\matone < -1
\quad \Longleftrightarrow \quad 
\sum_{j=1}^{d} a_j < \matone - \frac d2.
\end{talign}
Then applying \citet[Lemma~3.6]{NEWEY19942111}, we conclude that for all multi-indices $a$ such that $\sum_{j=1}^{d} a_j \leq \floor{\matone- \frac d2}$, the partial derivative
$\partial^{a, a}\kernel$ exists and is given by
\begin{talign}
c_{\kernel, d}  \int \partial^{a, a}\cos(\omega\tp(x-y)) (\mattwo^2 + \twonorm{\omega}^2)^{-\matone} d\omega,
\end{talign}
and we are done.
\end{prooffor}

\begin{prooffor}{\imq\ kernel: Part~\cref{item:imq_cover}}
The bounds~\cref{eq:imq_cover,eq:imq_const} follow from \citet[Ex.~3]{sun2008reproducing}, and noting that $\coveringnumber_2(\ball_2(r), \wtil{r}/2)$ is bounded by $(1+\frac{4r}{\wtil{r}})^d$~\citep[Lem.~5.7]{wainwright2019high}.
\end{prooffor}

\begin{prooffor}{\sinc\ kernel: Part~\cref{item:sinc_cover}}
Note that
\begin{talign}
    \frac{1}{2\pi}\int_{\real} \indicator(|\xi| \leq \theta)  e^{-iz\xi}d\xi 
    = \frac{1}{2\pi}\int_{-\theta}^{\theta} \cos(z\xi)d\xi 
    = \frac{1}{2\pi} \frac{2\sin(\theta z)}{z} = \frac{\theta}{\pi}  \sinc(\theta z).
\end{talign}
and hence $\kappa(z) = \prod_{j=1}^d \sinc(\theta z_j)$ is the Fourier transform (see \cref{lem:compact_covering}) of
\begin{talign}
    \what \kappa(\xi) = (\frac{\pi}{\theta})^d \prod_{j=1}^d \indicator(|\xi_j| \leq \theta).
\end{talign}
Now we can apply \cref{lem:compact_covering} with $a=\theta$ and $\sinfnorm{\what{\kappa}} = (\frac{\pi}{\theta})^d$, to obtain
\begin{talign}
    N_{\kappa, a, d} = \max\braces{1, \ \  \ceil{2\theta},   \ \ \log((\frac{3\theta}{2\pi})^d \cdot \frac{32d}{3\vareps^2} \cdot \frac{\pi^d}{\theta^d})}
    = \max\braces{1, \ \ \ceil{2\theta}, \ \ \log((\frac{3}{2})^d \cdot \frac{32d}{3\vareps^2})}.
\end{talign}
Now that for $x, y \in [-r, r]^d$, we can define vectors $x'$ and $y'$ in $[0, 1]^d$ with $x_j'= (x_j+r)/2r$ and $y_j' \defeq (y_j+r)/(2r)$ for each $j \in [d]$ such that
\begin{talign}
    \sinc(\theta(x-y)) = \sinc(r\theta(x'-y')).
\end{talign}
And hence for $\kernel(x, y) = \sinc(\theta(x-y))$, we can consider $\kernel'(x, y) = \sinc(r\theta(x-y))$ so that $\entropy([-r, r]^d, \vareps) = \entropy[\kernel']([0, 1]^d, \vareps)$. Substituting $\theta \gets  r\theta$ into the definition of $N_{\kappa, a, d}$ above and invoking the bound~\cref{eq:entropy_bound_zhou} from \cref{lem:compact_covering} implies the desired claim.
\end{prooffor}

\begin{prooffor}{\bspline\ kernel: Part~\cref{item:spline_kernel}}
The analytical expression for the $2\beta+2$-recursive convolution of $\indicator_{[-\half, \half]}$ from \citet[App.~O.4.1]{dwivedi2021kernel} shows that the function $h_{\beta}:\real\to[0, 1]$ can be represented as a linear combination of functions $x \mapsto \max(a+x, 0)^{2\beta+1}$ for multiple different thresholds $a$ and consequently that $h_{\beta}$ is continuously differentiable $2\beta$ times on $\real$. 
Hence $\kernel(\x,\y) = \kappa(\x-\y)$ for $\kappa(\z)=\mfk{B}_{2\splineparam+2}^{-d} \prod_{\j=1}^d h_{\splineparam}(\mattwo\z_{\j})$ is $\beta$-times continuously differentiable %
since for all multi-indices $\alpha_1,\alpha_2 \in \N_0^{d}$, we have $\frac{\partial^{|\alpha_1|+| \alpha_2|}}{\partial^{\alpha_1}x\partial^{\alpha_2}y} \kernel(x,y) = (-1)^{|\alpha_2|}(\frac{\partial^{|\alpha_1|+| \alpha_2|}}{\partial^{\alpha_1+\alpha_2}z}\kappa)(x-y)$. As a result, $\bspline(2\beta+1, \gamma)$ satisfies the conditions of \cref{general_rkhs_covering_bounds}\cref{item:differentiable_kernel} with $s=\beta$ yielding the claim.
\end{prooffor}

\section{Proof of \cref{table:explicit_mmd} results}
\label{proof_of_table_explicit_mmd}

In \cref{table:explicit_mmd}, the stated results for all the entries in the \tkt column follow directly by substituting the covering number bounds from \cref{rkhs_covering_numbers} in the corresponding entry along with the stated radii growth conditions for the target $\P$. (We substitute $m=\half\log_2 n$ since we thin to $\sqrt{n}$ output size.) For the KT+ column, the stated result follows by either taking the minimum of the first two columns (whenever the \rkt guarantee applies) or using the \akt guarantee. First we remark how to always ensure a rate of at least $\order(n^{-\quarter})$ even when the guarantee from our theorems are larger, using a suitable baseline procedure and then proceed with our proofs.

\begin{remark}[Improvement over baseline thinning]
\label{rem:baseline}
First we note that the \ktswaplink step ensures that, deterministically,  $\mmd_{\kernel}(\inputcoreset, \ktcoreset) \leq  \mmd_{\kernel}(\inputcoreset, \basecoreset)$ and $ \mmd_{\kernel}(\P, \ktcoreset) \leq  2\mmd_{\kernel}(\P, \inputcoreset) + \mmd_{\kernel}(\P, \basecoreset)$ for $\basecoreset$ a baseline thinned coreset of size $\frac{n}{2^m}$ and any target $\P$. For example if the input and baseline coresets are drawn \iid and $\kernel$ is bounded, then $\mmd_{\kernel}(\inputcoreset, \ktcoreset)$ and $\mmd_{\kernel}(\P, \ktcoreset)$ are $\order(\sqrt{{2^m/}{n}})$ with high probability \citep[Thm.~A.1]{tolstikhin2017minimax}, even if the guarantee of \cref{theorem:mmd_guarantee} is larger. As a consequence, in all well-defined KT variants, we can guarantee a rate of $n^{-\quarter}$ for $\mmd_{\kernel}(\inputcoreset, \ktcoreset)$ when the output size is $\sqrt n$  simply by using baseline as \iid thinning in the \ktswap step.
\end{remark}

\paragraph{\gauss\ kernel} The \tkt guarantee follows by substituting the covering number bound for the Gaussian kernel from \cref{rkhs_covering_numbers}\cref{item:gauss_cover} in \cref{eq:mmd_simplified_bound}, and the \rkt guarantee follows directly from \citet[Tab. 2]{dwivedi2021kernel}. Putting the guarantees for the \rkt and \tkt together (and taking the minimum of the two) yields the guarantee for \textsc{KT+}.

\paragraph{\imq\ kernel}
The \tkt guarantee follows by putting together the covering bound~\cref{rkhs_covering_numbers}\cref{item:imq_cover} and the MMD bounds~\cref{eq:mmd_simplified_bound}.

For the \rkt guarantee, we use a square-root dominating kernel $\wtil{\mbf k}_{\mrm{rt}}$ \imq$(\matone', \mattwo')$~\citet[Def.2]{dwivedi2021kernel} as suggested by \cite{dwivedi2021kernel}.  \citet[Eqn.(117)]{dwivedi2021kernel} shows that $\wtil{\mbf k}_{\mrm{rt}}$ is always defined for appropriate choices of $\matone', \mattwo'$. The best \rkt guarantees are obtained by choosing largest possible $\matone'$ (to allow the most rapid decay of tails), and \citet[Eqn.(117)]{dwivedi2021kernel} implies with $\matone < \frac d2$, the best possible parameter satisfies $\matone' \leq \frac d4 + \frac \matone 2$. For this parameter, some algebra shows that $\max(\rmin_{\wtil{\mbf k}_{\mrm{rt}},n}^{\dagger} \rmin_{\wtil{\mbf k}_{\mrm{rt}},n})\precsim_{d, \matone, \mattwo} n^{1/2\matone}$, leading to a guarantee worse than $n^{-\quarter}$, so that the guarantee degenerates to $n^{-\quarter}$ using \cref{rem:baseline} for \rkt. When $\matone \geq \frac d2$, we can use a \maternk\ kernel as a square-root dominating kernel from \citet[Prop.~3]{dwivedi2021kernel}, and then applying the bounds for the kernel radii~\cref{eq:rmin_k}, and the inflation factor~\cref{eq:err_simplified} for a generic \Matern kernel from \citet[Tab.~3]{dwivedi2021kernel} leads to the
entry for the \rkt stated in \cref{table:mmd_rates}. The guarantee for KT+ follows by taking the minimum of the two.

\paragraph{\maternk\ kernel}
For \tkt, substituting the covering number bound from \cref{rkhs_covering_numbers}\cref{item:matern_kernel} in \cref{eq:mmd_simplified_bound} with $R = \log n$ and $\l\defeq \floor{\matone-\frac{d}{2}}>0$ yields the MMD bound of order
\begin{talign}
\label{eq:matern_k_explicit}
\sqrt{\frac{\log n \cdot (\log n)^d}{n^{1-d/(2\l)}}} = 
\frac{(\log n)^{\frac{d+1}{2}}}{n^{(2\l-d)/4\l}}
\end{talign}
which decays faster than $n^{-\quarter}$ only when $\l>d$ or equivalently $\matone > 3d/2$. The rate in \cref{eq:matern_k_explicit} simplifies to the entry in the \cref{table:explicit_mmd} when $\matone-\frac d2$ is an integer, i.e., when $\l=\matone-\frac d2$.
When $\matone \leq 3d/2$, we can simply use baseline as \iid\ thinning to obtain an order $n^{-\quarter}$ MMD error as in \cref{rem:baseline}.

The \rkt (and thereby KT+) guarantees for $\matone\!>\!d$ follow from \citet[Tab.~2]{dwivedi2021kernel}. 

When $\matone \in (\frac d2, d]$, we use \akt with a suitable $\aroot$ to establish the KT+ guarantee.
For \maternk$(\matone, \mattwo)$ kernel, the $\aroot$-power kernel  is given by \maternk$(\aroot\matone, \mattwo)$ if $\aroot\matone>\frac{d}{2}$ (a proof of this follows from \cref{def:root} and \citet[Eqns~(71-72)]{dwivedi2021kernel}). Since \laplace$(\gaussparam)=\maternk(\frac {d+1}2, \sigma\inv)$, we conclude that its $\aroot$-power kernel is defined for $\aroot>\frac{d}{d+1}$. And using the various tail radii~\cref{eq:rmin_k}, and the inflation factor~\cref{eq:err_simplified} for a generic \Matern kernel from \citet[Tab.~3]{dwivedi2021kernel}, we conclude that $\wtil{\err}_{\aroot} \precsim_{d, \kroot, \delta} \sqrt{\log n \log \log n}$, and $\max(\rmin_{\kroot,n}^{\dagger} \rmin_{\kroot,n})\precsim_{d, \kroot} \log n$, so that $\rmin_{\max} = \order_{d, \kroot}(\log n)$~\cref{eq:rmin_P} for \subexp\ $\P$ setting. Thus for this case, the MMD guarantee for $\sqrt n$ thinning with \akt (tracking only scaling with $n$) is
\begin{talign}
&\parenth{\frac{2^m}{n} \sinfnorm{\kroot}}^{\frac{1}{2\aroot}} (2\cdot\wtil{\err}_{\aroot})^{1-\frac{1}{2\aroot}} 
    \parenth{2
    \!+ \!
    \sqrt{\frac{(4\pi)^{d/2}}{\Gamma(\frac{d}{2}\!+\!1)}}
    \!\cdot \rmin_{\max}^{\frac{d}{2}} \cdot  \wtil{\err}_{\aroot}   }^{\frac1\aroot-1}\\
    &\precsim_{d, \kroot, \delta}
    (\frac{1}{\sqrt n})^{\frac1{2\aroot}} 
    (\sqrt{\cn \log n})^{1-\frac{1}{2\aroot}} 
    \cdot ((\log n)^{\frac d2 + \half} \sqrt {\cn})^{\frac1\aroot-1}
    =
    (\frac{\cn (\log n)^{1\!+\!2d(1\!-\!\aroot)}}{n})^{\frac{1}{4\aroot}}
\end{talign} 
where $\cn = \log\log n$; and we thus obtain the corresponding entry (for KT+) stated in \cref{table:explicit_mmd}. 

\paragraph{\sinc\ kernel}
The guarantee for \tkt follows directly from substituting the covering number bounds from \cref{rkhs_covering_numbers}\cref{item:sinc_cover} in \cref{eq:mmd_simplified_bound} as $\ball_2(\rin)\subseteq [-\rin,\rin]^d$.

For the \rkt guarantee, we note that the square-root kernel construction of \citet[Prop.2]{dwivedi2021kernel} implies that  $\sinc(\sincparam)$ itself is a square-root of  \sinc$(\sincparam)$ since the Fourier transform of $\sinc$ is a rectangle function on a bounded domain. 
However, the tail of the \sinc\ kernel does not decay fast enough for the guarantee of \citet[Thm.~1]{dwivedi2021kernel} to improve beyond the $n^{-\quarter}$ bound of \citet[Rem.~2]{dwivedi2021kernel} obtained when running \rkt with \iid baseline thinning.

In this case, \tkt and KT+ are identical since $\ksqrt = \kernel$.

\paragraph{\bspline\ kernel}
The guarantee for \tkt follows directly from substituting the covering number bounds from \cref{rkhs_covering_numbers}\cref{item:spline_kernel} in \cref{eq:mmd_simplified_bound}.

For the \bspline$(2\splineparam+1, \mattwo)$ kernel, using arguments similar to that in \citet[Tab.4]{dwivedi2021kernel}, we conclude that (up to a constant scaling) the $\aroot$-power kernel is defined to be $\bspline(A+1, \mattwo)$ whenever $A \defeq 2\aroot\splineparam+2\aroot -2 \in 2\natural_0$. Whenever the $\aroot$-power kernel is defined, we can then apply the various tail radii~\cref{eq:rmin_k} and the inflation factor~\cref{eq:err_simplified} from \citet[Tab.~3]{dwivedi2021kernel} to conclude that the MMD error rates for the \akt\ for \bndcase\ $\P$ are the same as \rkt up to factors depending on $\aroot$ and $\splineparam$, which as per \citet[Tab.~2]{dwivedi2021kernel} are of order $\sqrt{\log n/n}$.

For odd $\splineparam$ we can always take $\aroot=\half$ and $\bspline(\splineparam, \mattwo)$ is a valid (up to a constant scaling) square-root kernel~\citep{dwivedi2021kernel}. In this case, the \rkt guarantee is  $\sqrt{\log n/n}$, and the KT+ guarantee follows by taking the minimum MMD error for \tkt and \rkt.

For even $\splineparam$, we can choose $\aroot \defeq \frac{p+1}{\splineparam+1} \in (\half, 1)$ with  $p = \ceil{\frac{\splineparam-1}{2}} = \frac{\splineparam}{2} \in \natural$, which is feasible as long as $\splineparam \geq 2$. Thus $\bspline(\splineparam+1, \mattwo)$ is a suitable $\kroot$ for $\bspline(2\splineparam+1, \mattwo)$ for even $\splineparam \geq 2$ with $\aroot= \frac{\splineparam+2}{2\splineparam+2} \in (\half, 1)$. Since $\kroot$ is compactly supported, \cref{thm:a_root_kt} implies that  $\wtil{\err}_{\aroot} = \order_d(\sqrt{\log n})$ and $\rmin_{\max} = \order_d(1)$, and hence the MMD guarantee for $\sqrt n$ thinning with \akt (tracking only the scaling with $n$) is
\begin{talign}
&\parenth{\frac{2^m}{n} \sinfnorm{\kroot}}^{\frac{1}{2\aroot}} (2\cdot\wtil{\err}_{\aroot})^{1-\frac{1}{2\aroot}} 
    \parenth{2
    \!+ \!
    \sqrt{\frac{(4\pi)^{d/2}}{\Gamma(\frac{d}{2}\!+\!1)}}
    \!\cdot \rmin_{\max}^{\frac{d}{2}} \cdot  \wtil{\err}_{\aroot}   }^{\frac1\aroot-1}\\
    &\precsim_{d, \kroot, \delta}
    (\frac{1}{\sqrt n})^{\frac1{2\aroot}} 
    (\sqrt{\log n})^{1-\frac{1}{2\aroot}} 
    \cdot (\sqrt{\log n})^{\frac1\aroot-1}
    =
    (\frac{ \log n}{n})^{\frac{1}{4\aroot}}
    = (\frac{ \log n}{n})^{\frac{\beta+1}{2\beta+4}}.
\end{talign} 
Taking the minimum MMD error for \tkt and  $\aroot$-\akt yields the entry for KT+ stated in \cref{table:explicit_mmd} for even $\splineparam$.

\end{document}